\documentclass[pdflatex,sn-mathphys-num]{sn-jnl}

\usepackage{graphicx}%
\usepackage{multirow}%
\usepackage{amsmath,amssymb,amsfonts}%
\usepackage{amsthm}%
\usepackage{mathrsfs}%
\usepackage[title]{appendix}%
\usepackage{xcolor}%
\usepackage{textcomp}%
\usepackage{manyfoot}%
\usepackage{booktabs}%
\usepackage{algorithm}%
\usepackage{algorithmicx}%
\usepackage{algpseudocode}%
\usepackage{listings}%
\usepackage{caption}
\usepackage{subcaption}
\usepackage{tikz}
\usetikzlibrary{backgrounds}
\usetikzlibrary{positioning}
\usepackage{geometry}
\begin{document}
\title[Article Title]{Online Physics-Informed Dynamic Mode Decomposition: Theory and Applications}

\author[1]{\fnm{Biqi}\sur{ Chen}}\email{23B954005@stu.hit.edu.cn}
\author*[2]{\fnm{Ying} \sur{Wang}}\email{yingwang@hit.edu.cn}

\affil[1]{\orgdiv{School of Civil and Environmental Engineering}, \orgname{Harbin Institute of Technology(Shenzhen)}, 
\orgaddress{\street{Taoyuan Street}, \city{Shenzhen}, \postcode{518055}, \state{Guangdong}, \country{China}}}

\affil[2]{\orgdiv{School of Civil and Environmental Engineering}, \orgname{Harbin Institute of Technology(Shenzhen)}, 
\orgaddress{\street{Taoyuan Street}, \city{Shenzhen}, \postcode{518055}, \state{Guangdong}, \country{China}}}

\abstract{Dynamic Mode Decomposition (DMD) has received increasing research attention due to its capability to analyze and model complex dynamical systems. However, it faces challenges in computational efficiency, noise sensitivity, and difficulty adhering to physical laws, which negatively affect its performance.
Addressing these issues, we present Online Physics-informed DMD (OPIDMD), a novel adaptation of DMD into a convex optimization framework. 
This approach not only ensures convergence to a unique global optimum, but also enhances the efficiency and accuracy of modeling dynamical systems in an online setting. Leveraging the Bayesian DMD framework, we propose a probabilistic interpretation of Physics-informed DMD (piDMD), examining the impact of physical constraints on the DMD linear operator. Further, we implement online proximal gradient descent and formulate specific algorithms to tackle problems with different physical constraints, enabling real-time solutions across various scenarios. Compared with existing algorithms such as Exact DMD, Online DMD, and piDMD, OPIDMD achieves the best prediction performance in short-term forecasting, e.g.  an \( R^2 \) value of 0.991 for noisy Lorenz system. The proposed method employs a time-varying linear operator, offering a promising solution for the real-time simulation and control of complex dynamical systems.}




\maketitle

\section{Introduction}
In the realm of scientific and engineering challenges, the ability to understand and predict the behaviours of complex systems is highly valuable\cite{brunton2022data,xu2020support,fu2023updating}. These systems are often intricate, composed of numerous interacting components 
that exhibit different dynamic behaviors\cite{abdullahi2022long,achillopoulou2020monitoring}.  Gaining an understanding of these systems is vital for a variety of purposes, such as evaluating current conditions,  enhancing performance efficiency, predicting future conditions, and formulating control 
strategies\cite{wang2015damage,zhou2023automated}.
However, traditional analysis methods, including linear stability analysis and modal analysis, often find it difficult
to cope with the intricacies of these systems\cite{bruntonModernKoopmanTheory2021}. This is particularly true in cases where the system displays nonlinear characteristics or the analysis involves handling extensive datasets.

DMD emerges as a robust and flexible tool in this context, providing a novel approach to uncovering the dynamic essence of time series data from
 these complex systems\cite{kutz2016dynamic,schmid2011applications,schmidDynamicModeDecomposition2010}. 
Due to its capability to reveal intricate spatial and temporal patterns within high-dimensional datasets, DMD has 
been widely used to analyse nonlinear dynamical
systems\cite{askhamVariableProjectionMethods2018,tuDynamicModeDecomposition2014,proctorDynamicModeDecomposition2016,
kordaLinearPredictorsNonlinear2018,baiDynamicModeDecomposition2020,grosekDynamicModeDecomposition2014}.
Its applications span fluid dynamics\cite{schmidDynamicModeDecomposition2010}, neuroscience\cite{brunton2016extracting}, 
and epidemiology\cite{proctor2015discovering}, demonstrating its versatility and depth of insight into system 
behaviors\cite{rowleyModelReductionFlow2017}. However, it encounters challenges in terms of computational efficiency, noise sensitivity, and difficulty in complying with physical laws, which adversely affect its performance in real-time applications\cite{dawsonCharacterizingCorrectingEffect2016}.

Integrating physical principles into DMD has led to substantial improvements
in model accuracy and relevance. These advances are marked by the development of sparsity-promoting algorithms for Koopman-invariant subspaces, 
which enhance the identification of key physical processes in nonlinear systems\cite{pan2021sparsity}. 
Innovations such as piDMD\cite{baddooPhysicsinformedDynamicMode2023}, physics-fusion 
DMD \cite{yina2023pf}, physics-Fusion DMD\cite{yina2023pf} and geometry-informed DMD\cite{li2023geometry} address the integration of various 
types of physical information and the reduction of noise interference, achieving accurate dynamic modeling and improving prediction accuracy. 
Researchers have demonstrated its application in real-time and predictive settings by creating surrogate models with DMD in multiple real or simulated cases, significantly enhancing the model’s interpretability and reliability\cite{sharmaReviewPhysicsInformedMachine2023, 
 karniadakisPhysicsinformedMachineLearning2021, baddooKernelLearningRobust2022}. However, the influence of physical constraints on the DMD linear approximation of dynamics remains an area that requires further investigation, providing an open avenue for exploration despite these advancements \cite{baddooPhysicsinformedDynamicMode2023}.

Although piDMD successfully enhances the accuracy and noise resistance of DMD by extending the low-rank constraint to a manifold constraint, 
it retains certain limitations. Like standard DMD, piDMD was originally solved using a non-probabilistic, linear model-based approach. 
Consequently, this method lacks optimality guarantees, and its statistical performance remains undetermined without further assumptions about the probabilistic data structure. Bayesian DMD has improved the method, introducing a 
probabilistic approach to modal analysis through prior knowledge integration and uncertainty quantification\cite{takeishiBayesianDynamicMode2017}. 
Extensions such as Bayesian sparse DMD and probabilistic DMD mixtures
enhance DMD by accounting for observation noise and enabling posterior analysis. Variational Matrix Factorization in Bayesian DMD aligns with traditional DMD while providing a fully Bayesian framework, offering notable benefits for incomplete data handling\cite{kawashima2023gaussian}. This synergy between Bayesian methods and
optimized DMD algorithms produces robust and stable forecasts for spatio-temporal data, emphasizing the quantification of essential uncertainty in both spatial and temporal dimensions\cite{sashidharBaggingOptimizedDynamic2022}.

As data volumes from simulations and experiments expand, the computational costs of modal extraction algorithms such as DMD increase significantly\cite{erichsonRandomizedDynamicMode2019}. This increase in computational complexity is directly related to the growing quantity and dimensions of the incoming data, resulting in substantial computational burdens\cite{zhangOnlineDynamicMode2019}. Handling large-scale measurement data thus becomes challenging, especially in scenarios requiring online or real-time analysis. This initiative is supported by various studies, especially compressed sensing-like approaches to data compression and algorithmic optimization
  \cite{brunton2016compressed, bai2020dynamic,  erichson2019compressed,zhang2022unsteady}.
The essence of these efforts is to tailor DMD for high fidelity in aerodynamics, fluid dynamics, and large-scale thermofluid systems, catering to both overconstrained and underconstrained data sets.  
Recent efforts to optimize DMD for online applications have primarily focused on a common scenario in which the system evolves slowly over time while data arrives continuously \cite{zhangOnlineDynamicMode2019,hematiDynamicModeDecomposition2014}.
Notable progress in wind tunnel testing, EEG data examination, and wind farm forecasting demonstrates the enhanced efficiency and real-world relevance of these methods \cite{nedzhibovExtendedOnlineDMD2023,matsumoto2018online,alfatlawi2019incremental,
nedzhibov2023online,liewStreamingDynamicMode2022}. Despite these advancements, integrating physical constraints into DMD's online learning algorithms remains an unaddressed challenge.

In this study, we introduce OPIDMD, a novel method for the online analysis of time-varying systems. It represents a linear operator that best captures the data characteristics at the current time. By integrating the physical rigor of DMD with advancements in online analytics, OPIDMD offers an effective solution for accurately capturing the system state in real time while demonstrating significant advantages in noise robustness and generalization compared to existing online algorithms. It is particularly well-suited for low-dimensional, nonlinear dynamical systems with large amounts of sample data. We leverage convex 
 optimization techniques to integrate physical constraints with the probabilistic modeling of the linear approximation of dynamics, examining 
 the adaptability of various physical constraints within convex frameworks and the conversion of non-convex to convex optimization
  challenges. To demonstrate the generality of OPIDMD, we select eight fundamental physical principles for comparison: implicit constraint, self-adjointness, localization, causality,  shift-invariance, low-rank , along with LASSO ($l_1$ norm) and Tikhonov ($l_2$ norm) regularizations. To tailor these constraints, we designed efficient convex optimization solutions.

\begin{figure}[htb]
  \centering
  \includegraphics[width=1\linewidth]{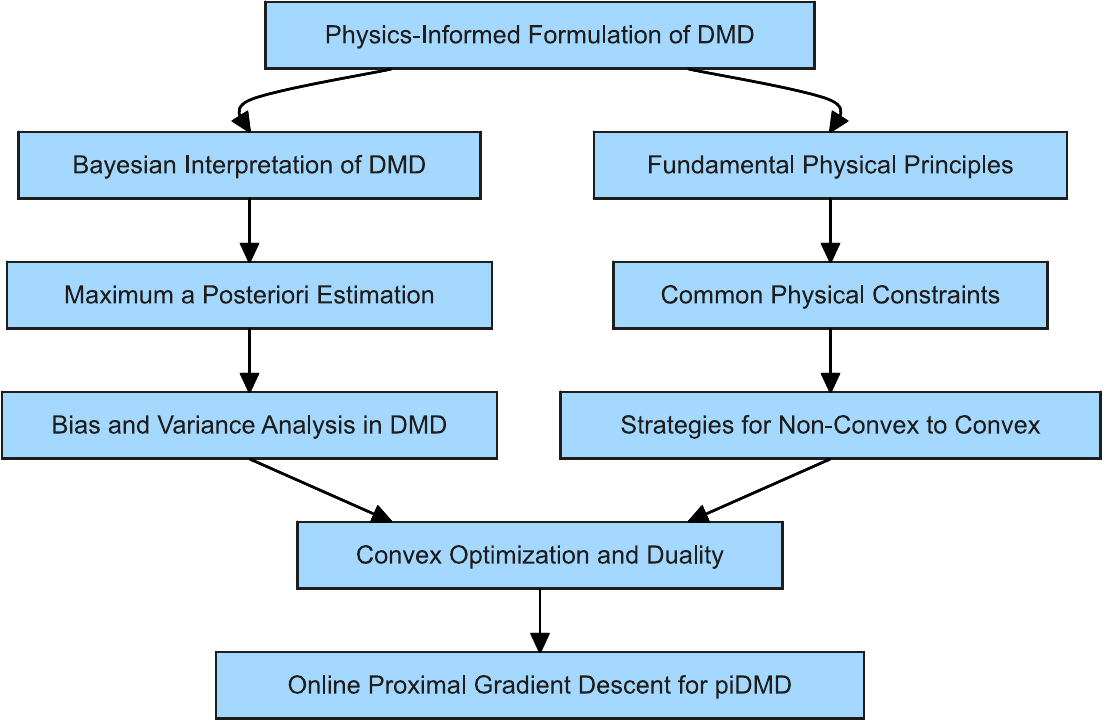}
  \caption{Framework of the Online Physics-Informed DMD. This diagram illustrates the comprehensive framework of this paper, emphasizing the physics-informed formulation of DMD,
   which can be interpreted in two ways. The left branch focuses on probabilistic aspects, such as Bayesian interpretation, 
   allowing us to perform bias-variance analysis in DMD. The right branch emphasizes physical principles; however, some physical 
   constraints may lead to non-convex optimization problems, necessitating a transition from non-convex to convex strategies. 
   By leveraging convex optimization and duality theory, both branches ultimately converge to the same solution. Finally, the 
   model can be solved using online proximal gradient descent for piDMD.}
  \label{fig1}
\end{figure}

Section \ref{sec:The theory of DMD} introduces the integration of physics insights with DMD and convex optimization, focusing on transforming non-convex problems into convex ones for improved analysis.
Section \ref{sec:Bias and Variance} merges DMD with Bayesian methods to explore their synergy and the impact of regularization on system forecasts, particularly with physical constraints.
Section \ref{sec:method} describes using online proximal gradient descent for convex optimization under physical constraints, emphasizing methodological precision and efficiency.
Section \ref{sec:result} presents the application and results of our proposed method, which uses the OPIDMD algorithm to learn approximate linear operators across multiple numerical examples.
Section \ref{sec:conclusion} concludes by highlighting our contributions and proposing future research directions. We also share an open-source Python implementation of OPIDMD at
   \href{https://github.com/Chen861368/Online-Physics-Informed-Dynamic-Mode-Decomposition/tree/Here%E2%80%99s-the-code-modified-according-to-the-reviewer%E2%80%99s-comments}{Github}. 
   The proposed framework is shown in Figure \ref{fig1}.
\section{Integrating Physical Knowledge into DMD via Convex Optimization}\label{sec:The theory of DMD}
\subsection{A physics-informed Formulation of DMD}
The traditional DMD is a data-driven approach designed to analyze the temporal evolution of a system by approximating its dominant dynamics. 
It assumes an unspecified dynamical system of the form \( y_j = x_{j+1} = F(x_j) \), where \( x_j, y_j \in \mathbb{R}^n \) are state vectors, and \( F : \mathbb{R}^n \rightarrow \mathbb{R}^n \) represents an unknown mapping. This framework is useful for analyzing discrete 
snapshots of continuous-time systems, enabling DMD to capture the underlying dynamics from sequential data snapshots \cite{schmidDynamicModeDecomposition2010}.

The implementation of DMD involves collecting pairs of system state snapshots at consecutive time steps, denoted by $\{ (x(t_k), y(t_k)) \}_{k=1}^m$. Here, each subsequent time \( t_{k+1} \) is defined as \( t_k + \Delta t \), where \( x_k = x(k\Delta t) \) represents the state at each time step. The interval $\Delta t$ is selected to be sufficiently small to accurately capture the system's highest frequency dynamics. These snapshots are then arranged into two data matrices, $X$ and $Y\in \mathbb{R}^{n\times m}$, as follows:
\begin{align}
  X & =\left[\begin{array}{cccc}
  \mid & \mid & & \mid \\
  x_1 & x_2 & \cdots & x_m \\
  \mid & \mid & & \mid
  \end{array}\right] 
\end{align}
\begin{align}
  Y & =\left[\begin{array}{cccc}
    \mid & \mid & & \mid \\
    y_1 & y_2 & \cdots & y_m \\
    \mid & \mid & & \mid
    \end{array}\right] 
\end{align}
Exact DMD optimizes a low-rank linear approximation operator \( A \) to model the dynamics of \( F \), such that the future state $y_j$ is approximated by $Ax_j$ for $j = 1, ..., m$. This is formalized by organizing the data into matrices $X$ and $Y$ and seeking $A$ as\cite{tuDynamicModeDecomposition2014}:
\begin{equation}\label{eqc_8}
  A = \underset{rank(A)\le r}{\mathrm{argmin}} \left\|Y - AX\right\|_F^2  
\end{equation}
where $\left \| \cdot  \right \|_F$ denotes the Frobenius norm. The rank constraint in equation \eqref{eqc_8} reflects the modal structure of the system but may not maintain 
other physical properties. One issue with DMD is that the solution to equation \eqref{eqc_8} is constrained to the column space of $Y$, potentially limiting the  generalizability of the model. To address this issue, Baddoo\cite{baddooPhysicsinformedDynamicMode2023} incorporated physics-informed constraints into the learning process.
 In piDMD , we can constrain $A$ within a matrix manifold reflective of physical laws:
\begin{equation}\label{eqc_9}
  A = \underset{A \in \mathcal{M}}{\mathrm{argmin}} \left\|Y - AX\right\|_F^2  
\end{equation}
Here, the manifold \( \mathcal{M} \subseteq \mathbb{R}^{n \times n}  \) is selected to guarantee that the solution matrix \( A \) conforms to specific established behaviors of the system. This choice embeds physical understanding into the optimization process delineated by equation \eqref{eqc_9}, thereby enriching the solution with meaningful physical constraints.

\subsection{Physics-Informed DMD via Convex Optimization}\label{sec:via}
In the field of DMD and its physics-informed variant, the analysis and modeling of dynamical systems are commonly formulated as optimization problems. 
This methodology effectively utilizes data to uncover the underlying system dynamics. However, after incorporating physical constraints, equation \eqref{eqc_9} often becomes challenging to solve. Typically, this requires constructing analytical solutions on a case-by-case basis. Deriving analytical expressions frequently needs to use numerical analysis, Fourier analysis,
  and other techniques, significantly complicating the resolution process. 

However, if we can transform piDMD problems into convex optimization problems, we can bypass complex mathematical solving techniques and directly apply efficient convex optimization methods.
A introduction to the convex optimization theory relevant to this paper is provided in Section 1 of the supplementary materials.
To incorporate the optimization function proposed for piDMD into the discussions of convex optimization, we first express the standard DMD algorithm equivalently in the following form:
\begin{align}\label{eqc_{18}}
  A &= \underset{A}{\mathrm{argmin}} \left\|Y - AX\right\|_F^2  \\
  &=\underset{A}{\mathrm{argmin}}\left [\text{tr}(X^TA^TAX-2Y^TAX)\right ]\label{eqc_{17}}
\end{align}
where \(\text{tr}(A)\) is the trace of matrix \(A\), defined as \(\sum_{i=1}^{n} a_{ii}\).
Clearly, equation \eqref{eqc_{17}} is an unconstrained convex optimization problem. 
If physical information about the dynamical system being studied is available, constraints derived from this information can be integrated,
 leading to the form of equation \eqref{eqc_9}.
 By examining the definition of convex optimization problems, we obtain Table \ref{table:convexity_matrix_constraints} , which clearly outlines how various matrix structures  influence the convexity of optimization problems.
\begin{table}[!h]
\centering
\caption{Convexity of optimization problems with matrix constraints}
\label{table:convexity_matrix_constraints}
\begin{tabular}{lll}
\hline
Matrix Structure $\mathcal{M}$ & Convex Set & Convex Problem  \\
\hline
Shift-Invariant (Circulant) & Yes & Yes \\
Self-Adjoint (Symmetric) & Yes & Yes  \\
Causal (Upper Triangular) & Yes & Yes \\
Local (Tri-Diagonal) & Yes & Yes  \\
Low-rank\cite{heas2022low}& No & No  \\
\hline
\end{tabular}
\vspace*{-4pt}
\end{table}

This table highlights the critical role of matrix structure in defining the convexity of optimization problems. Shift-invariant, self-adjoint, causal, and local matrices enable convex optimization due to their structural properties, which align with the requirements of convex sets. In contrast, low-rank constraints often introduce non-convex 
 challenges. To address the difficulties posed by non-convex low-rank optimization problems, we employ the following relaxation techniques.

To enforce a low-rank structure in matrices within optimization frameworks, a common strategy is to utilize the trace norm as a
 convex approximation of the rank function \cite{candes2010power}. The trace norm, defined as the sum of a matrix's singular 
 values and denoted by \( \|A\|_{*} = \sum_{i=1}^{r}\sigma_i(A) \), encourages low-rank solutions because it is the tightest 
 convex relaxation of the rank function. The optimization problem can be formulated as follows:
\begin{equation}\label{eqc_{25}}
  \underset{A}{\mathrm{argmin}}\quad \|Y - AX\|_F^2 \quad \text{subject to} \quad \|A\|_{*} \leq t
\end{equation}
where \( t \in (0, +\infty) \) serves as an upper bound that constrains the norm of the matrix \( A \). Selecting an appropriate threshold \( t \) is essential for controlling the sparsity of the singular values, thus influencing the low-rank characteristics 
of \( A \). Determining the optimal value of \( t \) typically requires empirical assessment, often through cross-validation. This allows us to examine how different values 
of \( t \) impact model performance, towards minimizing generalization error.

\section{Statistical Interpretation of DMD}\label{sec:Bias and Variance}
\subsection{Frequentist Point of View}
 To facilitate subsequent analysis of the DMD system matrix $A$ under physical constraints, we present a set of probabilistic assumptions in this section under which the standard DMD algorithm emerges as a highly intuitive method \cite{takeishiBayesianDynamicMode2017}.  We posit a relationship between the target variables and the inputs through the equation
\begin{equation}
  y_{i}=Ax_i+\epsilon_i
\end{equation}
where $\epsilon_i$ represents unmodeled effects or random noise. We assume the error term $\epsilon_i$ is independently and identically distributed (IID) according to a Gaussian distribution with zero mean and variance $\sigma^2 I$. 
The likelihood function for the observed data set, considering the 
independence of the $\epsilon_i$ term, is:
\begin{equation}
  L(A)=L(A;X,Y)=p(Y|X; A)= \prod_{i=1}^{m} p(y_{i}|x_i; A)
\end{equation}
The notation $p(y_{i}|x_i; A)$ denotes the distribution of $y_{i}$ given $x_i$, with $A$ as a 
deterministic but unknown parameter. Maximizing the likelihood principle suggests selecting $A$ that makes the observed data as probable as possible. 
Therefore, maximizing $l(A)$ is equivalent to minimizing\cite{jiang2022correcting}
\begin{equation}
  \left\| Y - A X \right\|_F^2 
\end{equation}
Under the probabilistic assumptions previously outlined, the DMD algorithm is consistent with the principle of maximum likelihood estimation. 
We infer that \( A \), although constant, is unknown, aligning with the frequentist interpretation of statistics. Given certain noise assumptions, 
the estimator’s statistical performance can be guaranteed. 
\subsection{Bayesian Point of View}
In contrast, from a Bayesian standpoint, $A$ is treated as a random variable, and its value is unknown\cite{bishop2006pattern}. In this approach, we 
 articulate our prior beliefs about the parameters through a prior distribution $P(A)$. The Bayesian view allows us to compute the 
 posterior distribution of the parameters, particularly when predicting new values of $x_i$. However, computing this posterior distribution is 
  computationally challenging due to the high-dimensional integrals over $A$, which are typically not feasible to solve in closed form.
 
  In practice, to circumvent these computational challenges, we approximate the posterior distribution for $A$. 
  A prevalent approach is to condense the posterior distribution into a single point estimate known as the maximum a posterior (MAP) 
  estimate\cite{bertsekas2008introduction}. The MAP estimate is calculated by maximizing the product of the likelihood and the prior
    probability of $A$:
 \begin{equation}\label{eqc_{22}}
   A_{\text{MAP}} = \arg\max_{A} \prod_{i=1}^{m} p(y_i|x_i, A)p(A).
 \end{equation}
 This approach is similar to the MLE estimate but incorporates the prior \( p(A) \). For parameters represented in vector form, Gaussian or Laplace distributions 
 are commonly used as priors. The inclusion of a Gaussian or Laplace prior often leads to a solution for \( A_{\text{MAP}} \) with a smaller 
 norm than the MLE estimate, thereby potentially reducing the risk of overfitting. 
 For matrix parameters, 
 we can consider a matrix normal distribution for $A$\cite{gupta2018matrix}. The probability 
 density function for a matrix $X \in \mathbb{R}^{n \times m}$ following the matrix normal distribution is given by:
\begin{equation}
  p(X | M, E, F) = \frac{\exp\left(-\frac{1}{2} \text{tr}\left[F^{-1} (X - M)^T E^{-1} (X - M)\right]\right)}{(2\pi)^{nm/2}|F|^{n/2}|E|^{m/2}}
\end{equation}
where \( M \) is \(  \mathbb{R}^{n \times m} \), \( E \) is \(  \mathbb{R}^{n \times n }\), and \( F \) is \(  \mathbb{R}^{m \times m }\), representing the mean matrix, row covariance matrix, and column covariance matrix of the distribution, respectively. The density is understood as the probability density function with respect to the standard Lebesgue measure in \( \mathbb{R}^{n \times m} \), i.e., the measure corresponding to integration with respect to \( dx_{11} dx_{21} \dots dx_{n1} dx_{12} \dots dx_{nm} \).
 
 The matrix normal distribution relates to the multivariate normal distribution in such a way that a matrix $X$ follows a matrix normal
 distribution $\mathcal{MN}_{n\times m}(M, E, F)$ if and only if the vectorization of $X$, denoted as $\mathrm{vec}(X) \sim \mathcal{N}_{nm}(\mathrm{vec}(M), F \otimes E)$
 where $\otimes$  denotes the Kronecker product\cite{gupta2018matrix}.
 Under the assumption that $A$ follows a matrix normal distribution,
  the log-likelihood function for $A$ can be derived, leading to the MAP estimate. 
 \begin{align}
   l(A) &= \log \prod_{i=1}^{m} p(y_i|x_i, A)p(A)  \\
   &= \log \prod_{i=1}^{m} \frac{\exp \left(-\frac{1}{2\sigma^2}\left \| y_i-Ax_i \right \|_2^2  \right)}{(2\pi)^{n/2}\sigma^n }  \frac{\exp\left(-\frac{1}{2} \text{tr}\left[F^{-1} (A- M)^T E^{-1} (A - M)\right]\right)}{(2\pi)^{n^2/2}|F|^{n/2}|E|^{n/2}}
 \end{align}
 This estimate minimizes the sum of the squared residuals, adjusted by a term derived from the prior distribution:
 \begin{equation}\label{eqc_{21}}
   A_{\text{MAP}} =\underset{A}{\mathrm{argmin}} \left ( \frac{1}{2\sigma^2} \sum_{i=1}^{m} \|y_i - A x_i\|_2^2 + \frac{1}{2} \text{tr} \left( F^{-1} (A - M)^T E^{-1} (A - M) \right) \right ) 
 \end{equation}
 If we further assume $A \sim \mathcal{MN}_{n \times n}(0, \frac{1}{\lambda} I, \sigma^2 I)$, this MAP estimation is equivalent to regularizing the standard DMD algorithm with a 
 $l_2$ norm penalty, effectively balancing the fit to the data against the complexity of the model:
 \begin{align}
   A_{\text{MAP}} &= \underset{A}{\mathrm{argmin}} \left ( \frac{1}{2\sigma^2} \sum_{i=1}^{m} \|y_i - A x_i\|_2^2 + \frac{1}{2} \text{tr} \left( (\sigma ^2I)^{-1} A^T (\frac{1}{\lambda} I)^{-1} A  \right) \right ) \\
   &=\underset{A}{\mathrm{argmin}}\left ( \left\|Y - AX\right\|_F^2+\lambda \left \| A\right \|_F^2\right ) 
 \end{align}
 
 In the context of matrix variate distributions, it is possible to construct properties similar to the Laplace distribution through the probability density 
 function in matrix form. The probability density function of the matrix variable $X$ under the matrix Laplace distribution $\mathcal{ML}_{n\times m}(M,b)$ is expressed as:
\begin{equation}
  p(X|M,b) = \frac{1}{(2b)^{nm}} \exp \left(-\frac{1}{b} \|X - M\|_1 \right), \quad   \|X - M\|_1 = \sum_{i=1}^{n} \sum_{j=1}^{m} |x_{ij} - m_{ij}|
\end{equation}
Here, \( \|X - M\|_1 \) denotes the entrywise \( l_1 \) norm of matrix \( X \) relative to matrix \( M \),
 where \( x_{ij} \) and \( m_{ij} \) represent the elements of \( X \) and \( M \), respectively.
 The parameters $n$ and $m$ represent the dimensions of the matrix.
 This measure is used to promote sparsity in the matrix variable $X$. When outliers are present in the observed data, 
 the robustness of the matrix Laplace distribution becomes advantageous.
 If we assume the prior $A$ follows a matrix Laplace distribution $\mathcal{ML}_{n \times n}(M,b)$, the MAP estimate becomes:
\begin{equation}
  A_{\text{MAP}} = \underset{A}{\mathrm{argmin}} \left ( \frac{1}{2\sigma^2} \sum_{i=1}^{m} \|y_i - A x_i\|_2^2 + \frac{1}{b} \|A - M\|_1 \right ) 
\end{equation}
 If we further assume $A \sim \mathcal{ML}_{n \times n}(0, \frac{2\sigma^2}{\lambda} )$, this MAP estimation becomes
\begin{equation}
  A_{\text{MAP}} =\underset{A}{\mathrm{argmin}}\left ( \left\|Y- AX\right\|_F^2+\lambda \|A\|_1 \right ) 
\end{equation}
 This adjustment encourages a sparse solution, which may enhance the DMD algorithm's interpretability in scenarios where simplicity 
 and robustness are key\cite{tibshirani1996regression}. As mentioned earlier, piDMD is represented by equation \eqref{eqc_9}. In contrast, 
 introducing a probabilistic prior through MAP estimation within a Lagrangian framework transforms the problem from a constrained 
 optimization to an unconstrained model with a penalty term:
\begin{equation}\label{eqc_{23}}
 A = \underset{A}{\mathrm{argmin}} \left\|Y - AX\right\|_F^2 + \lambda R(A)
\end{equation}
Here, $R(A)$ serves as a regularization term that implicitly imposes the physical constraints or probabilistic prior, 
while $\lambda \geq 0 $ acts as a Lagrange multiplier that quantifies the trade-off between data fidelity and adherence to the physical laws.
For the constraints discussed in this paper, as long as the constraints satisfy Slater's condition, the equivalence between optimization 
problem \eqref{eqc_9} and optimization problem \eqref{eqc_{23}} holds. Detailed proof can be found in Section 1 of the supplementary materials.

This unified approach not only simplifies computational complexity but also 
  ensures that the optimal solution is consistent with both physical understanding and probabilistic reasoning.  By utilizing the 
  probabilistic framework proposed earlier, we can analyze the impact of the physical constraint $R(A)$ on $A$ as well as its 
  effect on the prediction error. Generally speaking, implementing physical constraints can reduce variance, but it may also introduce bias because these constraints do not 
align with the underlying data structure.  Since this analysis can stand alone and is not closely related to the subsequent sections of the paper, it is provided in section 3 of the supplementary materials for consistency.
Based on the above analysis, we summarize the differences between OPIDMD and piDMD in Table \ref{table:comparison}.
  
  \begin{table}[!h]
  \centering
  \caption{Comparison of OPIDMD and piDMD}
  \label{table:comparison}
  \begin{tabular}{lll}
  \hline
  Aspect & OPIDMD & piDMD  \\
  \hline
  Form of Optimization Function & Essentially Identical & Essentially Identical \\
  Solution Method & Convex Optimization & Case-by-Case  \\
  Noise Assumption & Yes & No  \\
  Bias-Variance Analysis & Yes & No \\
  Interpretability & Yes & Yes  \\
  \hline
  \end{tabular}
  \vspace*{-4pt}
  \end{table}

  \section{Online Physics-informed Dynamic Mode Decomposition}\label{sec:method}
\subsection{Online Learning Algorithm}
Given the continuous arrival of data over time and the gradual evolution of system parameters, the ability for online updating becomes crucial. For instances where the system exhibits time-varying dynamics, it is expected to update the model continuously as 
new data becomes available. Given a series of snapshot pairs $(x_j, y_{j})$ for $j = 1, \ldots, k$, we construct the data matrices $X_k$ and $Y_k$, each with dimensions $n \times k$, as follows:
\begin{equation}
  X_k = [x_1 \quad x_2 \quad \ldots \quad x_k], \quad Y_k = [y_1 \quad y_2 \quad \ldots \quad y_{k}]
\end{equation}
Our objective is to determine an $n \times n$ matrix $A_k$ that approximates the relationship
 $A_kX_k \approx Y_k$. The DMD matrix $A_k$ is ascertained by minimizing the cost function:
\begin{equation}\label{eqc_{10}}
  J_k(A_k) = \sum_{i=1}^{k} \left\| y_{i} - A_k x_i \right\|^2_2 = \left\| Y_k - A_k X_k \right\|_F^2
\end{equation}
where \(\left \| \cdot  \right \|^2\) denotes the Euclidean norm on vectors.
Online learning algorithms are designed to incrementally update a model as new data becomes available\cite{hoi2021online,anderson2004towards}. 
The proposed methodology aims to minimize a loss function online, which incorporates both data fidelity and adherence to physical constraints.

In such scenarios, the online proximal gradient descent method plays a crucial role in achieving this objective. Detailed information on proximal gradient descent can be found in section 1 of the supplementary materials.
Leveraging the Bayesian prior analysis and the results from convex optimization regarding the equivalence of certain constrained and unconstrained optimization problems. The  table \ref{table:convexity_matrix} outlines the convex optimization problems associated with different matrix structures and their respective solution methods. The indicator function $I_C(x)$ is 
 defined as 
\begin{equation}
  I_C(x)=\left\{\begin{array}{ll}0 & x \in C \\ \infty & x \notin C\end{array}\right.
\end{equation}
 It succinctly emphasizes the concept that \( x \) must belong to the set \( C \) \cite{combettes2011proximal}. This type of indicator function can be interpreted as the result of a MAP estimation, where the prior probability \( P(A) \) is assumed to have a uniform distribution.
 
The primary approach involves using optimization techniques like projected gradient descent and the proximal gradient method for various constraint-induced scenarios. 
Notably, the implicit constraint exhibited by the stochastic gradient descent algorithm, as mentioned in Table \ref{table:convexity_matrix}, 
refers to its tendency to find a simpler solution even without an explicit norm constraint in the objective function\cite{woodworth2020kernel,haochen2021shape}. While the theoretical basis of this phenomenon is not yet fully understood, extensive empirical research indicates that stochastic gradient descent often converges to solutions that align more closely with the fundamental characteristics of the data. This typically results in solutions with a smaller norm, thereby reducing the risk of overfitting\cite{zou2021benefits}. Further details can be found in section 1 of the supplementary materials.

\begin{table}[!h]
  \centering
  \renewcommand{\arraystretch}{1.5} 
  \caption{Convexity of Optimization Problems with Matrix Constraints} 
  \label{table:convexity_matrix} 
  \begin{tabular}{lll} 
  \hline
  Matrix Structure  & Convex Problem & Solution Method \\
  \hline
  Implicit Constraint   & $  \|Y_k - A_kX_k\|_F^2$ & Stochastic Gradient Descent\\
  Shift-Invariant  &  $ \|Y_k - A_kX_k\|_F^2 + I_{\text{Circulant}}$ & Projected Gradient Descent \\
  Self-Adjoint   & $ \|Y_k - A_kX_k\|_F^2 + I_{\text{Symmetric}}$& Projected Gradient Descent \\
  Causal   & $ \|{Y_k} - A_kX_k\|_F^2 + I_{\text{Upper Triangular}}$  & Projected Gradient Descent \\
  Local   & $\|Y_k - A_kX_k\|_F^2 + I_{\text{Tri-Diagonal}}$ & Projected Gradient Descent \\
  Low-Rank  & $ \|Y_k - A_kX_k\|_F^2 +\lambda \|A_k\|_{*}$ & Proximal Gradient Method \\
  Sparse  & $\|Y_k - A_kX_k\|_F^2 +\lambda \|A_k\|_1$ & Proximal Gradient Method \\
  Norm Shrinkage & $\left\|Y_k - A_kX_k\right\|_F^2+\lambda \left \| A_k \right \|_F^2$ &  Gradient Descent Method \\
  \hline
  \end{tabular}
  \vspace*{-4pt}
  \end{table} 

\subsection{Online Proximal Gradient Descent for OPIDMD}

For the various convex optimization challenges presented in Table \ref{table:convexity_matrix}, there exists a myriad of specialized 
and efficient convex optimization algorithms tailored to each type\cite{boyd2004convex,nesterov2013introductory,rockafellar2015convex}.
 However, to maintain focus and simplicity, we introduce a general methodology applicable across these challenges rather than exploring the optimal algorithm for each distinct scenario. The convex optimization problems outlined in Table \ref{table:convexity_matrix},
  can be succinctly described by the following equation:
 \begin{equation}
 \min_{x} f(x) = g(x) + h(x),
 \end{equation}
 where $g$ represents a convex and differentiable function, and $h$ denotes a convex but potentially nondifferentiable function. 
Since the problems outlined in Table \ref{table:convexity_matrix} can be formulated in the form of equation \eqref{eqc_{26}}, they are amenable to optimization through mini-batch or stochastic proximal gradient descent methods\cite{bottou2010large,hinton2012neural}. 
\begin{equation}\label{eqc_{26}}
  A_k =\underset{A_k}{\mathrm{argmin}} \left\|Y_k - A_kX_k\right\|_F^2 + \lambda R(A_k)= \underset{A_k}{\mathrm{argmin}} \sum_{i=1}^{k} \left\| y_{i} - A_k x_i \right\|^2_2 + \lambda R(A_k)
 \end{equation}
 In this paper, given that data arrives sequentially over time, we adopt the online proximal gradient descent approach \cite{bottou2010large}.
 This method optimizes $A_k$ using one new data point at a time through proximal gradient descent, with the algorithm detailed in Algorithm \ref{algo:opdmd} .
 \begin{algorithm}
  \caption{Online Physics-informed Dynamic Mode Decomposition}\label{algo:opdmd}
  \begin{algorithmic}
  \State \textbf{Input}: Initial point $A_0$, step sizes $\{t_k\}$ chosen to be fixed and small, or via backtracking line search\cite{civicioglu2013backtracking}.
  \State \textbf{Goal}:  minimize 
  $$
  f(A) = g(A) + h(A)= \sum_{i=1}^{k} \left\| y_{i} - A x_i \right\|^2_2 + \lambda R(A)
  $$ over time, with data arriving online.
  \For{each iteration $k = 1, 2, \ldots$ with new data point $(x_{\text{new}}, y_{\text{new}})$}
      \State Update the objective function $g(A_k)=\left\| y_{i} - A_k x_i \right\|^2_2$ based on new data $(x_{\text{new}}, y_{\text{new}})$.
      \State Compute the gradient of updated $g_{\text{new}}(A_k)=\left\| y_{\text{new}} - A_k x_{\text{new}} \right\|^2_2$ at $A_k$: 
      $$
      \nabla g_{\text{new}}(A_k)=-2\times \left ( y_{\text{new}}-A_kx_{\text{new}} \right ) x_{\text{new}}^T
      $$
      \State Update $A_k$ using the online proximal gradient step:
      \State \quad $$A_{k+1} = \text{prox}_{t_kR}\left(A_k - t_k \nabla g_{\text{new}}(A_k)\right)$$
      \State \textbf{where} the proximal mapping $\text{prox}_{tR}(x)$ is defined as:
      \State \quad $$\text{prox}_{tR}(x) = \arg\min_{z} \left\{ \frac{1}{2t}\|x - z\|^2 + R(z) \right\}$$
  \EndFor
  \State \textbf{return}: A sequence of solutions $A_k$ adapting to the new data over time.
  \end{algorithmic}
\end{algorithm}

It's important to note that online proximal gradient descent does not guarantee that each update step will yield an exact solution for $A_k$. However, 
as the number of iterations increases and with the application of backtracking line search, the optimization algorithm will finally converge to an accurate solution. However, if the system's parameters are time-varying, then employing fixed step sizes $t_k$ is advisable. In practice, selecting different $t_k$ values can lead to varying rates of convergence, necessitating careful choice tailored to the specific problem at hand\cite{bottou2012stochastic}. Other optimization algorithms may be capable of adaptively adjusting $t_k$, but they are beyond the scope of this discussion. 

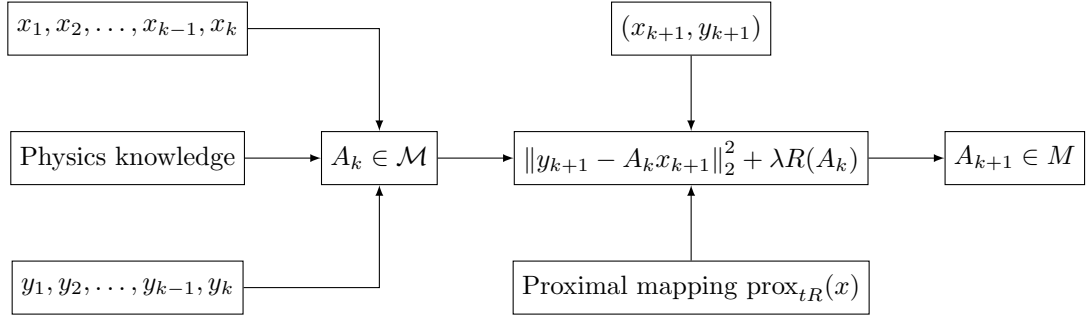
\begin{figure}[htbp]
    \centering
    \begin{tikzpicture}[node distance=1cm and 1cm, auto]
        \tikzstyle{block} = [rectangle, draw, text centered, minimum height=2em, minimum width=3em]
        \tikzstyle{data} = [draw, rectangle, minimum width=3em, minimum height=2em]
    \tikzstyle{block} = [rectangle, draw, text centered, minimum height=2em, minimum width=3em]
    \tikzstyle{data} = [draw, rectangle, minimum width=3em, minimum height=2em]
    
    \node [data] (xvalues) {\(x_1, x_2,  \ldots ,x_{k-1}, x_k\)};
    \node [block, below=of xvalues] (physics) {Physics knowledge};
    \node [data, below= of physics] (yvalues) {\(y_1, y_2,  \ldots,y_{k-1}, y_k\)};
    \node [block, right=of physics] (Ak) {\(A_k \in \mathcal{M}\)};
  
  \node [data, right=of Ak] (algorithm) {\(\left\| y_{k+1} - A_k x_{k+1} \right\|^2_2 + \lambda R(A_k)\)};
    \node [data, above=of algorithm] (xkplus) {\((x_{k+1},y_{k+1})\)};
    \node [block, right=of algorithm] (Akplus) {\(A_{k+1}\in M\)};
    \node [data, below=of algorithm] (proximal) {Proximal mapping $\text{prox}_{tR}(x)$};
  
    \draw [-latex] (xvalues.east) -| (Ak);
    \draw [-latex] (yvalues.east) -| (Ak);
    \draw [-latex] (physics.east) -- (Ak);
    \draw [-latex] (proximal) -- (algorithm);
    \draw [-latex] (xkplus) -- (algorithm);
    \draw [-latex] (Ak) -- (algorithm);
    \draw [-latex] (algorithm) -- (Akplus);
  \end{tikzpicture}
  \caption{A schematic representation of the OPIDMD algorithm process.}
  \label{fig:algorithm_process}
  \end{figure}

The setup of the OPIDMD algorithm is similar to that of the online DMD, but with added physical constraints, as illustrated in Figure \ref{fig:algorithm_process}. 
The foregoing section has outlined the general procedure for utilizing
 proximal gradient descent. Detailed algorithms for the proximal operator corresponding to each physical constraint, along with a comparison of computational times for OPIDMD, piDMD, and other online algorithms, are provided in the supplementary materials.

\section{Application and results}\label{sec:result}

To verify the effectiveness and performance of the OPIDMD algorithm, this paper presents six numerical examples, ranging from linear and nonlinear systems to those with chaotic characteristics. To test the robustness of the method against noise, 25\% non-stationary Gaussian white noise was added to all training data, while keeping the test data noise-free to  evaluate each algorithm's predictive performance. During the prediction phase, the DMD modal decomposition method was used, utilizing eigenvalues and eigenvectors to predict noise-free test data.

Since all physical laws are approximations of reality, with varying degrees of precision, or serve to simplify model complexity, this paper treats physical constraints primarily as tools to enhance prediction accuracy rather than as strict representations of the system's physical nature. Through these examples, the paper explores the predictive capabilities of the OPIDMD algorithm in noisy environments. This section, based on the $R^2$ prediction metric, presents results from multiple systems that demonstrate the clear advantage of the OPIDMD algorithm in short-term prediction accuracy.

Particularly, OPIDMD demonstrates strong generalization performance in high-dimensional nonlinear systems with limited sample data. In Table \ref{table:dmd_comparison}, both piDMD and OPIDMD show prediction results under their respective optimal physical constraints. True DMD refers to the model obtained using noise-free data and the Exact DMD algorithm, while all other models were trained on noisy data. Standard DMD refers to the DMD operator without low-rank constraints. Table \ref{table:constraints_comparison} 
summarizes the predictive performance of OPIDMD with different physical constraints across various systems, showing the effect of each constraint under
 different configurations in detail. 

\begin{table}[!h]
  \centering
  \caption{Performance Comparison of Different DMD Algorithms on Various Systems}
  \label{table:dmd_comparison}
  \begin{tabular}{lccccc}
  \hline
  System & True DMD & Exact DMD & piDMD & OPIDMD & ODMD \\
  \hline
  2D Advection & 0.920 & 0.808 & 0.818 & 0.830 & $-9.73\times 10^{196}$ \\
  Schrödinger Equation & 0.915 & 0.808 & 0.838 & 0.885 & 0.755 \\
  Advection-Diffusion & 0.870 & 0.705 & 0.723 & 0.846 & $-2.56\times 10^{204}$ \\
  Cylinder Flow & 0.941 & 0.938 & 0.935 & 0.935 & $-1.43\times 10^{39}$ \\
  Lorenz system & 0.950 & 0.066 & 0.062 & 0.991 & 0.066\\
  \hline
  \end{tabular}
  \vspace*{-4pt}
  \end{table}
Only a subset of examples is presented in the main paper, while additional examples for each OPIDMD constraint are provided in Section 2 of the supplementary materials. The specific details of the Schrödinger equation and Advection-Diffusion can be found in the supplementary material, with state dimensions of 200 and 100, respectively. The training data consists of 9,850 and 4,850 samples, while the testing data includes 150 samples for each. The noise-free data at the final training step is used as the initial condition for predicting the subsequent 150 steps. The cases in \ref{table:constraints_comparison}  focus on prediction performance. Furthermore, we include an example of a five-degree-of-freedom system to demonstrate how incorporating physical information can make the OPIDMD matrix structure more closely resemble that of the True DMD.

\begin{table}[!h]
  \centering
  \caption{Performance Comparison of Different OPIDMD Constraints on Various Systems}
  \label{table:constraints_comparison}
  \begin{tabular}{lp{1.5cm}p{1.5cm}p{1.5cm}p{1.5cm}p{1.5cm}}
  \hline
  Constraints & 2D Advection & Schrödinger Equation & Advection Diffusion & Cylinder Flow & Lorenz  system\\
  \hline
  OGD & 0.821 & 0.504 & 0.846 & 0.935 & 0.237 \\
  \( l_1 \) Norm & 0.813 & 0.874 & 0.846 & 0.935 & 0.235 \\
  \( l_2 \) Norm & 0.821 & 0.884 & 0.846 & 0.935 & 0.235 \\
  Low Rank & 0.814 & 0.885 & 0.846 & - & 0.235 \\
  Symmetric & 0.830 & 0.846 & 0.842 & 0.849 & - \\
  Circulant & - & 0.861 & - & - & - \\
  Tri-Diagonal & - & - & - & - & 0.991 \\
  \hline
  \end{tabular}
  \vspace*{-4pt}
\end{table}

Due to the unique dynamic characteristics 
of each system, it is essential to consider the impact of different physical constraints on prediction accuracy when selecting them.  
Based on this consideration, only general constraints such as online gradient descent (OGD), \( l_1 \) norm, \( l_2 \) norm, and low-rank 
constraints are applied to all systems to ensure model applicability and result comparability. 
In contrast, more targeted hard constraints, 
such as symmetry constraints, circulant constraints, and tri-diagonal structures, are only used in systems with similar characteristics, 
where they are expected to be more effective.

\subsection{Two-dimensional Advection Equation}

In this section, we use the two-dimensional advection equation as a test problem for numerical validation. The full form of the partial differential equation, along with the initial and boundary conditions, is described as follows:
\begin{equation}
  \begin{cases}
    \frac{\partial u(x, y, t)}{\partial t} + c_x \frac{\partial u(x, y, t)}{\partial x} + c_y \frac{\partial u(x, y, t)}{\partial y} = 0, & (x, y) \in (0, L_x) \times (0, L_y), \, t > 0 \\[10pt]
    u(x, y, 0) = \sin(\pi x) \sin(\pi y), & (x, y) \in [0, L_x] \times [0, L_y] \\[10pt]
    u(0, y, t) = u(L_x, y, t), \quad u(x, 0, t) = u(x, L_y, t), & t > 0
    \end{cases}
\end{equation}
Here, the advection velocities are set to \( c_x = 0.5 \) and \( c_y = 0.5 \), with the spatial domain lengths \( L_x = 1.0 \) and \( L_y = 1.0 \). The initial condition \( u(x, y, 0) = \sin(\pi x) \sin(\pi y) \) defines the system's state at \( t = 0 \), and the periodic boundary conditions ensure continuity of the solution along the boundaries in the \( x \) and \( y \) directions.

\begin{figure}[!ht]
  \centering
  \begin{subfigure}{0.41\textwidth}
      \centering
      \includegraphics[width=\textwidth]{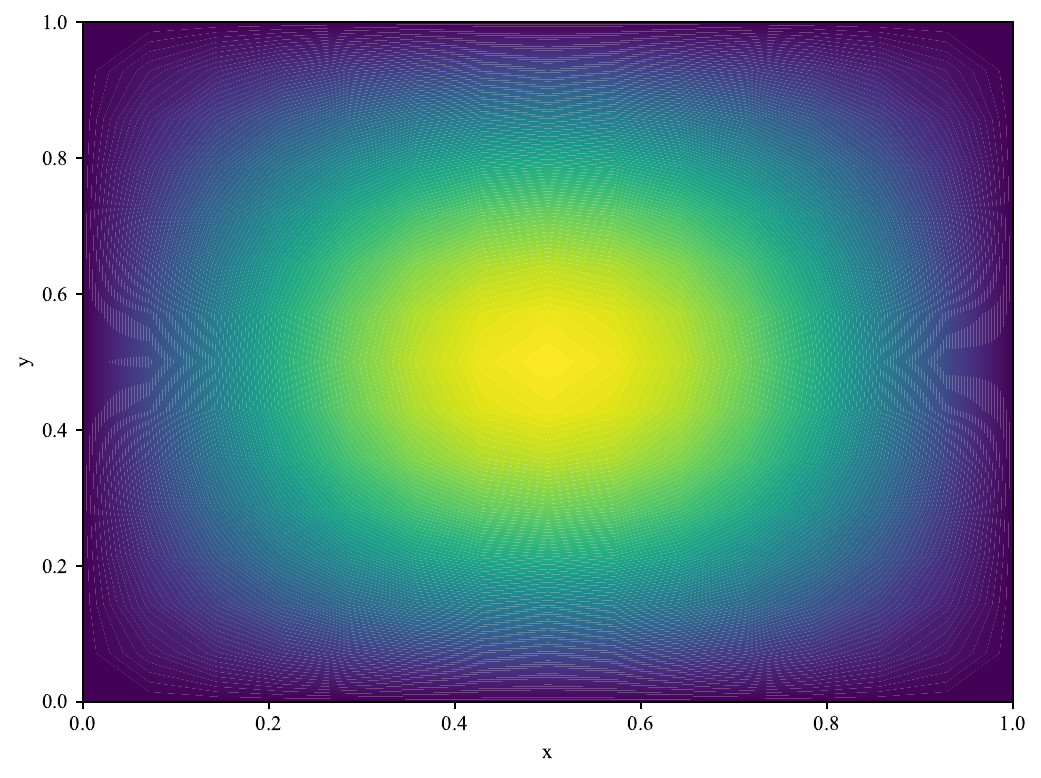}
      \caption{Without Noise}
  \end{subfigure}
  \hspace{1em}
  \begin{subfigure}{0.41\textwidth}
      \centering
      \includegraphics[width=\textwidth]{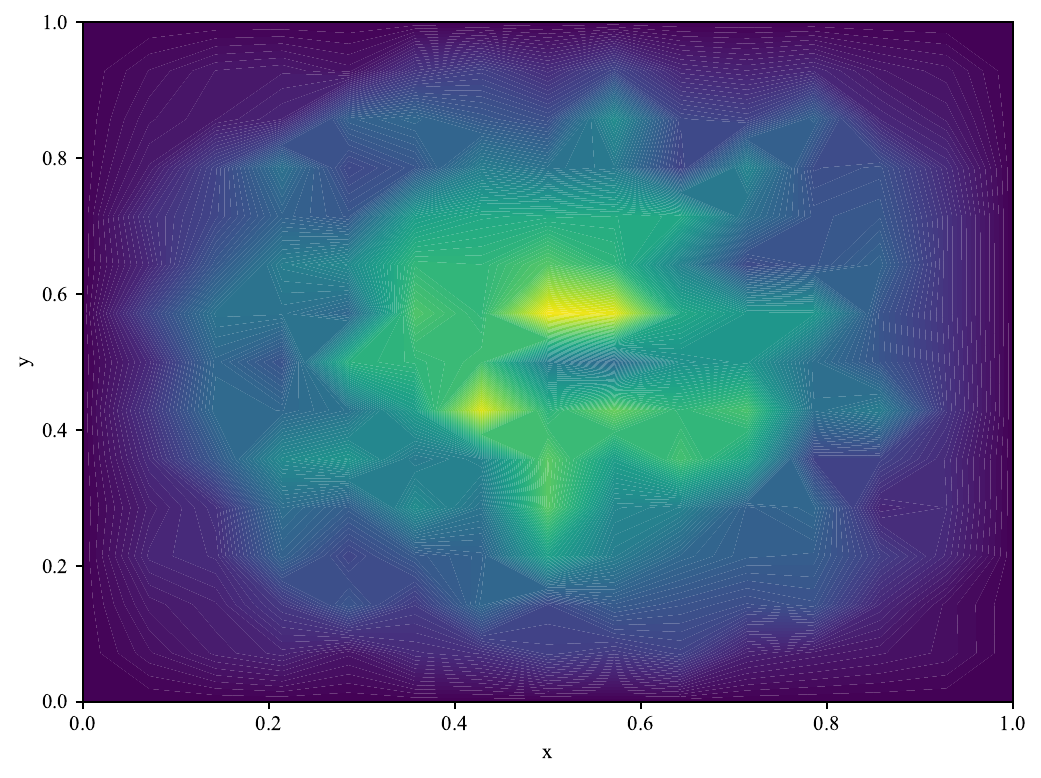}
      \caption{With Noise}
  \end{subfigure}
  \caption{Initial Conditions of the Two-Dimensional Advection Equation}
  \label{fig:150}
\end{figure}

To generate a uniform grid in both the \( x \) and \( y \) directions, we set the number of grid points as \( N_x = 15 \) and \( N_y = 15 \). For temporal discretization, the total time \( T = 1.0 \) is divided into \( N_t = 5000 \) time steps. An upwind scheme is used for the numerical discretization of the two-dimensional advection equation, which effectively reduces numerical dispersion. The numerical distribution of the initial condition is shown in Figure \ref{fig:150}.

\begin{figure}[!ht]
    \centering
    \begin{subfigure}{0.29\textwidth}
        \centering
        \includegraphics[width=\textwidth]{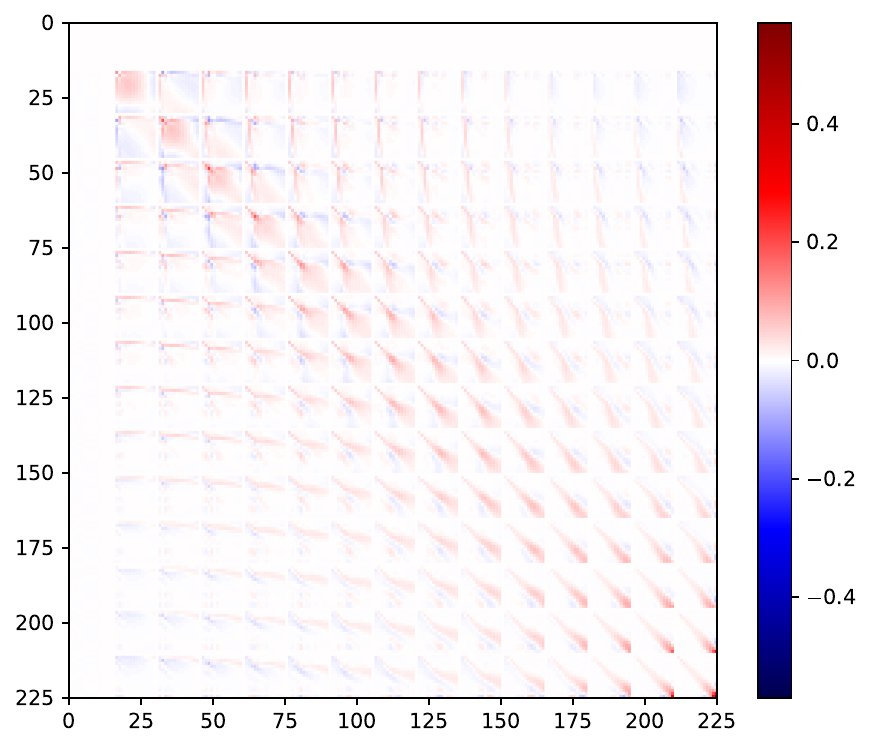}
        \caption{True DMD}
    \end{subfigure}
    \hspace{2em}
    \begin{subfigure}{0.29\textwidth}
        \centering
        \includegraphics[width=\textwidth]{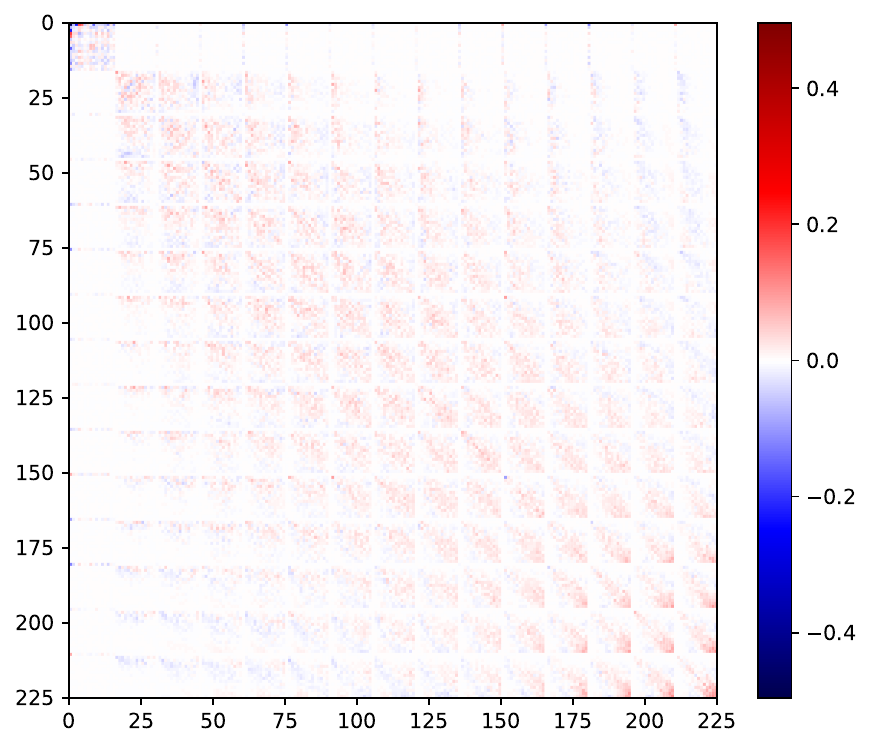}
        \caption{Symmetric piDMD}
    \end{subfigure}
    \hspace{2em}
    \begin{subfigure}{0.29\textwidth}
        \centering
        \includegraphics[width=\textwidth]{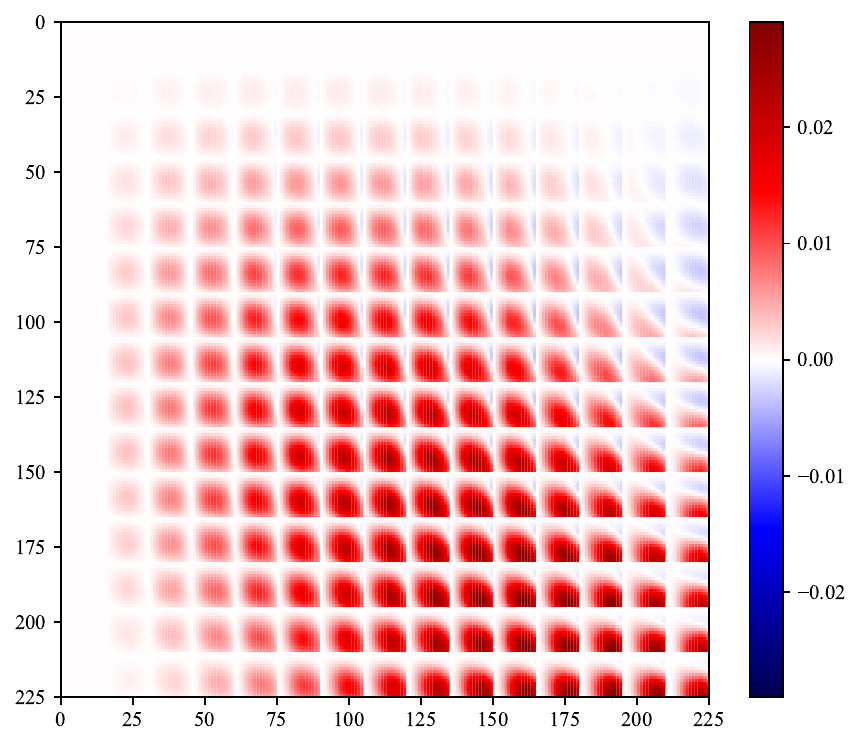}
        \caption{OGD}
    \end{subfigure}
  
    \vspace{1em}

    \begin{subfigure}{0.29\textwidth}
        \centering
        \includegraphics[width=\textwidth]{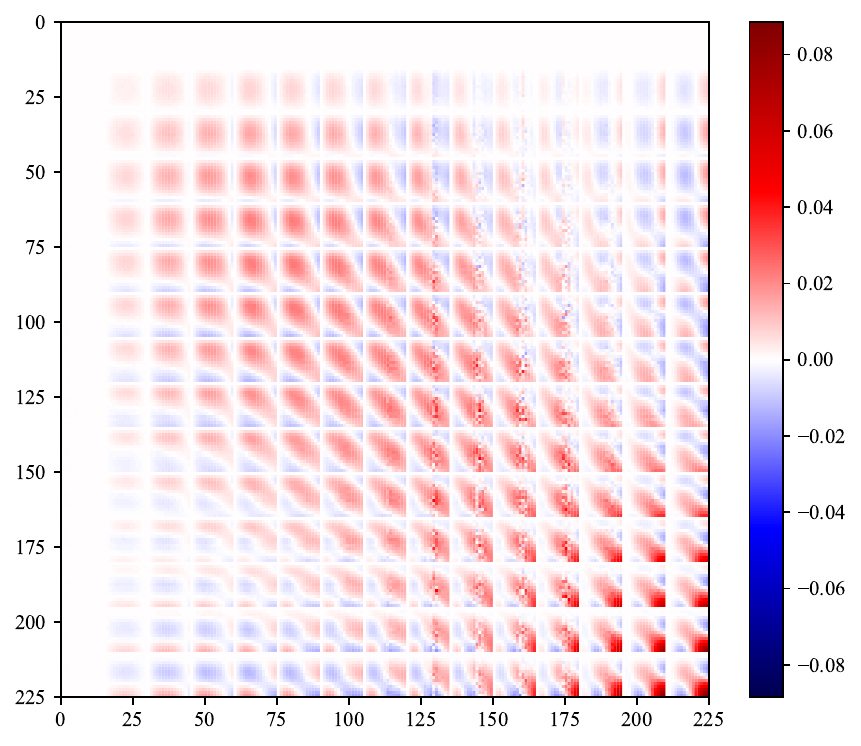}
        \caption{Exact DMD}
    \end{subfigure}
    \hspace{2em}
    \begin{subfigure}{0.29\textwidth}
        \centering
        \includegraphics[width=\textwidth]{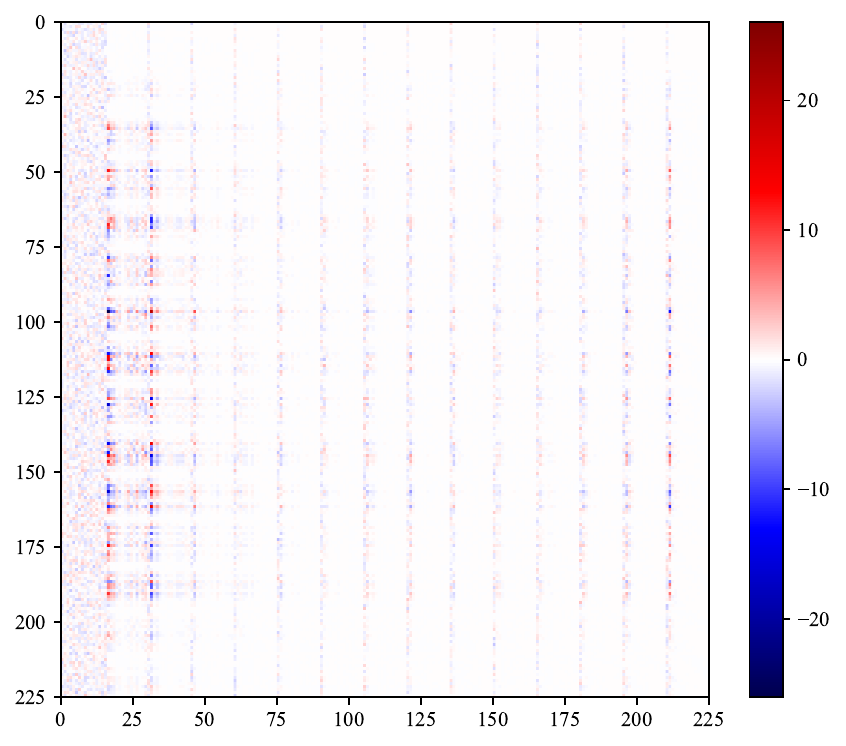}
        \caption{Online DMD}
    \end{subfigure}
    \hspace{2em}
    \begin{subfigure}{0.29\textwidth}
        \centering
        \includegraphics[width=\textwidth]{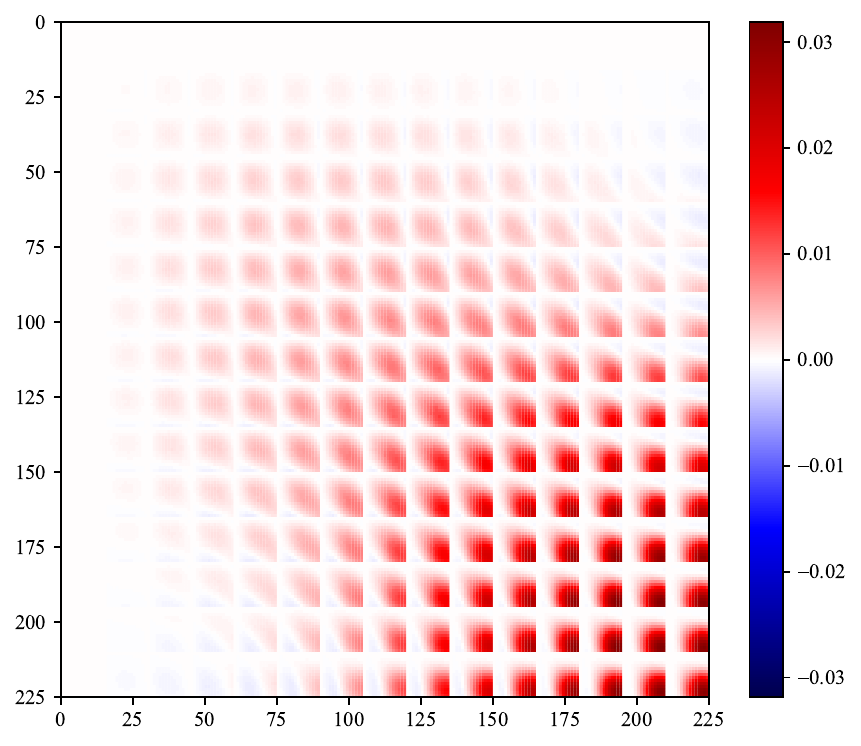}
        \caption{Symmetric OPIDMD}
    \end{subfigure}
    \caption{Comparison of DMD Matrices}
    \label{fig:Two-dimensional1}
\end{figure}

\begin{figure}[!ht]
    \centering
    \begin{subfigure}{0.29\textwidth}
        \centering
        \includegraphics[width=\textwidth]{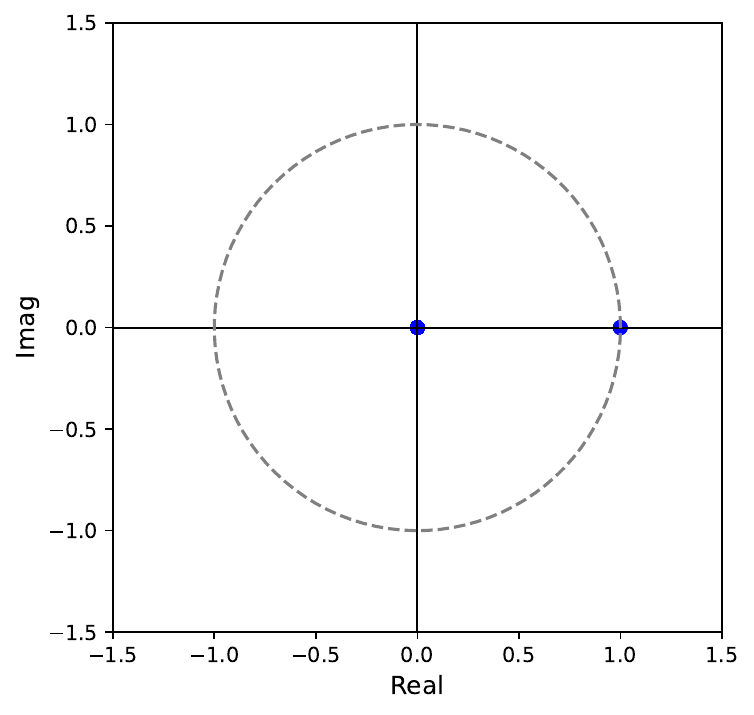}
        \caption{True DMD}
    \end{subfigure}
    \hspace{2em}
    \begin{subfigure}{0.29\textwidth}
        \centering
        \includegraphics[width=\textwidth]{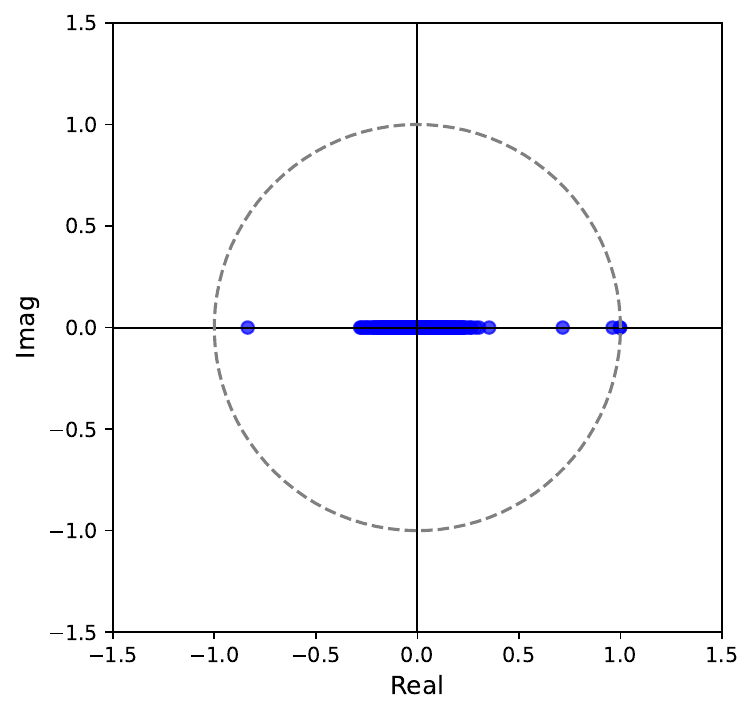}
        \caption{Symmetric piDMD}
    \end{subfigure}
    \hspace{2em}
    \begin{subfigure}{0.29\textwidth}
      \centering
      \includegraphics[width=\textwidth]{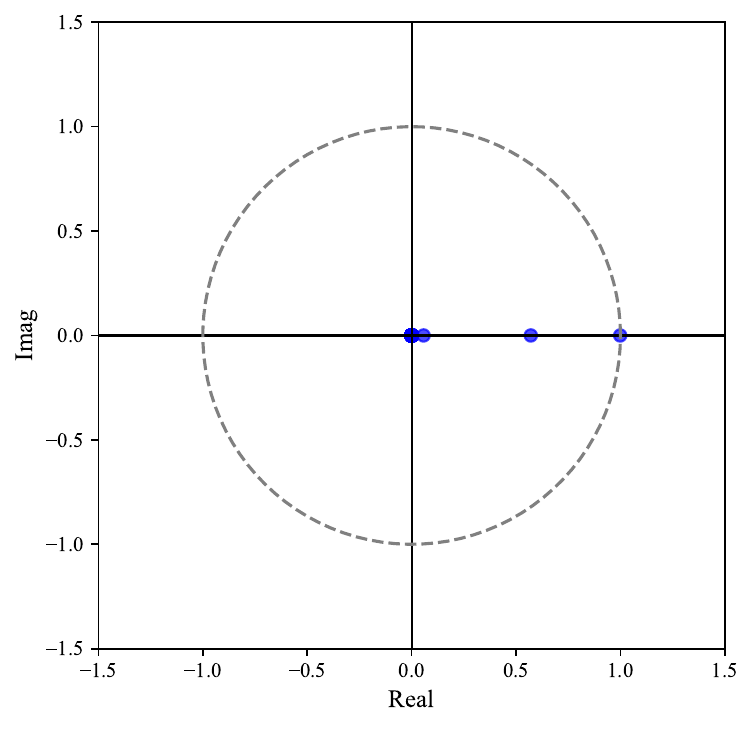}
      \caption{OGD}
  \end{subfigure}
  
    \vspace{1em}
    \begin{subfigure}{0.29\textwidth}
        \centering
        \includegraphics[width=\textwidth]{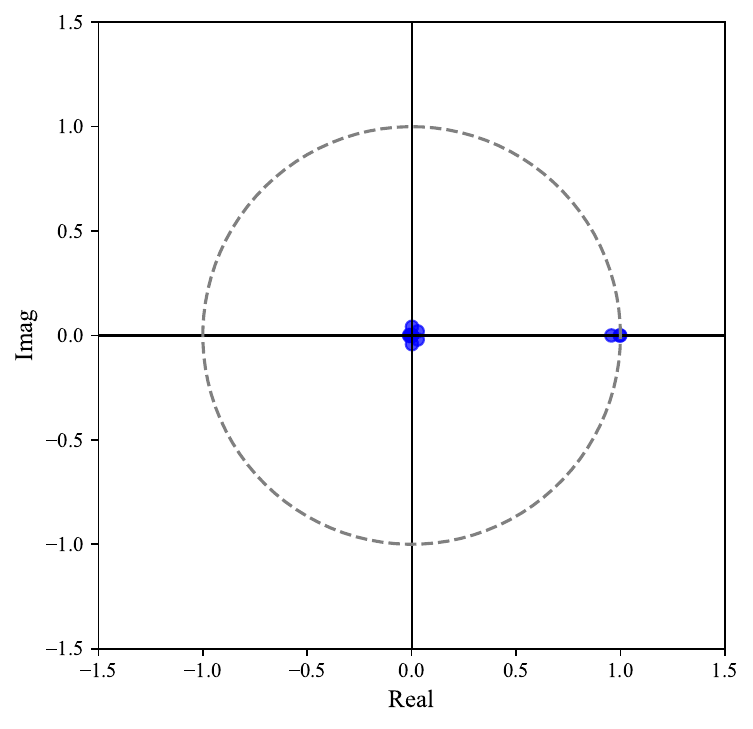}
        \caption{Exact DMD}
    \end{subfigure}
    \hspace{2em}
    \begin{subfigure}{0.29\textwidth}
        \centering
        \includegraphics[width=\textwidth]{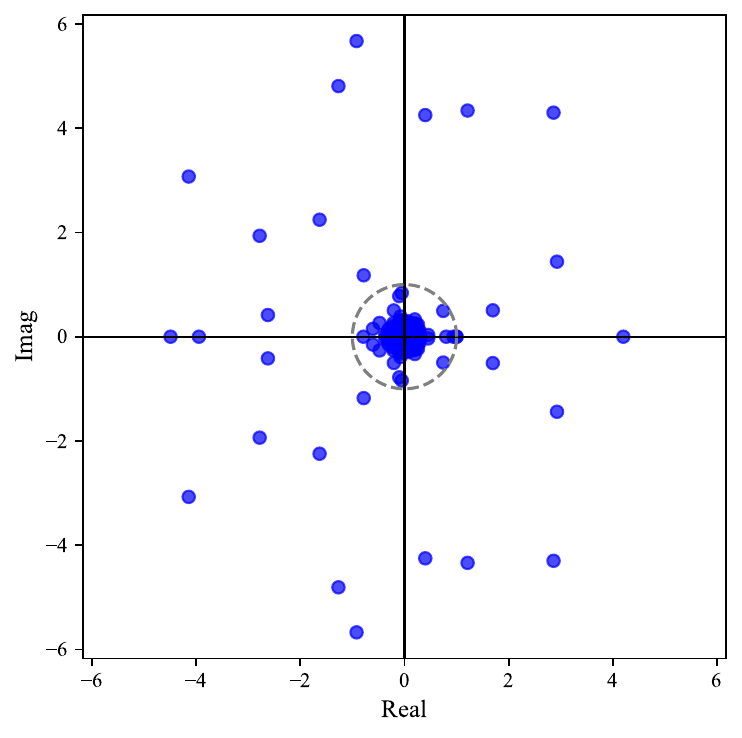}
        \caption{Online DMD}
    \end{subfigure}
    \hspace{2em}
    \begin{subfigure}{0.29\textwidth}
      \centering
      \includegraphics[width=\textwidth]{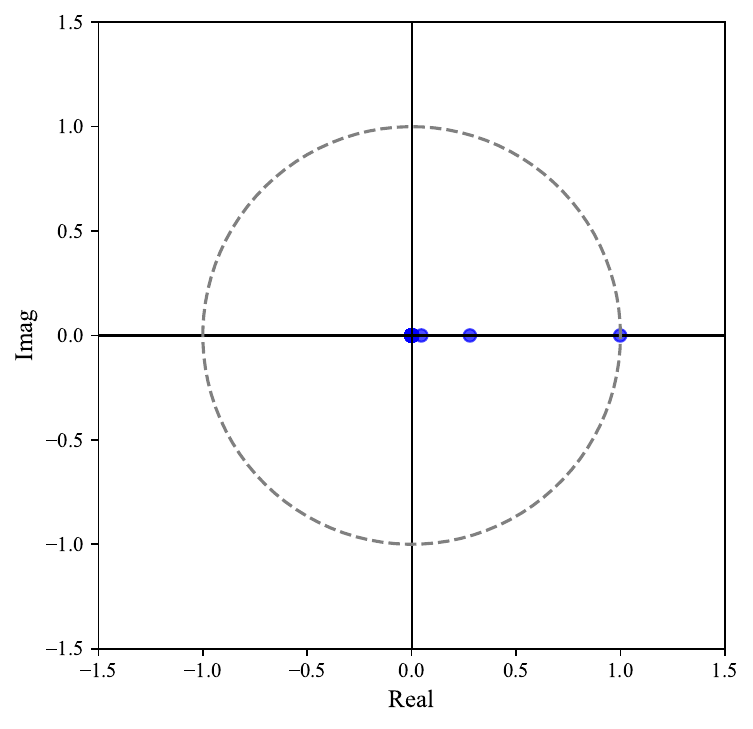}
      \caption{Symmetric OPIDMD}
  \end{subfigure}
  \caption{Eigenvalue Distributions}
    \label{fig:Two-dimensional2}
  \end{figure}

In this study, we use the first 4850 noisy time steps to train the model, with the noise-free data at step 4850 as the initial condition for 
predicting the final 150 noise-free steps. Figures \ref{fig:Two-dimensional1} and \ref{fig:Two-dimensional2} display the DMD matrices and their
 corresponding eigenvalues obtained from different algorithms.
The majority of the eigenvalues of the Exact DMD matrix are close to the origin, and the matrix structure also resembles that of the True DMD.

Despite the absence of any explicit constraints, the matrix structure obtained by the OGD method is highly simplified, resembling
 a low-rank matrix. This characteristic is also reflected in its eigenvalue distribution, with most eigenvalues clustered near zero and only a few 
 located around the unit circle. This low-rank behavior indicates the strong smoothing properties of this method, making it well-suited for noisy data
  environments, effectively suppressing noise.

Under symmetric constraints, the matrix generated by the OPIDMD method exhibits notable symmetry, with a relatively concentrated eigenvalue distribution, most of which lie on or near the origin. This result suggests that the OPIDMD method effectively approximates the primary modes of the system and inherits certain characteristics from the OGD matrix. In contrast, the symmetric piDMD matrix also displays a high degree of symmetry; however, its eigenvalues are more densely distributed along the real axis over a wider range, implying that the matrix rank may be high.

The matrix structure produced by the online DMD method is more complex, with a dispersed eigenvalue distribution, some of which are far from the unit circle. This behavior reflects potential stability issues within the system and may be attributed to the overfitting of noisy data by the online DMD method, which compromises model stability and increases the risk of overfitting.

\subsection{Cylinder Flow}

To assess OPIDMD’s performance in a nonlinear system, we selected the cylinder flow problem, a benchmark in fluid dynamics known for its energy-conserving 
properties and exhibiting \( n \gg k \) characteristics. The Reynolds number is set to \( Re = \frac{DU}{\nu} = 100 \), where \( D \) is the cylinder 
 diameter, \( U \) is the free-stream velocity, and \( \nu \) is the kinematic viscosity.

\begin{figure}[!ht]
  \centering
  \begin{subfigure}{0.38\textwidth}
      \centering
      \includegraphics[width=\textwidth]{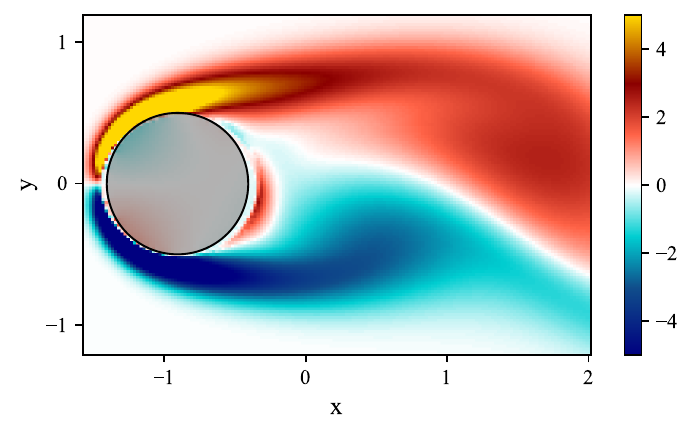}
      \caption{Without Noise}
  \end{subfigure}
  \hspace{1em}
  \begin{subfigure}{0.401\textwidth}
      \centering
      \includegraphics[width=\textwidth]{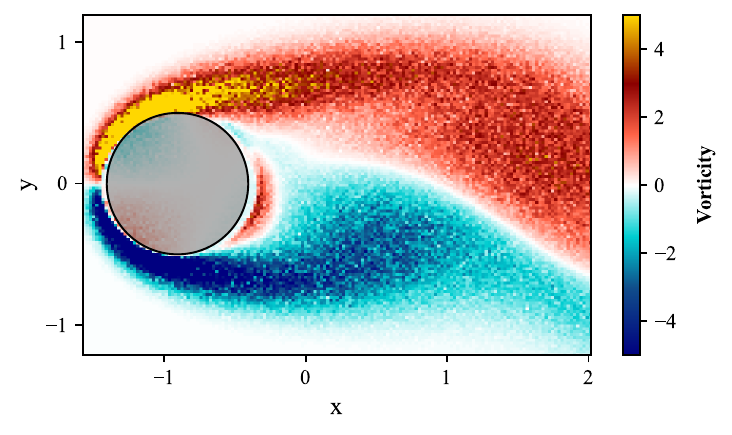}
      \caption{With Noise}
  \end{subfigure}
  \caption{Initial Conditions of the Cylinder Flow Data}
  \label{fig:112}
\end{figure}

The dataset is sourced from \cite{kutz2016dynamic}, with an original dimension of \( (199 \times 449, 151) \). For computational efficiency, a
 portion of the region was selected and reduced to \( (119 \times 161, 151) \). Figure \ref{fig:112} displays both the original data and its noisy 
 version. In this experiment, the model was trained on the first 140 time steps, while the remaining 11 time steps were used for testing to evaluate 
 predictive performance.

Since we have not derived physical constraints for the conservative system, we validated the OPIDMD algorithm with no physical constraints
 (i.e., degenerating into the OGD algorithm) and with \( l_1 \) and \( l_2 \) norm constraints. Given the computational 
 complexity of calculating low-rank norms, it was excluded from this experiment. Due to limited data, multiple training rounds were  performed 
 on the OPIDMD model, and the final results were compared with Exact DMD, piDMD, and online DMD. 
  Since all eigenvalues lie on the unit circle, piDMD used all eigenvalues and eigenvectors for prediction,  while other models utilized the 
  first 50 eigenvalues and their corresponding eigenvectors

\begin{figure}[!ht]
   \centering
   \begin{subfigure}{0.29\textwidth}
       \centering
       \includegraphics[width=\textwidth]{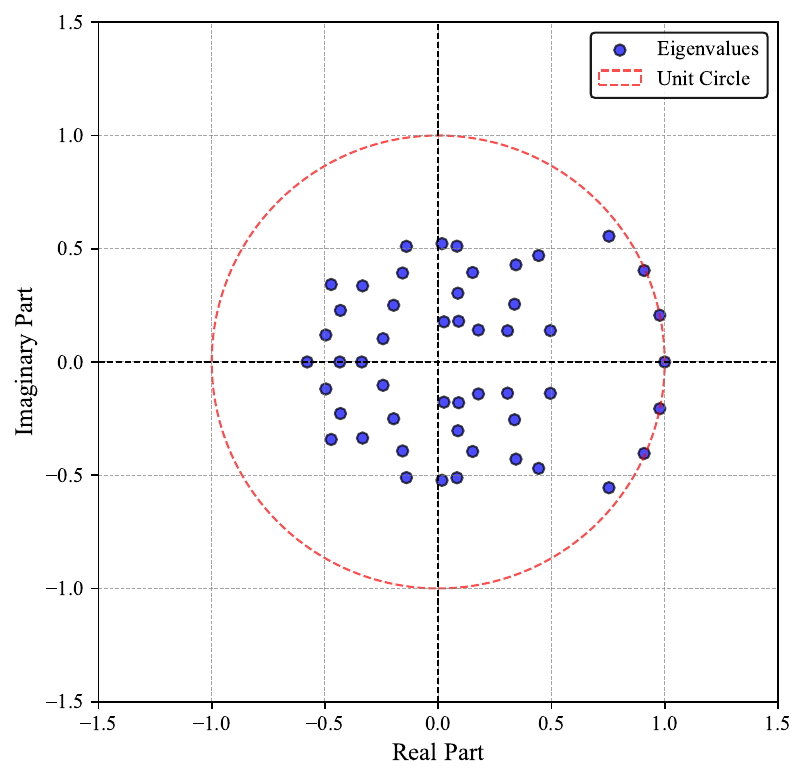}
       \caption{Exact DMD}
   \end{subfigure}
   \hspace{0.5em}
   \begin{subfigure}{0.29\textwidth}
       \centering
       \includegraphics[width=\textwidth]{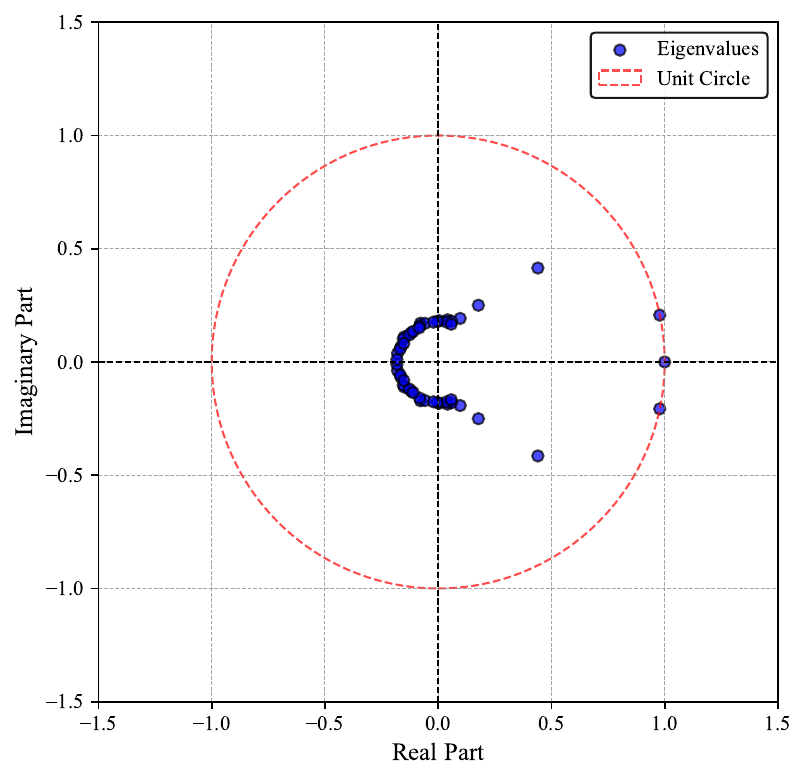}
       \caption{OPIDMD}
   \end{subfigure}
   \hspace{0.5em}
   \begin{subfigure}{0.29\textwidth}
       \centering
       \includegraphics[width=\textwidth]{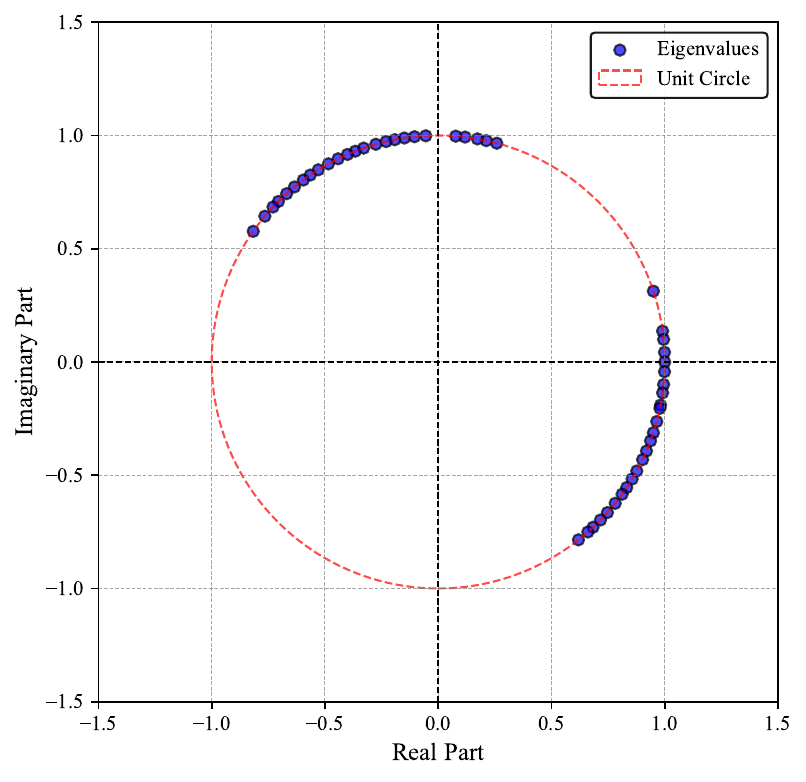}
       \caption{piDMD}
   \end{subfigure}
   \caption{Visualizations of the first 50 eigenvalues used in different algorithms}
   \label{fig:117}
\end{figure}

In the classic cylinder flow problem, DMD modes typically exhibit symmetry along the horizontal axis, forming symmetric vortex structures on 
either side of the cylinder. This flow characteristic leads to a symmetric eigenvalue distribution in the complex plane, reflecting similar 
dynamical behavior in symmetric directions. Figure \ref{fig:17} shows the eigenvalue distributions of the first 50 eigenvalues for Exact DMD,
 piDMD, and OPIDMD. Both Exact DMD and OPIDMD display symmetry, revealing the dynamic symmetry of the system. However, piDMD, due to its 
 constrained eigenvalues on the unit circle, struggles to distinguish the primary 50 eigenvalues in numerical solutions, failing to capture 
 the same symmetry. Since the eigenvalues of online DMD lie entirely outside the unit circle, they are not shown in Figure \ref{fig:117}.

The \( R^2 \) values for Exact DMD, OGD, Online DMD, and piDMD were 0.938, 0.935, \(-1.426 \times 10^{39}\), and 0.935, respectively.
 Testing showed that the one-step predictions of Online DMD perfectly fit the noisy test data, leading to overfitting and poor prediction results. For this 
 dataset, given the short temporal dimension, Exact DMD proved optimal as a global linear model and achieved the best predictive performance. 
The results indicated that both the unconstrained OGD and the \( l_1 \) and \( l_2 \)-regularized OPIDMD models performed similarly in terms of predictive accuracy. Since the \( l_1 \) and \( l_2 \)-regularized OPIDMD models showed only slight structural variations, only the results for OGD are shown in the figure.

This result suggests that even without explicit physical constraints, OPIDMD can achieve satisfactory predictive performance in nonlinear systems. It 
can be inferred that in scenarios where \( k \gg n \), OPIDMD may outperform other algorithms in local predictive performance. Furthermore, the eigenvalue 
distribution reveals that, similar to other cases, OGD tends to learn simpler models, with a few large eigenvalues while the rest are 
concentrated near the origin. Consequently, even without physical constraints, the model exhibits strong noise resistance.

\subsection{Lorenz System}

The Lorenz system is one of the most classic examples in deterministic dynamical systems and a typical model for studying chaotic phenomena. This system is characterized by a set of three coupled nonlinear ordinary differential equations describing its dynamic behavior, given as follows:
\begin{align}
   \dot{x} &= \sigma (y - x), \\
   \dot{y} &= x(\rho - z) - y, \\
   \dot{z} &= xy - \beta z,
\end{align}
where \( x \), \( y \), and \( z \) are the state variables representing the current state of the system. The parameters \( \sigma \), \( \rho \), and \( \beta \) are control parameters, with a common parameter set being \( \sigma = 10 \), \( \rho = 28 \), and \( \beta = \frac{8}{3} \). Under these values, the Lorenz system exhibits complex nonlinear behavior and displays characteristic chaotic dynamics.

In this experiment, the training dataset consists of 149,850 noisy data points, capturing the full dynamics of the system. An additional 150 noise-free data 
points are used to validate the model and assess its predictive performance. This volume of data far exceeds the number of model parameters, ensuring that the online DMD algorithm avoids overfitting in this low-dimensional system.

\begin{figure}[!ht]
  \centering
  \begin{subfigure}{0.31\textwidth}
      \centering
      \includegraphics[width=\textwidth]{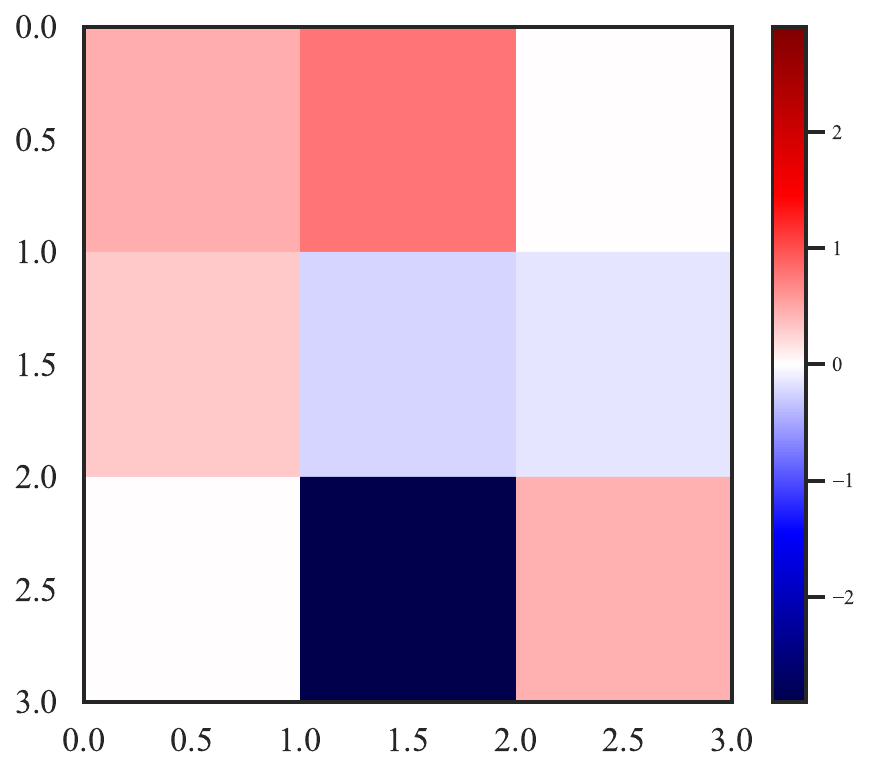}
      \caption{Matrix}
  \end{subfigure}
  \hspace{1em}
  \begin{subfigure}{0.285\textwidth}
      \centering
      \includegraphics[width=\textwidth]{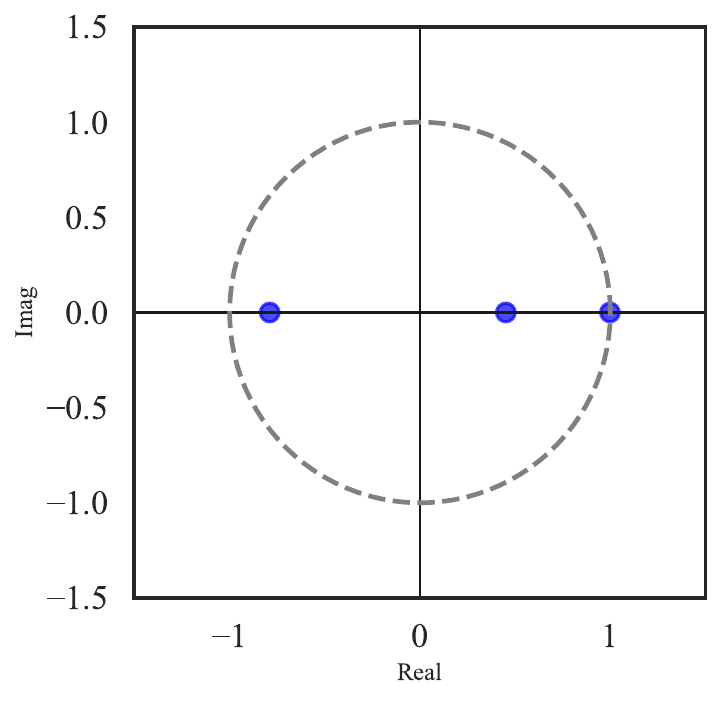}
      \caption{Eigenvalues}
  \end{subfigure}
  \caption{OPIDMD Matrix and Eigenvalue Visualizations}
  \label{fig:Lorenz3}
\end{figure}

\begin{figure}[!ht]
  \centering
  \begin{subfigure}{0.29\textwidth}
      \centering
      \includegraphics[width=\textwidth]{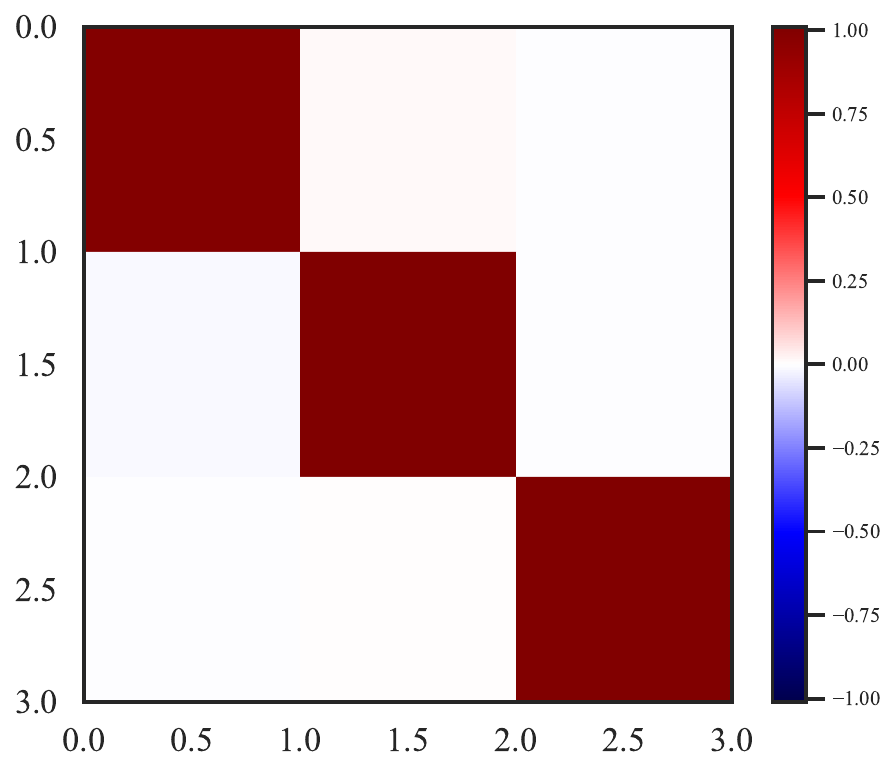}
      \caption{True DMD}
  \end{subfigure}
  \hspace{2em}
  \begin{subfigure}{0.29\textwidth}
      \centering
      \includegraphics[width=\textwidth]{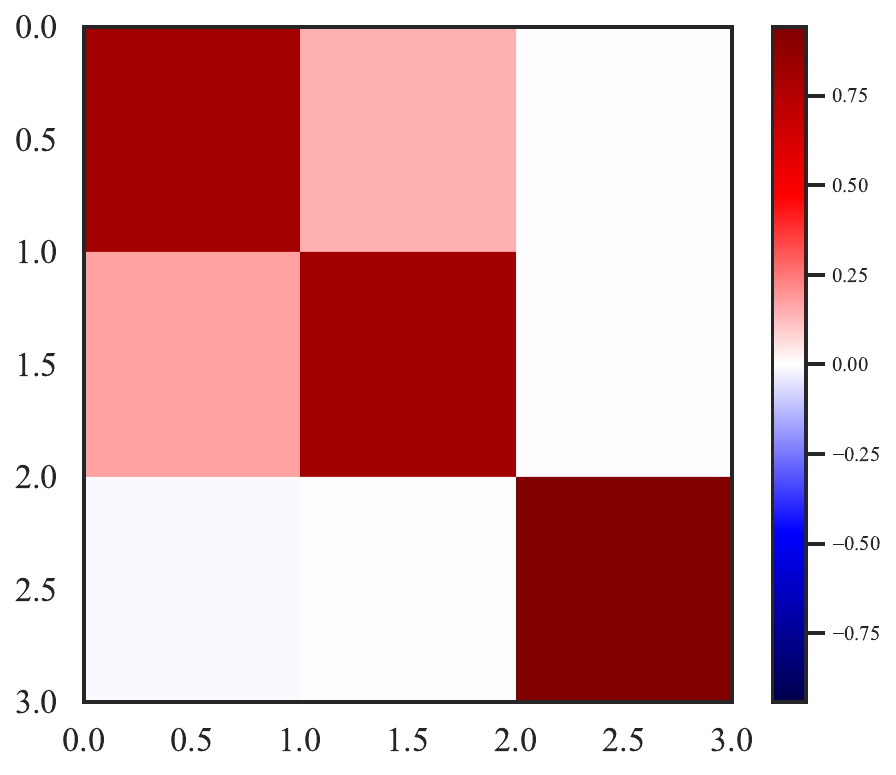}
      \caption{Online DMD}
  \end{subfigure}
  \hspace{2em}
  \begin{subfigure}{0.29\textwidth}
      \centering
      \includegraphics[width=\textwidth]{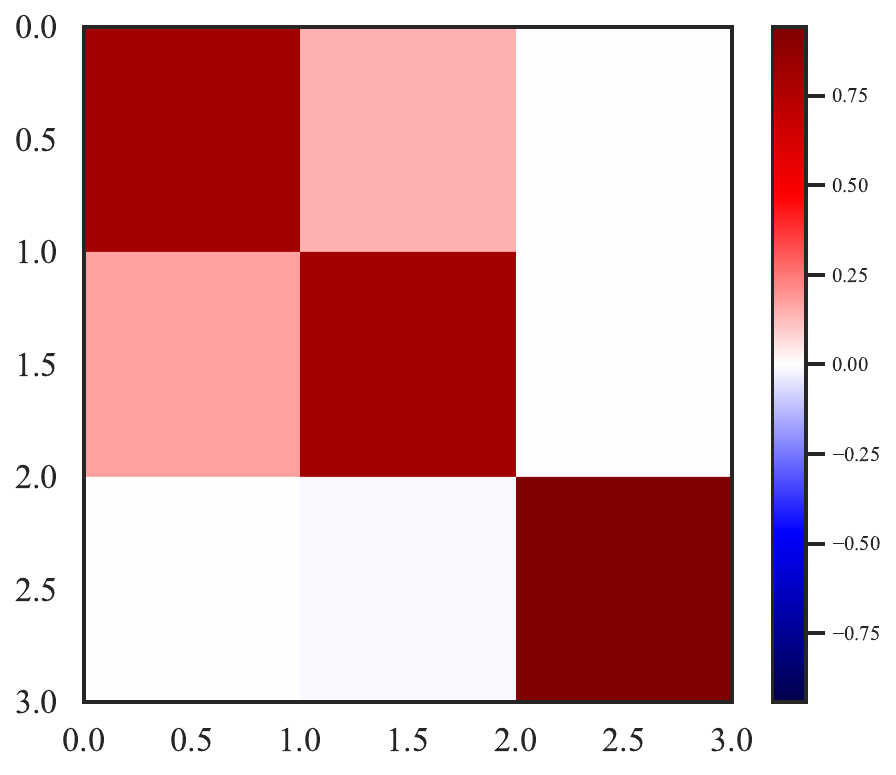}
      \caption{piDMD}
  \end{subfigure}

  \vspace{1em}

  \begin{subfigure}{0.29\textwidth}
      \centering
      \includegraphics[width=\textwidth]{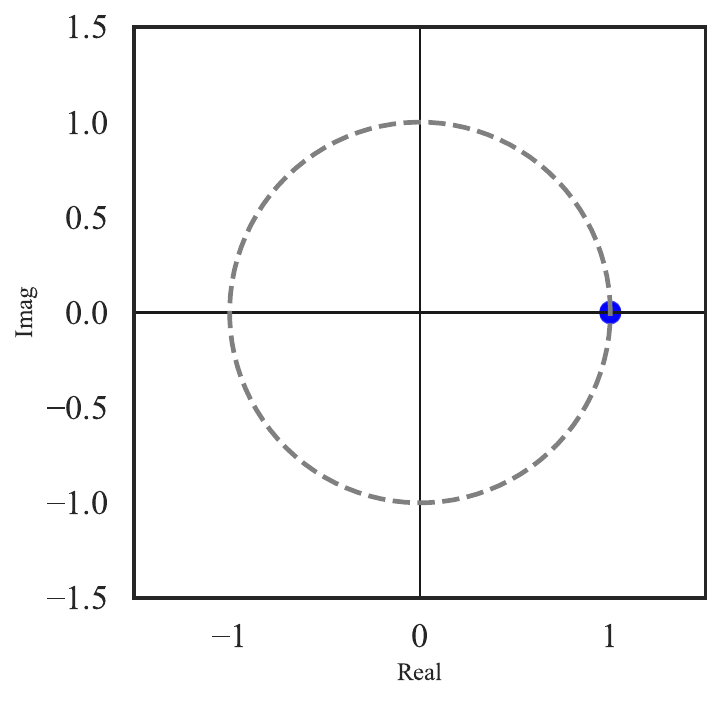}
      \caption{True DMD}
  \end{subfigure}
  \hspace{2em}
  \begin{subfigure}{0.29\textwidth}
      \centering
      \includegraphics[width=\textwidth]{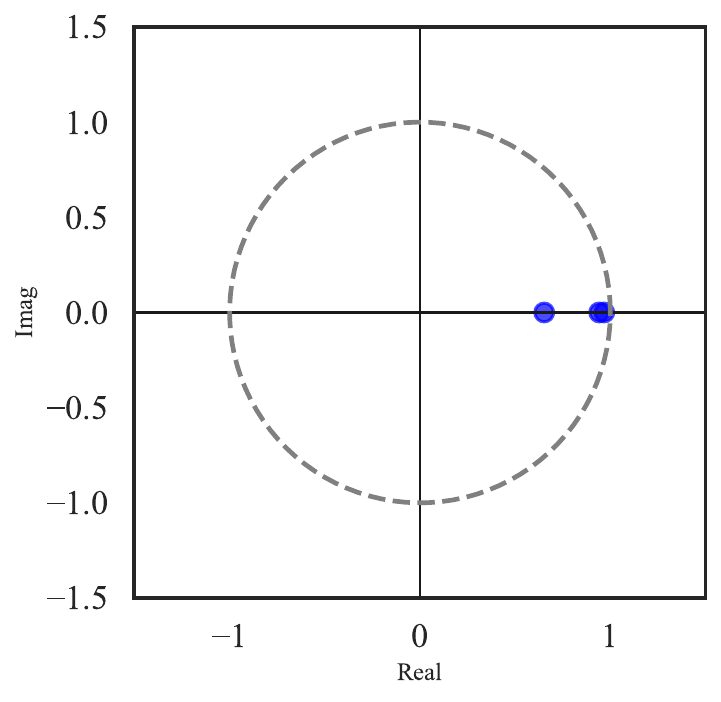}
      \caption{Online DMD}
  \end{subfigure}
  \hspace{2em}
  \begin{subfigure}{0.29\textwidth}
      \centering
      \includegraphics[width=\textwidth]{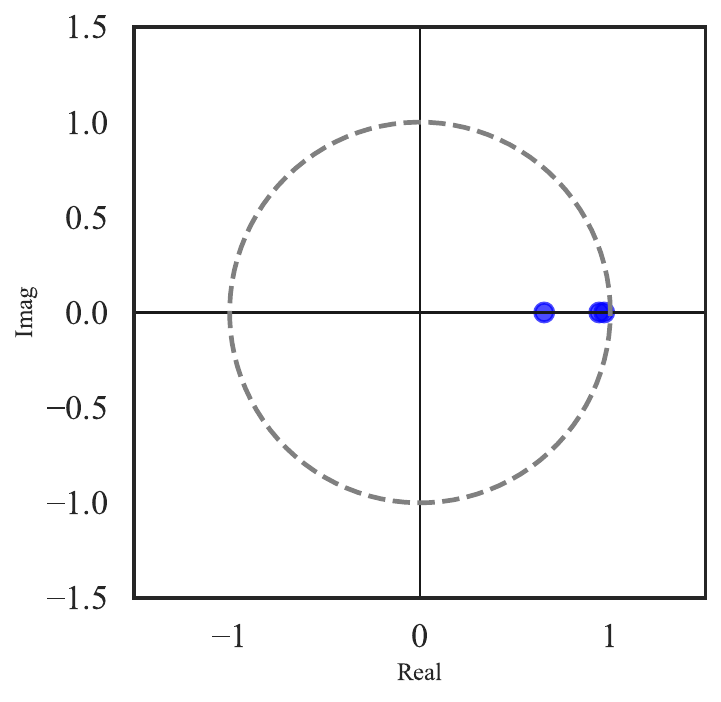}
      \caption{piDMD}
  \end{subfigure}
  \caption{Comparison of DMD Matrices and Spectral Analysis}
  \label{fig:Lorenz2}
\end{figure}

Due to the chaotic nature of the Lorenz system, its sensitivity to initial conditions implies that the current state is primarily influenced 
by recent states, reflecting a strong local dependency. This feature leads the True DMD matrix to exhibit a tridiagonal structure. Hence, we 
verify the physical validity of the model by imposing the tridiagonal matrix as a physical constraint. Figures \ref{fig:Lorenz3} and \ref{fig:Lorenz2} 
display this matrix structure and its eigenvalue visualizations.

Since chaotic systems lack a globally linear representation in three-dimensional space, algorithms such as online DMD and piDMD, which 
approximate the system globally as a linear model, often struggle to achieve satisfactory results. These algorithms tend to oversimplify 
the system’s dynamic behavior, making them susceptible to noise in high-noise environments and significantly lowering prediction accuracy.
 In contrast, OPIDMD demonstrates a distinct advantage, as its local linearization approach better accommodates the Lorenz system’s local 
 characteristics. It effectively separates noise during long-duration noisy training, achieving higher prediction accuracy.

\begin{figure}[!ht]
  \centering
  \begin{subfigure}{0.31\textwidth}
      \centering
      \includegraphics[width=\textwidth]{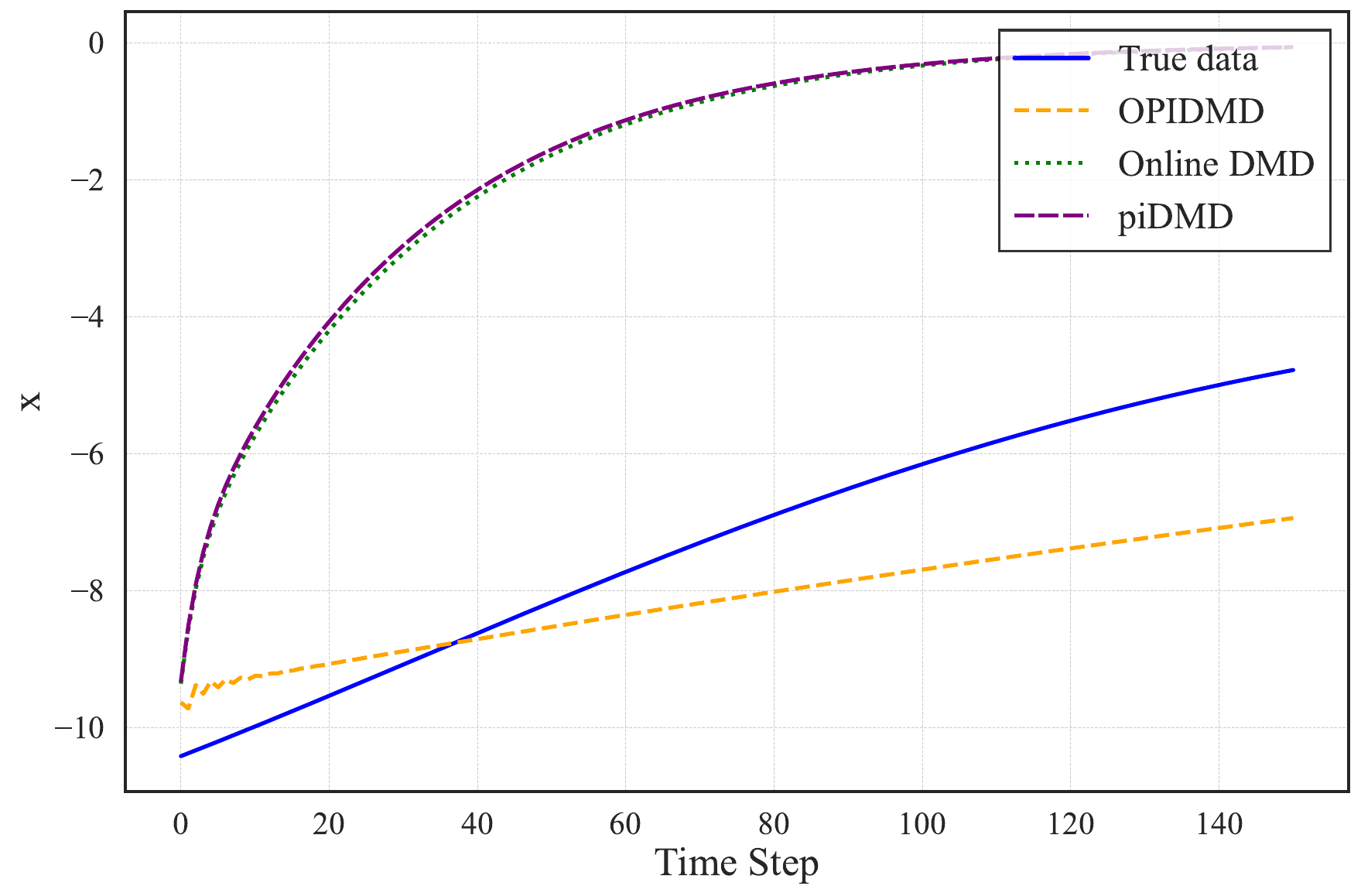}
      \caption{$x$}
  \end{subfigure}
  \hspace{0.5em}
  \begin{subfigure}{0.31\textwidth}
      \centering
      \includegraphics[width=\textwidth]{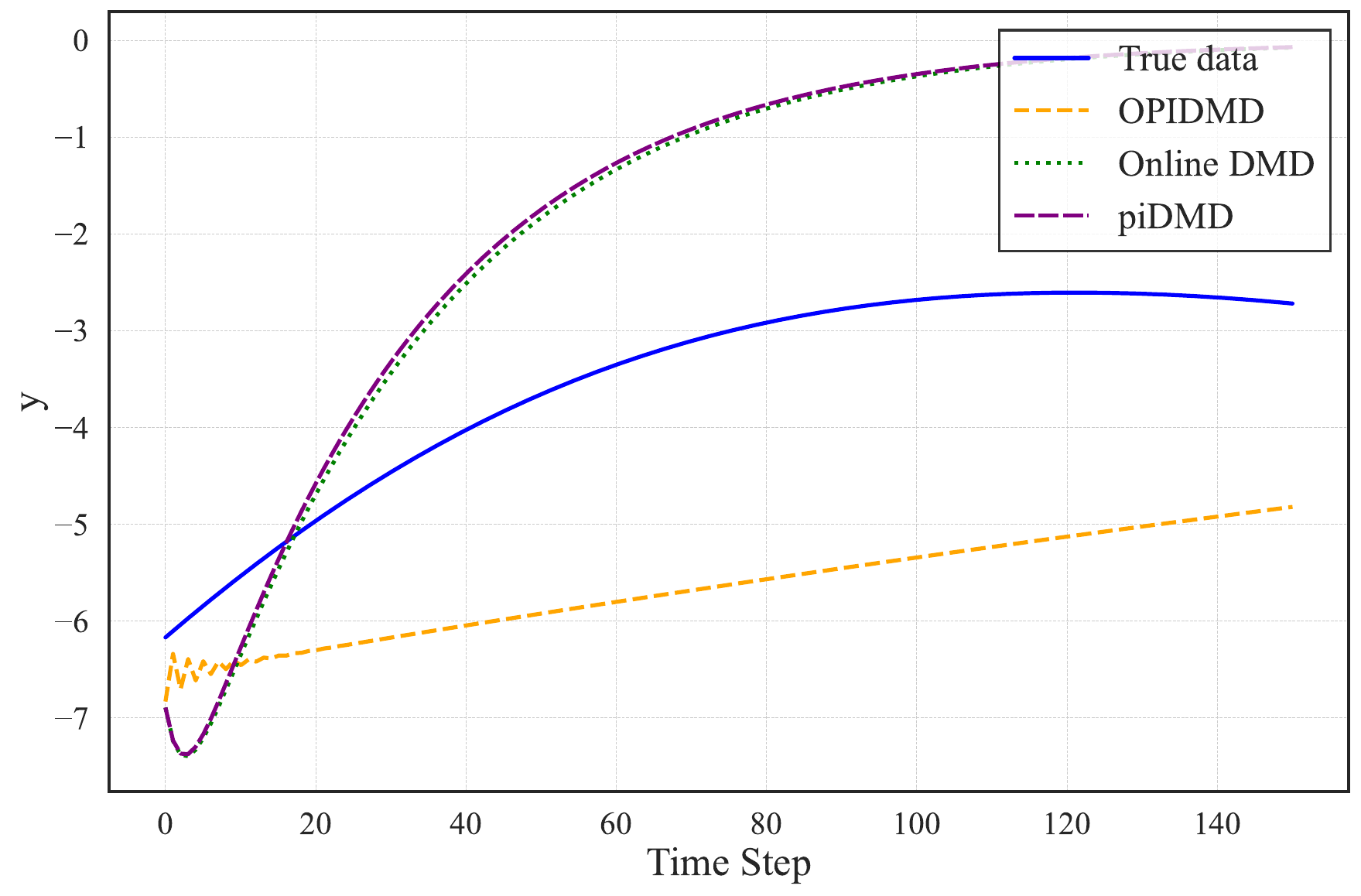}
      \caption{$y$}
  \end{subfigure}
  \hspace{0.5em}
  \begin{subfigure}{0.31\textwidth}
      \centering
      \includegraphics[width=\textwidth]{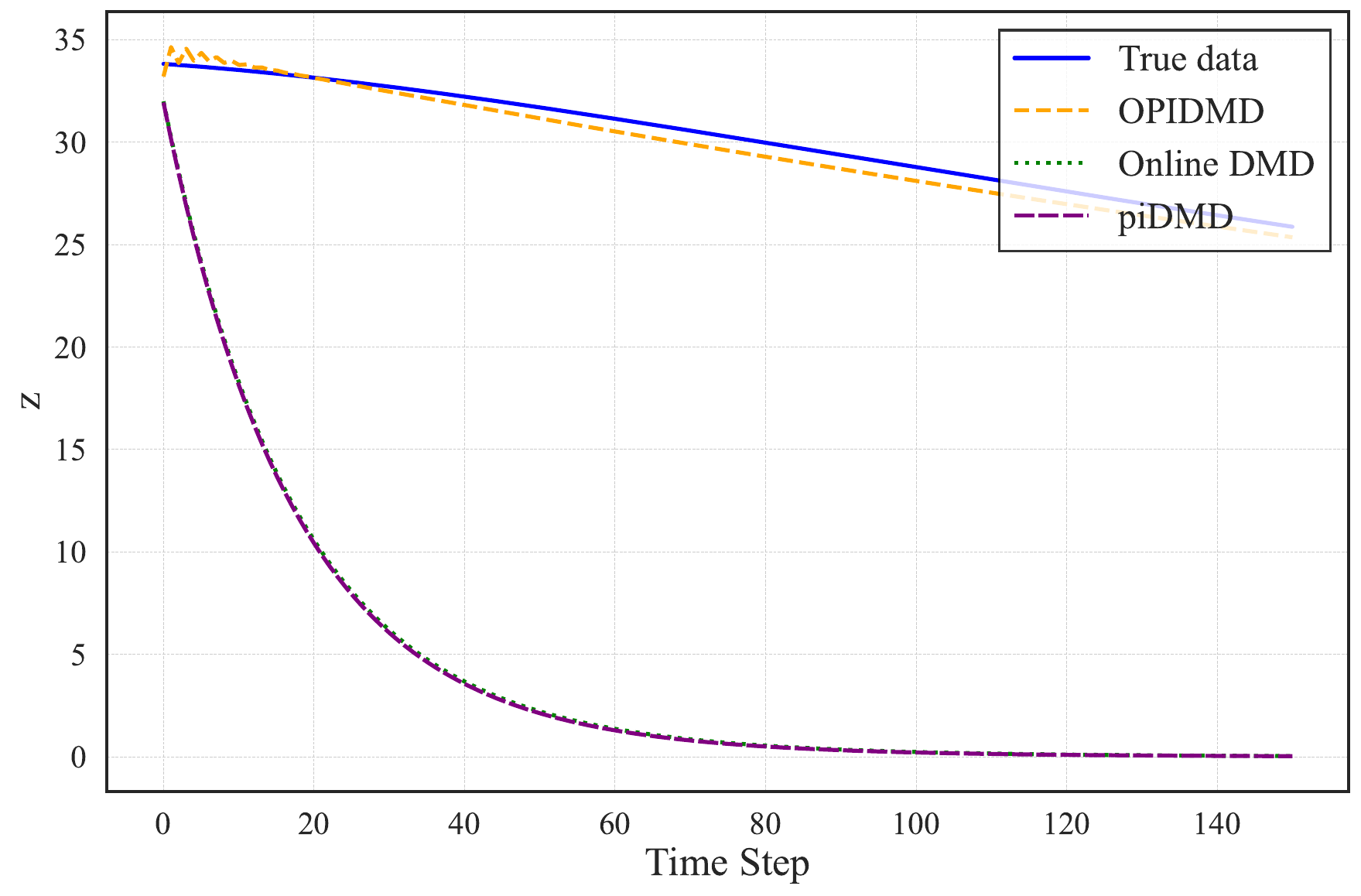}
      \caption{$z$}
  \end{subfigure}
  \caption{Comparison of Prediction Results in the Lorenz System}
  \label{fig:Lorenz1}
\end{figure}

Figure \ref{fig:Lorenz1} compares the prediction accuracy of different algorithms. Although the Lorenz system is challenging to
 predict with traditional linear methods, OPIDMD, with its localized linear modeling strategy, excels in short-term predictions. The \( R^2 \) values for OPIDMD, Online DMD, and piDMD were 0.991, 0.066, and 0.062, respectively.

\section{Conclusion}\label{sec:conclusion}
In this study, we introduce OPIDMD, a novel method developed for the online analysis of time-varying dynamical systems. We demonstrate that, 
for the constraints mentioned in this paper, under conditions satisfying Slater’s condition, the MAP estimation and piDMD are equivalent. To address the nonconvex physical constraints, we relax them and reframe them as penalties, allowing us to combine them with convex terms derived from the probabilistic modeling of the \(A\) parameter. Our analysis examines the impact of these constraints on the bias and variance of \(A\) under \(l_2\) norm constraints.

Additionally, we leverage online proximal gradient descent to analyze eight fundamental physical principles, such as norm shrinkage and sparsity, 
and conclude with relevant case studies. Our approach extends the capabilities of traditional DMD by allowing for the analysis of larger and more diverse datasets, extracting dominant spatiotemporal patterns efficiently. This enhanced method shows improved noise resistance due to the physical constraints in the optimization process. In all numerical examples, the  short-term prediction performance of OPIDMD reaches or approaches the optimal level of existing algorithms.

Notably, compared with existing algorithms, OPIDMD demonstrates the best prediction performance in short-term forecasting for noisy Lorenz system numerical examples with chaotic characteristics, achieving an \( R^2 \) value of 0.991.
We also found that online gradient descent has an implicit constraint effect, often leading to a simpler and smoother solution. As a result, OPIDMD maintains strong generalization performance even in high-noise, small-sample scenarios. However, for systems with an extremely high number of state variables, storing a large system matrix may exceed the capacity of conventional computers. Therefore, further research is needed to explore feasible approaches for running online algorithms in such cases.

  This research opens up new avenues and presents challenges
  for data-driven dynamical systems.  A promising direction 
 for future research involves a more thorough exploration of the OPIDMD algorithms' application in system identification, particularly 
 for scenarios involving dynamic variations or the need for real-time control. Accurate and adaptive reduced-order models are crucial for effective control in these contexts, and the methods proposed herein hold significant potential for addressing these requirements.

\backmatter

\bmhead{Supplementary information}

This supplementary material includes the following sections:
\begin{itemize}
  \setlength{\itemsep}{2mm} 
  \item \textbf{Convex Optimization and Proximal Gradient Descent}: This section provides an overview of the convex optimization concepts and algorithms relevant to the paper, including a brief introduction to implicit constraints.
  \item \textbf{Numerical Examples}: Detailed explanations of additional results that were not included in the main text are presented in this section.
  \item \textbf{Statistical Analysis of DMD}: This section explores how probabilistic assumptions are used to conduct a bias-variance analysis of the DMD algorithm. It concludes by reinterpreting the constraints in Exact DMD, piDMD, and OPIDMD from this perspective.
  \item \textbf{Computational Time}: A comparison of the computational efficiency of OPIDMD with existing algorithms is provided.
\end{itemize}

\bmhead{Acknowledgements}

This research has been supported by the China National Key R\&D Program (2022YFB2602103), and National Natural Science Foundation of 
China (General Program, 52378294). This research compared various algorithms, including online DMD, piDMD, and streaming DMD, by referencing
 open-source libraries such as PyDMD and 
online DMD codebases \cite{demo2018pydmd, ichinaga2024pydmd, zhangOnlineDynamicMode2019, hematiDynamicModeDecomposition2014, baddooPhysicsinformedDynamicMode2023}. 
We are grateful for these open-source contributions.

\begin{appendices}

\section{Convex Optimization and Proximal Gradient Descent}\label{convex_opi}
The primary advantage of convex optimization in practical physical problems often lies in its ability to guarantee global optima, which is
crucial for dynamical system analysis to avoid being trapped in local optima\cite{boyd2004convex}. 
The simplicity and clarity of convex problems facilitate direct and efficient solutions,
enhancing both efficiency and performance of dynamical system modeling. 
However, not all DMD 
problems naturally lead to convex optimization, which poses a limitation given the inherent complexity of many dynamical systems. 

This paper has discussed
 the convexity of common piDMD problems and utilized convex relaxation techniques to convert certain non-convex challenges into convex problems. Convex relaxation involves replacing
 the original problem's constraints with convex functions, which serve as underestimators of the original functions, ensuring that the relaxed problem stays within feasible bounds.
A convex optimization problem  is:
\begin{equation}
\begin{aligned}
& \underset{x}{\text{minimize}}
& & f(x) \\
& \text{subject to}
& & h_i(x) \leq 0, \; i = 1, \ldots, m \\
& & & \ell_j(x) = 0, \; j = 1, \ldots, r
\end{aligned}
\end{equation}
where $f(x)$ and $h_i(x), i = 1, \ldots, m$ are all convex, and $\ell_j(x), j = 1, \ldots, r$ are affine functions.
Convex optimization offers a simpler and more predictable solution path compared to nonconvex optimization, 
where the presence of multiple local minima complicates the solution process\cite{rockafellar2015convex}.
The Lagrange dual function, constructed from the Lagrangian, plays a pivotal role in duality theory. The Lagrangian for a given optimization problem is defined as:
\begin{equation}
L(x, u, v) = f(x) + \sum_{i=1}^{m} {u_i}'  h_i(x) + \sum_{j=1}^{r} {v_j}' \ell_j(x)
\end{equation}
where \( {u_i}'  \geq 0 \) for all \( i \), reflecting the non-negativity constraint on the dual variables associated with the inequality constraints, and \( {v_j}' \in \mathbb{R} \) for all \( j \), indicating that the dual variables corresponding to equality constraints can take any real value.
The dual problem, then, is to maximize the dual function subject to $u \geq 0$:
\begin{equation}
\begin{aligned}
& \underset{u, v}{\text{maximize}}
& & g(u, v) \\
& \text{subject to}
& & u \geq 0
\end{aligned}
\end{equation}
where $g(u, v) = \min_{x} L(x, u, v)$. It is crucial to note that the dual problem is always convex, meaning that $g$ is always concave, even 
if the primal problem is not convex.
This framework of convex optimization and duality lays the groundwork for advanced solution techniques, which will be explored next. 

\subsection{Constrained and Lagrange Forms}
The main purpose of this section is to prove that the Lagrange forms in Table 3 of the main text can be equivalent to 
the constrained form under certain conditions. For the physical constraints mentioned in Table 3, it is evident that for 
the first five types, due to the presence of the indicator function \( I_C(x) \), we can readily observe that
\begin{equation}
  \underset{A}{\mathrm{argmin}} \|Y - AX  \|_F^2 + I_C(\mathcal{M}) = \underset{A \in \mathcal{M}}{\mathrm{argmin}} \left\|Y - AX\right\|_F^2  
\end{equation}
Here, \( \mathcal{M} \) refers to the first five constraints in Table 3. For the remaining types, further analysis is required. 
This paper first proves the general case and then specifically addresses the situations listed in Table 3. In the constrained form, suppose our objective is to minimize the function \( \|Y - AX\|_F^2 \), where \( t \in \mathbb{R} \) serves as a tuning parameter to control the constraint applied on \( R(A) \):
\begin{equation}\label{eqc_1}
  \min_A \|Y - AX\|_F^2 \quad \text{subject to} \quad R(A) \leq t 
\end{equation}
The Lagrange form introduces a Lagrange multiplier \( \lambda \geq 0 \), which also acts as a tuning parameter. 
\begin{equation}\label{eqc_2}
  \min_A \|Y - AX\|_F^2 + \lambda \cdot R(A)
\end{equation}
Here, we assume that \( R(A) \) is a convex function. The goal of our 
investigation is to examine the conditions under which this equivalence holds.
If the constrained problem \eqref{eqc_1} is strictly feasible—in other words, if it satisfies Slater’s condition—then strong duality holds. 
This implies the existence of some \( \lambda \geq 0 \) (a dual solution) such that any solution \( A^* \) in \eqref{eqc_1} also minimizes the Lagrange form:
\begin{equation}
  \|Y - AX\|_F^2 + \lambda \cdot (R(A) - t)
\end{equation}
Therefore, \( A^* \) is also a solution to \eqref{eqc_2}. Conversely, if \( A^* \) is a solution in \eqref{eqc_2},
 then by setting \( t = R(A^*) \), the Karush-Kuhn-Tucker (KKT) conditions for \eqref{eqc_1} are satisfied. This implies
  that \( A^* \) is also a solution to \eqref{eqc_1}.
In summary, we can derive the following relationships between the solution sets of \eqref{eqc_2} and \eqref{eqc_1}:
\begin{equation}
  \bigcup_{\lambda \geq 0} \{ \text{solutions in \eqref{eqc_2}} \} \subseteq \bigcup_{t} \{ \text{solutions in \eqref{eqc_1}} \}
\end{equation}
\begin{equation}
  \bigcup_{\lambda \geq 0} \{ \text{solutions in \eqref{eqc_2}} \} \supseteq \bigcup_{t \text{ such that \eqref{eqc_1}  is strictly feasible}} \{ \text{solutions in \eqref{eqc_1}} \}
\end{equation}
This result is nearly a perfect equivalence. Notably, when the only value of \( t \) that leads to a feasible but not strictly 
feasible constraint set is \( t = 0 \), then we do get perfect equivalence.
If \( R(A) \geq 0 \) and both \eqref{eqc_1} and \eqref{eqc_2} are feasible for all \( t, \lambda \geq 0 \), then we achieve perfect equivalence 
across the solution sets. 

In Table 3 of the main text, the \( R(A) \) functions we use are all some form of norm applied to \( A \), so they satisfy \( R(A) \geq 0 \). Additionally,
 we have specified \( \lambda \geq 0 \) and \( t > 0 \), which ensures that perfect equivalence can be obtained.

\subsection{Proximal Gradient Descent}
The convex optimization problems discussed in this paper can be concisely expressed by the following equation:
 \begin{equation}
 \min_{x} f(x) = g(x) + h(x),
 \end{equation}
 where $g$ represents a convex and differentiable function, and $h$ denotes a convex but potentially nondifferentiable function. 
The proximal gradient method is specifically designed for optimization challenges of this nature, proving especially advantageous for 
 managing the nonsmooth regularization terms that are prevalent in fields such as machine learning and signal processing\cite{combettes2005signal}.
 The specific steps can be seen in Algorithm \ref{algo:proximal_gradient}.
\begin{algorithm}
  \caption{Proximal Gradient Method}\label{algo:proximal_gradient}
  \begin{algorithmic}
  \State \textbf{Input}: Initial point $x^{(0)}$, step sizes $\{t_k\}$ chosen to be fixed and small, or determined through backtracking line search as
   described in Algorithm \ref{algo:Backtracking}.
  \State \textbf{Goal}: Minimize $f(x) = g(x) + h(x)$, where $g$ is convex and differentiable, and $h$ is convex but possibly nondifferentiable.
  
  \For{$k = 0, 1, 2, \ldots$ until convergence}
      \State Compute the gradient of $g$ at $x^{(k)}$:$$\nabla g(x^{(k)})$$
      \State Update $x^{(k+1)}$ using the proximal gradient step:
      \State \quad $$x^{(k+1)} = \text{prox}_{t_kh}\left(x^{(k)} - t_k \nabla g(x^{(k)})\right)=x^{(k)}-t_kG_t(x^{(k)})$$
      
      \State \textbf{where} the proximal mapping $\text{prox}_{t}(v)$ and the generalized gradient $G_t(x)$ are defined as:
      \State \quad $$\text{prox}_{th}(x) = \arg\min_{z} \left\{ \frac{1}{2t}\|x - z\|^2 + h(z) \right\}, \quad G_{th}(x) = \frac{x - \text{prox}_{th}(x - t \nabla g(x))}{t}$$
  \EndFor
  
  \State \textbf{return}: The sequence $\{x^{(k)}\}$ converging to a solution of $\min_{x} f(x) = g(x) + h(x)$.
  \end{algorithmic}
\end{algorithm}

This proximal gradient method is particularly powerful when the proximal operator for $h$ can be computed efficiently, which is the 
case for many important functions in optimization, allowing for the handling of complex problems involving sparsity, 
constraints, and regularization in a computationally efficient manner\cite{combettes2005signal,combettes2011proximal}.

Backtracking line search is a common technique in optimization algorithms, especially used in scenarios like gradient descent or Newton’s method for
finding the minimum value of a function\cite{civicioglu2013backtracking}. The advantages of backtracking line search include its simplicity and
flexibility, allowing for the adaptive adjustment of step size to suit the local
characteristics of the function. The specific steps can be seen in Algorithm \ref{algo:Backtracking}.
\begin{algorithm}
  \caption{Backtracking Line Search for Proximal Gradient Method}\label{algo:Backtracking}
  \begin{algorithmic}
  \State \textbf{Input}: Current iterate $x$, initial step size $t_{\text{init}}$, shrinkage parameter $\beta$ where $0<\beta<1$.
  \State Initialize step size $t = t_{\text{init}}$.
  \State Compute the generalized gradient $G_{th}(x)$ 
  \While{
  $g(x - t G_{th}(x)) > g(x) - t \nabla g(x)^T G_{th}(x) + \frac{t}{2} \|G_{th}(x)\|_2^2$}
      \State Shrink the step size: $t = \beta t$.
      \State Update the generalized gradient $G_{th}(x)$ for the new $t$.
  \EndWhile
  \State \textbf{return} step size $t$.
  \end{algorithmic}
\end{algorithm}

The convergence accuracy to optimal solutions in convex optimization problems, where the 
objective function $f(x) = g(x) + h(x)$ combines a $L$-Lipschitz continuous differentiable function $g(x)$\cite{schmidt2011convergence,suzuki2013dual}. The choice of an appropriate step size is critical, as it ensures the iterates $x^{(k)}$ 
converge to the optimal solution $x^*$ at a rate of $O(1/\sqrt{k})$, where $k$ is the iteration count. This setting provides a theoretical 
convergence rate, yet the practical computational complexity of executing proximal gradient descent varies with different forms of $h(x)$, 
impacting the convergence time.

The choice of step size $t_k$ can be constant or determined through backtracking line
search techniques to ensure convergence. 
This advantage is further augmented by implementing variations of the gradient descent 
algorithm designed to expedite convergence, such as Variance Reduction\cite{johnson2013accelerating,schmidt2017minimizing,defazio2014saga}, 
Acceleration and Momentum\cite{nitanda2014stochastic,jin2018accelerated,qian1999momentum,liu2020accelerating}, 
and Adaptive Step Sizes\cite{ward2020adagrad,duchi2011adaptive,kingma2014adam,zou2019sufficient}. These methodologies
can significantly enhance the efficiency and responsiveness of the algorithm to changes in data over time.

The proximal gradient descent method, also referred to as the generalized gradient descent, is adaptable to a broad spectrum of 
optimization problems. It operates on the principle of minimizing a composite function $f = g + h$, where\cite{combettes2005signal}:
\begin{itemize}
  \item $h = 0$, the method reverts to the classic gradient descent approach, where it purely follows the negative gradient of the function $g$.
  \item $h = I_C$, proximal gradient descent method transforms into projected gradient descent. This is achieved by projecting the gradient steps
onto $C$, thereby maintaining the feasibility of the iterates with respect to the constraints defined by $C$.
\end{itemize}

The formulation given in equation \eqref{eqc_{27}} presents a minimization problem that aims to balance two objectives: to minimize the 
distance to \(x\) and to ensure that \(z\) remains within the set \(C\)\cite{combettes2005signal}. 
The second line streamlines this problem into finding the point within \(C\) that is closest to \(x\), effectively turning it into a projection operation.
\begin{align}\label{eqc_{27}}
\operatorname{prox}_{t,I_C}(x) & =\underset{z}{\operatorname{argmin}} \frac{1}{2 t}\|x-z\|_2^2+I_C(z) \\
& =\underset{z \in C}{\operatorname{argmin}}\|x-z\|_2^2
\end{align}
Therefore the proximal gradient update step is\cite{combettes2005signal}:
\begin{equation}
  x^{+}=P_C(x-t \cdot \nabla g(x))
\end{equation}
The equation \(\operatorname{prox}_{t,I_C}(x) = P_C(x)\) indicates that the proximal operator functions as a projection operator \(P_C(x)\) onto the set \(C\). 
This process identifies the point \(z\) in \(C\) closest to \(x\), minimizing their distance and reducing the value of \(g(x)\) while ensuring compliance 
with the constraints represented by \(C\). This is crucial in constrained optimization, where direct gradient descent might exit the feasible region \(C\). 
Projecting back onto \(C\) maintains adherence to constraints at every step, effectively integrating gradient optimization with constraint satisfaction. 

Moreover, the inherent nature of convex problems guarantees that, with 
the sequential accumulation of online data, algorithms are assured to converge to this unique global optimum. This ensures a consistent and 
efficient approach in solving for $A_k$.
In contrast, non-convex problems lack this assurance of converging to a global optimum, often necessitating more sophisticated and 
computationally demanding strategies to find satisfactory solutions\cite{rockafellar2015convex}.

\subsection{OPIDMD Tailored to Specific Physical Principles}
This section will discuss the specific implementation of the proximal mapping $\text{prox}_{th}(x)$ for the constraints mentioned in Table 2 of the main text.  First, the algorithms for the following types: Circulant matrix, Self-Adjoint matrix, Upper Triangular matrix, and Tri-Diagonal matrix, all belong to the projection type in an online context. The implementation of projected gradient descent, as previously introduced, ensures adherence to physical constraints, thus guaranteeing that $A_k$ belongs to the set $\mathcal{M}$. The specific proximal mapping $\text{prox}_{th}(x)$ can be found in Algorithms \ref{algo:projection_circulant}, \ref{algo:projection_symmetric}, \ref{algo:projection_upper_triangular}, and \ref{algo:projection_tridiagonal}.

\begin{algorithm}
  \caption{Projection to Circulant Matrix}\label{algo:projection_circulant}
  \begin{algorithmic}
  \State \textbf{Input}: Matrix $M$ of dimension $n \times n$.
  \State \textbf{Output}: Circulant matrix $C$ that is closest to $M$.
  \State Initialize an empty vector $c$ of length $n$.
  \For{$k = 0$ to $n-1$}
    \State Compute the mean of elements on the $k$-th circulant diagonal of $M$: 
    \[
    c_k  = \frac{1}{n} \sum_{i=0}^{n-1} M[i, (i + k) \bmod n]
    \]
  \EndFor
  \State Construct circulant matrix $C$ using $c = [c_0, c_1, \dots, c_{n-1}]$ as the first row. 
  \State \textbf{return} $C$
  \end{algorithmic}
\end{algorithm}

\begin{algorithm}
  \caption{Projection to Symmetric Matrix}\label{algo:projection_symmetric}
  \begin{algorithmic}
  \State \textbf{Input}: Matrix $A$ of dimension $n \times n$.
  \State \textbf{Output}: Symmetric matrix $S$ based on $A$.
  \State $S = (A + A^T) / 2$
  \State \textbf{return} $S$
  \end{algorithmic}
\end{algorithm}

\begin{algorithm}
  \caption{Projection to Upper Triangular Matrix}\label{algo:projection_upper_triangular}
  \begin{algorithmic}
  \State \textbf{Input}: Matrix $A$ of dimension $n \times n$.
  \State \textbf{Output}: Upper triangular matrix  based on $A$.
  \For{$i = 1$ to $n$}
    \For{$j = 1$ to $n$}
      \If{$i > j$}
        \State $a[i, j] = 0$
      \EndIf
    \EndFor
  \EndFor
  \State \textbf{return} $A=a[i, j]$
  \end{algorithmic}
\end{algorithm}

\begin{algorithm}
  \caption{Projection to Tridiagonal Matrix}\label{algo:projection_tridiagonal}
  \begin{algorithmic}
  \State \textbf{Input}: Matrix $A$ of dimension $n \times n$.
  \State \textbf{Output}: Tridiagonal matrix based on $A$.
  \For{$i = 1$ to $n$}
    \For{$j = 1$ to $n$}
      \If{$\lvert i - j \rvert > 1$}
        \State $a[i, j] = 0$
      \EndIf
    \EndFor
  \EndFor
  \State \textbf{return} $A=a[i, j]$
  \end{algorithmic}
\end{algorithm}

Using a trace norm regularization method is in alignment with the foundational goals of exact DMD, which emphasizes 
identifying low-rank approximations as a means to achieve succinct and interpretable representations of dynamical systems.
The specific proximal mapping $\text{prox}_{th}(x)$ can be found in Algorithm \ref{algo:unclear Norm Constraint}.
The significance of this approach extends beyond mere approximation, highlighting its capacity for real-time generation of 
low-rank solutions.

\begin{algorithm}
    \caption{Proximal Operator for unclear Norm Constraint\cite{mazumder2010spectral}}\label{algo:unclear Norm Constraint}
    \label{algo:proximal_unclear_norm}
    \begin{algorithmic}
    \State \textbf{Input}: Matrix $A \in \mathbb{R}^{n \times n}$, regularization parameter $\lambda > 0$
    \State \textbf{Output}: Matrix $B$ after applying unclear norm  soft thresholding on $A$ 
    \State \textbf{Step 1:} Perform SVD of $A$:
    \State \quad $U, \Sigma, V^T = \textbf{SVD}(A)$, where $\Sigma$ is a diagonal matrix of singular values
    \State \textbf{Step 2:} Apply the unclear norm constraint to all singular values $\Sigma$:
    \State \quad $\Sigma' = \max(\Sigma - \lambda I, 0)$
    \State \textbf{Step 3:} Reconstruct the projected matrix $B$:
    \State \quad $B = U \Sigma' V^T$
    \State \textbf{return} $B$
    \end{algorithmic}
  \end{algorithm}
  
   By leveraging the trace norm constraint, we can dynamically adjust 
  the complexity of the model in response to new data, ensuring that the resulting solutions are not only accurate and tailored 
  to the current dataset but also inherently low-rank. This aspect of real-time adaptability is crucial for applications requiring 
  continuous monitoring and updating, such as in streaming data scenarios or in systems where conditions may rapidly change.

  The sparsity induced by utilizing the $l_1$ norm is not merely a mathematical convenience but mirrors the intrinsic properties of many physical systems, 
  which often feature inherently sparse matrices that represent system dynamics. This sparsity enhances model interpretability, providing 
  clear insights into critical relationships and interactions within the system. The corresponding proximal mapping $\text{prox}_{th}(x)$ is
   detailed in Algorithm \ref{algo:l1 Norm Constraint}.

   \begin{algorithm}
    \caption{ Proximal Operator for $l_1$ Norm Constraint\cite{beck2009fast}}\label{algo:l1 Norm Constraint}
    \label{algo:l1_soft_thresholding}
    \begin{algorithmic}
    \State \textbf{Input}: Matrix $A \in \mathbb{R}^{n \times m}$, threshold $\lambda > 0$
    \State \textbf{Output}: Matrix $B$ after applying $l_1$ soft thresholding on $A$
    \State \textbf{for} each element $a_{ij}$ in $A$ \textbf{do}
    \State \quad $b_{ij} = \text{sign}(a_{ij}) \times \max(\lvert a_{ij} \rvert - \lambda, 0)$
    \State \textbf{end for}
    \State \textbf{return} $B = [b_{ij}]$, the matrix after $l_1$ soft thresholding
    \end{algorithmic}
  \end{algorithm}
  
   Moreover, a sparse representation
     is inherently more resistant to noise, a common challenge in the analysis of physical systems. This attribute makes the $l_1$ norm 
     constraint particularly potent for identifying the essential sparse structure of physical systems' matrices, thereby enabling a more 
     accurate and noise-resistant characterization of the dynamics.
  
   The $l_2$ norm constraint serves as a powerful regularization technique that mitigates the impact of noise and avoids overfitting
     by ensuring the model does not adhere too closely to noisy training data. The reduction in the Frobenius norm indicates that the matrix
     $A$ is becoming "smoother," with its elements being constrained in a manner that promotes generalization over memorization of the training data.
     The specific proximal mapping $\text{prox}_{th}(x)$ can be found in Algorithm \ref{algo:l2 Norm Constraint}.
  
  \begin{algorithm}
    \caption{Gradient Descent Method for $l_2$ Norm Constraint}\label{algo:l2 Norm Constraint}
    \begin{algorithmic}
    \State \textbf{Input}: Initial point $A_0$, step sizes $\{t_k\}$ chosen to be fixed and small, or via backtracking line search.
    \State \textbf{Goal}:  minimize 
    $$
    f(A) = g(A) + h(A)= \sum_{i=1}^{k} \left\| y_{i} - A x_i \right\|^2_2 + \lambda \left \| A_k \right \|_F^2
    $$ over time, with data arriving online.
    \For{each iteration $k = 1, 2, \ldots$ with new data point $(x_{\text{new}}, y_{\text{new}})$}
        \State Update the objective function $$
        f(A_k)=\left\| y_{i} - A_k x_i \right\|^2_2 + \lambda \left \| A_k \right \|_F^2$$
        based on new data $(x_{\text{new}}, y_{\text{new}})$.
        \State Compute the gradient of updated $f_{\text{new}}(A_k)=\left\| y_{\text{new}} - A_k x_{\text{new}} \right\|^2_2+ \lambda \left \| A_k \right \|_F^2$ at $A_k$: 
        $$
        \nabla f_{\text{new}}(A_k)=-2\times \left ( y_{\text{new}}-A_kx_{\text{new}} \right ) x_{\text{new}}^T + 2 \lambda A_k
        $$
      \State Update $A_k$ using the gradient descent step:
        $$A_{k+1} = A_k - t \nabla f_{\text{new}}(A_k)$$
    \EndFor
    \State \textbf{return}: A sequence of solutions $A_k$ adapting to the new data over time.
    \end{algorithmic}
  \end{algorithm}

  \subsection{Implicit Constraints in Stochastic and Online Gradient Descent}
  In recent years, researchers have observed the "double descent" phenomenon in machine learning models\cite{belkin2020two,hastie2022surprises,belkin2019reconciling}. This phenomenon refers to a "double peak" shape in the training and test error curves as model parameters increase (e.g., by adding layers or neurons), which contrasts with traditional bias-variance analysis. Although the double
   descent phenomenon has not been fully explained theoretically, some studies suggest that the implicit regularization effect in Stochastic gradient descent (SGD)
   may be one of its main causes\cite{woodworth2020kernel,haochen2021shape}.
  
   SGD  is a classical optimization algorithm that minimizes a loss function by iteratively moving in the direction
   of the negative gradient, aiming to find a solution that achieves a local or global minimum of the loss function. Implicit regularization refers to the tendency of SGD to find "simpler" or "smoother" solutions even in the absence of explicit 
  regularization terms in the objective function, a property that is especially prominent in over-parameterized models\cite{zou2021benefits}.

  In neural networks and
    high-dimensional models, the parameter space is typically vast, allowing for multiple solutions that can reduce the training error to near
     zero, corresponding to different parameter combinations. Among all possible solutions, the ones with minimal norm are often of particular 
     interest, as they tend to have better generalization or uniqueness properties. 
   The role of 
  implicit regularization in the double descent phenomenon can be understood as follows:
  
  \textbf{1. Preference for Smooth Solutions}: In high-dimensional parameter spaces, SGD tends to find solutions of low complexity, 
  that is, solutions with smaller norms or simpler structures. Such solutions can achieve good generalization even when the number 
  of model parameters far exceeds the amount of data. This preference for low-complexity solutions is one of the reasons for the decrease
   in test error in the over-parameterized region.
  
  \textbf{2. Stochasticity of Gradient Updates}: Each iteration in SGD uses only a mini-batch of data, introducing randomness to the 
  solution, which helps to avoid convergence to local minima. This randomness enables the model to escape overfitting regions, leading 
  to solutions with better generalization and imparting a certain level of resistance to overfitting.
  
  \textbf{3. Synergy between Over-parameterization and Implicit Regularization}: In over-parameterized neural networks, where the number 
  of parameters greatly exceeds the amount of training data, the model has sufficient flexibility to choose among multiple solutions. Implicit
   regularization guides the model to favor simpler solutions with smaller norms or smoother surfaces, thus achieving good generalization performance.
  
  During training, the optimization path of SGD naturally tends to avoid regions of high complexity or oscillation, which prevents the model
   from converging to complex solutions and instead makes it more likely to find smooth, shallow minima. Additionally, even without explicit
    physical constraints, SGD exhibits some resistance to noise. Although the final solution from SGD may not always be the minimum-norm 
    solution\cite{sekhari2021sgd}, its stochastic update process tends to find solutions with good generalization properties, i.e., solutions
     that do not overfit 
    the training data.
  
  Unlike SGD, Online Gradient Descent (OGD) updates model parameters in a sequential manner, with each update relying on the current data point and the
   model state from the previous time step. This makes OGD particularly well-suited for time-series models, as it can effectively adapt to continuously 
   changing data and capture short-term variations. By processing data in a sequential manner, OGD reduces randomness-induced perturbations during training. 
   Theoretically, this should result in a weaker implicit regularization effect compared to SGD.
     
  However, in practice, we observed that the OGD algorithm also exhibits an "implicit regularization" effect, tending to learn simpler solutions. 
  This property provides it with greater robustness when dealing with high-dimensional, small-sample datasets. Even in scenarios where the number 
  of model parameters significantly exceeds the amount of data, the OGD algorithm is able to maintain strong generalization performance.
  
  Furthermore, when processing time-series data, OGD avoids overfitting to earlier data points because each update relies primarily on the most 
  recent data. Although OGD lacks the stochastic perturbations present in SGD (i.e., randomness between samples), it is more prone to developing 
  a short-term "memory" effect. In nonlinear systems, OGD effectively fits a local linear model at each time step. Unlike the random sampling 
  characteristic of SGD, the update path of OGD is more strongly influenced by temporal correlations, enabling it to better capture the evolving
   dynamics of the system.

  In contrast, the online DMD algorithm tends to construct a model that precisely fits the training data. While this fitting ability is advantageous 
  in some cases, it often leads to overfitting, especially when the number of model parameters far exceeds the data quantity and the data contain noise, 
  resulting in poor predictive performance. Thus, online DMD is less robust in the face of high noise and small sample sizes, leading to reduced prediction accuracy.

  Online DMD is a type of Recursive Least Squares (RLS) method, which is characterized by its ability to fit data precisely. RLS iteratively 
  adjusts parameters to closely fit each new data point, assuming that data errors can be gradually minimized through optimization. As a result,
   in the presence of noisy data, RLS tends to fit noise as if it were part of the data, lacking the inherent robustness to noise seen in 
   gradient descent's implicit regularization. RLS closely tracks the data, making it challenging to achieve low-rank simplification under 
   noisy conditions. Compared to SGD, recursive methods are more inclined to fit the data precisely, including the noise, which may lead the 
   model to "memorize" noise characteristics.
  
  \section{Numerical Examples}
  This section provides additional numerical examples that were not included in the main text.
  Numerical experiment results indicate that, compared to traditional DMD methods, introducing appropriate physical constraints in OPIDMD generally achieves
   the best predictive performance. Suitable physical constraints help reduce noise interference, allowing OPIDMD to not only fit the data accurately but 
   also generalize well to unseen data. However, similar to piDMD, if the imposed physical constraints do not align with the actual system, OPIDMD's
    predictive performance may decline significantly.
  
  \subsection{Schrödinger Equation}\label{examples}
   First, we use the one-dimensional Schrödinger equation in a free potential field as a test case. The full partial differential
   equation, along with the initial and boundary conditions, is given by:
  \begin{equation}
    \begin{cases}
      i \hbar \frac{\partial \psi(x, t)}{\partial t} = -\frac{\hbar^2}{2m} \frac{\partial^2 \psi(x, t)}{\partial x^2}, & x \in \left(-\frac{L_x}{2}, \frac{L_x}{2}\right), \, t > 0 \\[10pt]
      \psi(x, 0) = \exp \left(-\frac{(x - x_0)^2}{2 \sigma^2}\right) \cdot \exp(i k_0 x), & x \in \left[-\frac{L_x}{2}, \frac{L_x}{2}\right] \\[10pt]
      \psi\left(-\frac{L_x}{2}, t\right) = \psi\left(\frac{L_x}{2}, t\right), & t > 0
      \end{cases}
  \end{equation}
  where Planck’s constant \( \hbar = 1 \), particle mass \( m = 1 \), spatial domain length \( L_x = 10 \), initial wave packet 
  position \( x_0 = 0 \), initial momentum \( k_0 = 20 \), and wave packet width \( \sigma = 1 \). The initial condition \( \psi(x, 0) \) 
  describes the wave function's state at \( t = 0 \), and periodic boundary conditions ensure continuity of the wave function at the boundaries 
  along the \( x \)-axis.
  
\begin{figure}[!ht]
    \centering
    \vspace{-0.5em}
    \begin{subfigure}{0.41\textwidth}
        \centering
        \includegraphics[width=\textwidth]{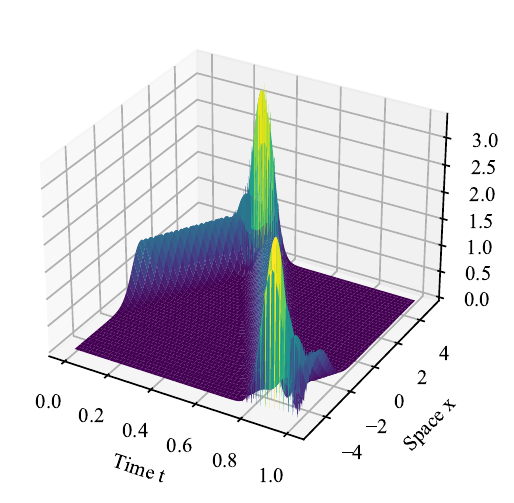}
        \caption{Without noise}
    \end{subfigure}
    \hspace{1em}
    \begin{subfigure}{0.41\textwidth}
        \centering
        \includegraphics[width=\textwidth]{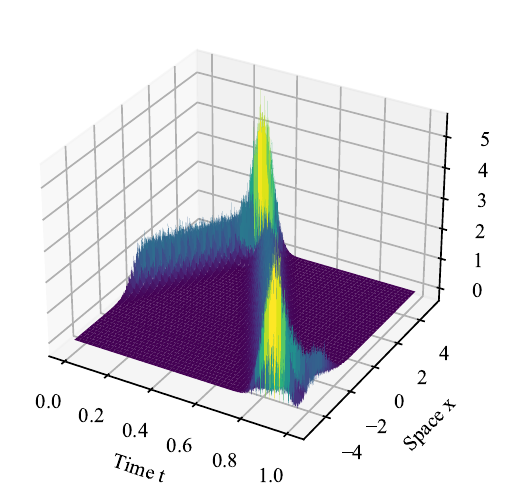}
        \caption{With noise}
    \end{subfigure}
    \caption{Solution of the one-dimensional Schrödinger equation in a free potential field}
    \label{fig:wave_packet}
  \end{figure}

  For the numerical implementation, we create a uniform spatial grid with \( N_x = 200 \) grid points, resulting in a spatial step 
  size \( dx = \frac{L_x}{N_x} \). For time discretization, the total simulation time \( T = 1.0 \) is divided into \( N_t = 10000 \) 
  time steps, giving a time step size \( dt = 0.0001 \) to ensure numerical stability. The Crank-Nicolson method is used to discretize and 
  solve the Schrödinger equation. The solution to the Schrödinger Equation is shown in Figure \ref{fig:wave_packet}.
  
\begin{figure}[!ht]
    \centering
    \begin{subfigure}{0.29\textwidth}
        \centering
        \includegraphics[width=\textwidth]{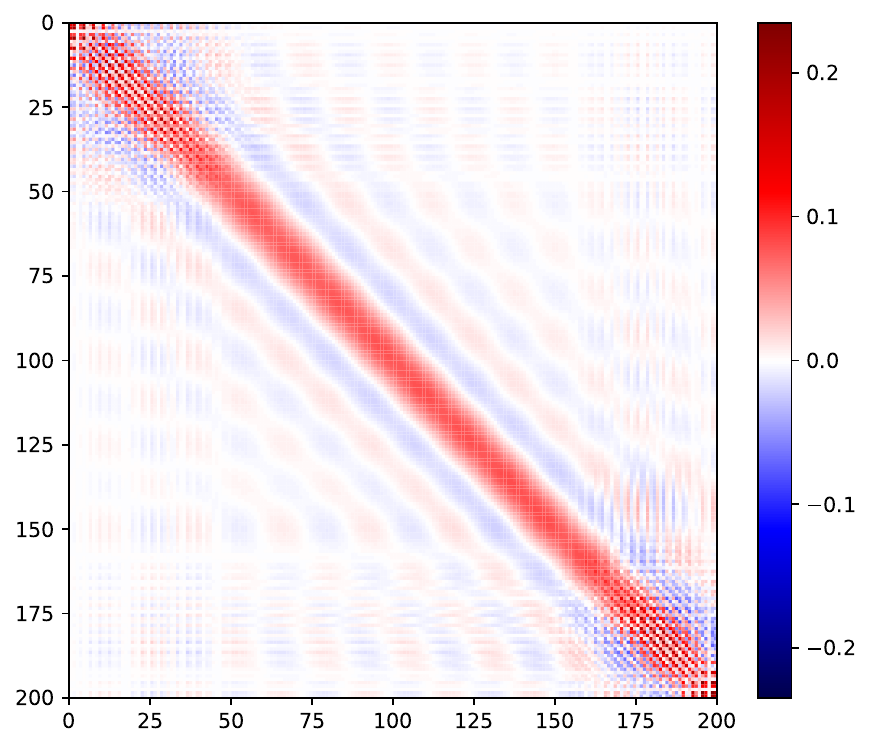}
        \caption{True DMD}
    \end{subfigure}
    \hspace{2em}
    \begin{subfigure}{0.28\textwidth}
        \centering
        \includegraphics[width=\textwidth]{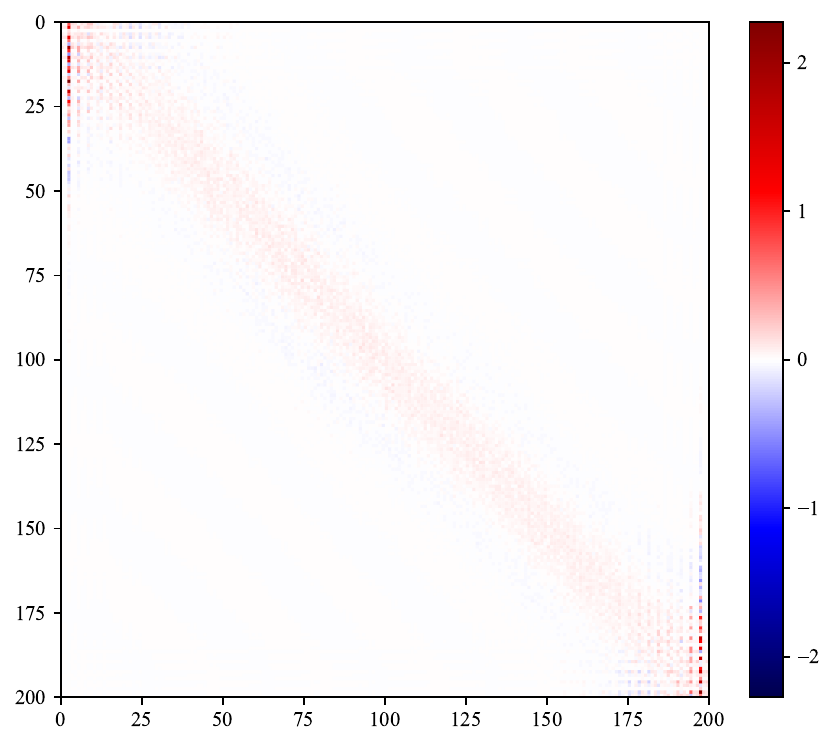}
        \caption{Standard DMD}
    \end{subfigure}
    \hspace{2em}
    \begin{subfigure}{0.29\textwidth}
        \centering
        \includegraphics[width=\textwidth]{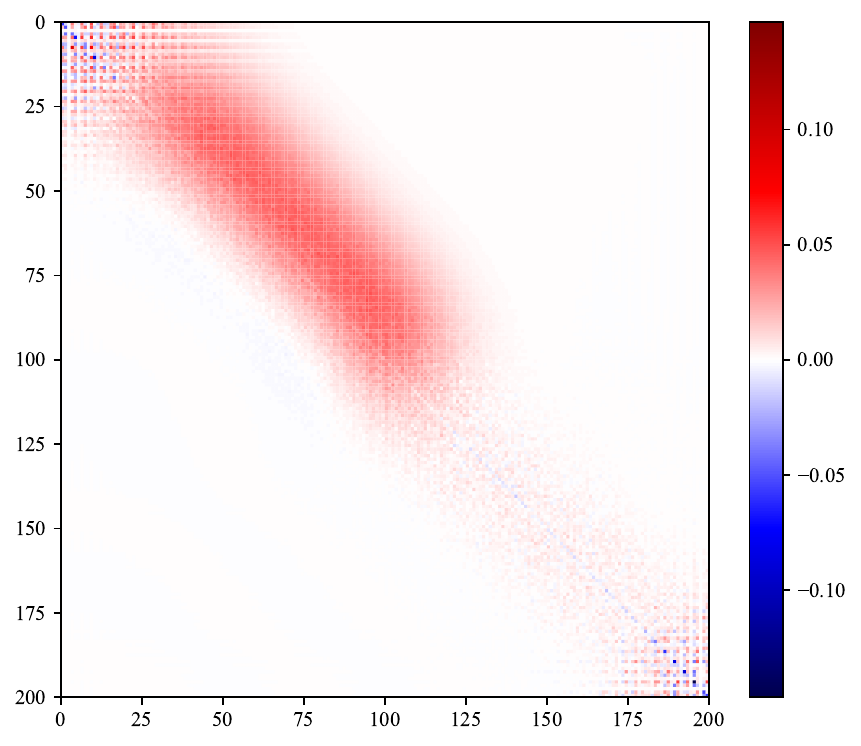}
        \caption{OGD}
    \end{subfigure}
  
    \vspace{-0.5em}
  
    \begin{subfigure}{0.29\textwidth}
        \centering
        \includegraphics[width=\textwidth]{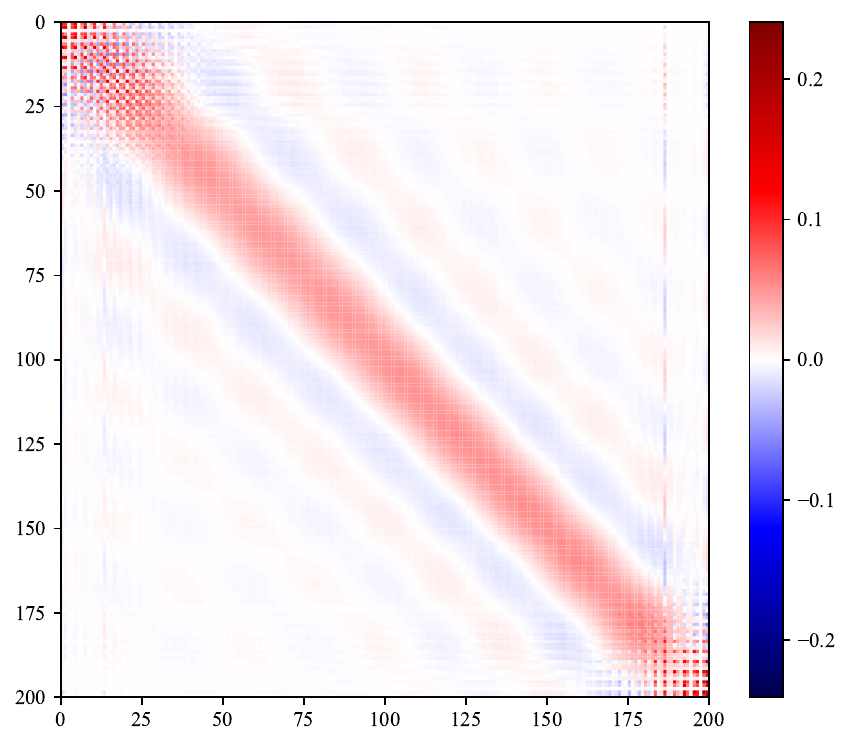}
        \caption{Exact DMD}
    \end{subfigure}
    \hspace{2em}
    \begin{subfigure}{0.29\textwidth}
        \centering
        \includegraphics[width=\textwidth]{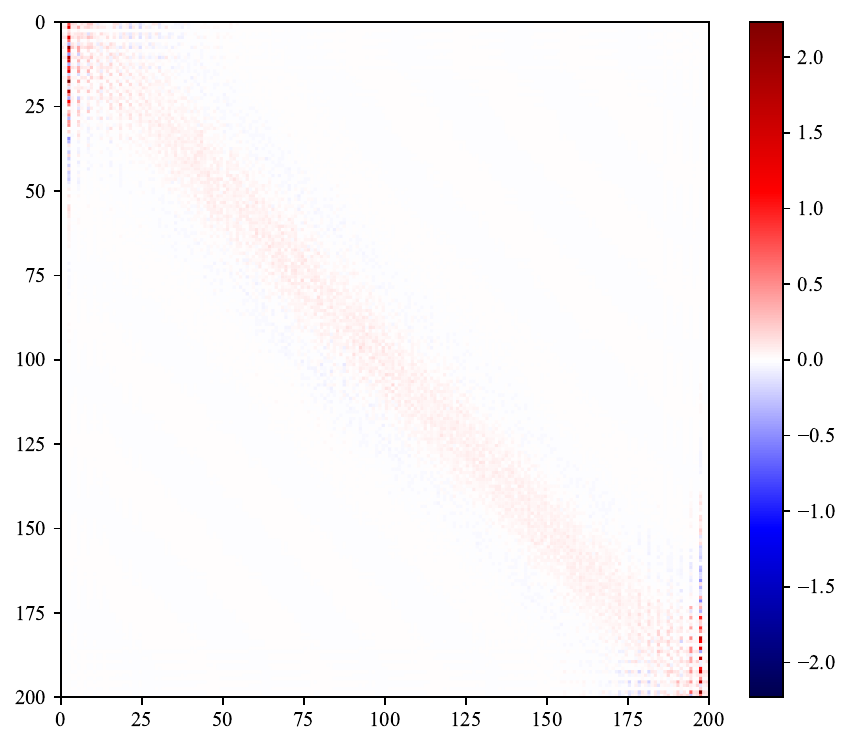}
        \caption{Online DMD}
    \end{subfigure}
    \hspace{2em}
    \begin{subfigure}{0.29\textwidth}
        \centering
        \includegraphics[width=\textwidth]{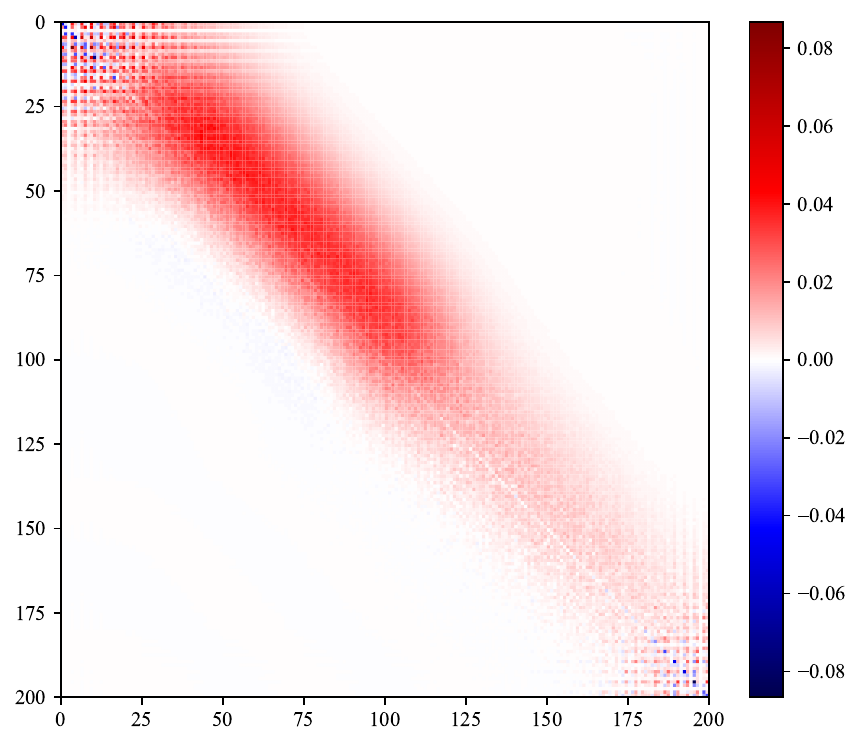}
        \caption{\( l_2 \) Norm OPIDMD}
    \end{subfigure}
    \caption{Comparison of DMD Matrices}
    \label{fig:Schrodinger1}
  \end{figure}

\begin{figure}[!ht]
   \centering
   \begin{subfigure}{0.29\textwidth}
       \centering
       \includegraphics[width=\textwidth]{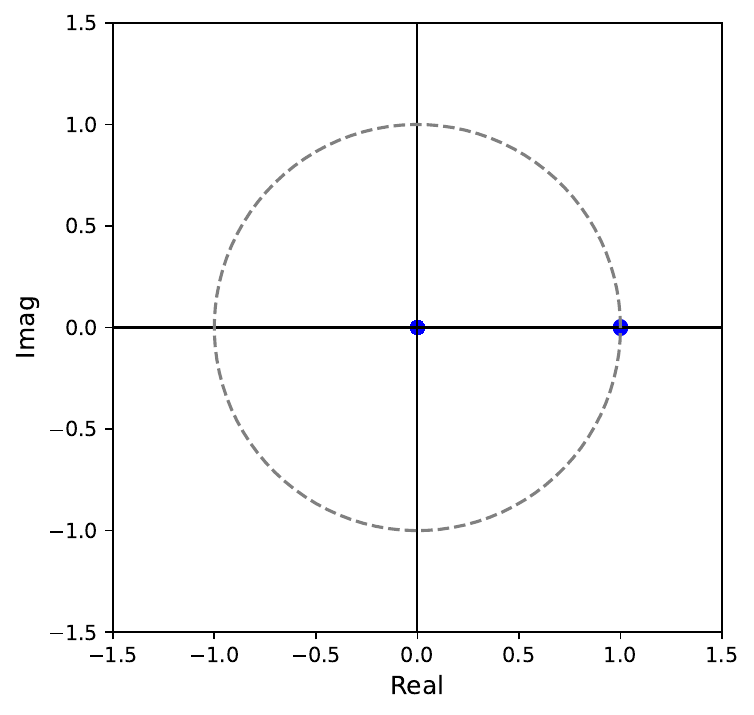}
       \caption{True DMD}
   \end{subfigure}
   \hspace{2em}
   \begin{subfigure}{0.29\textwidth}
       \centering
       \includegraphics[width=\textwidth]{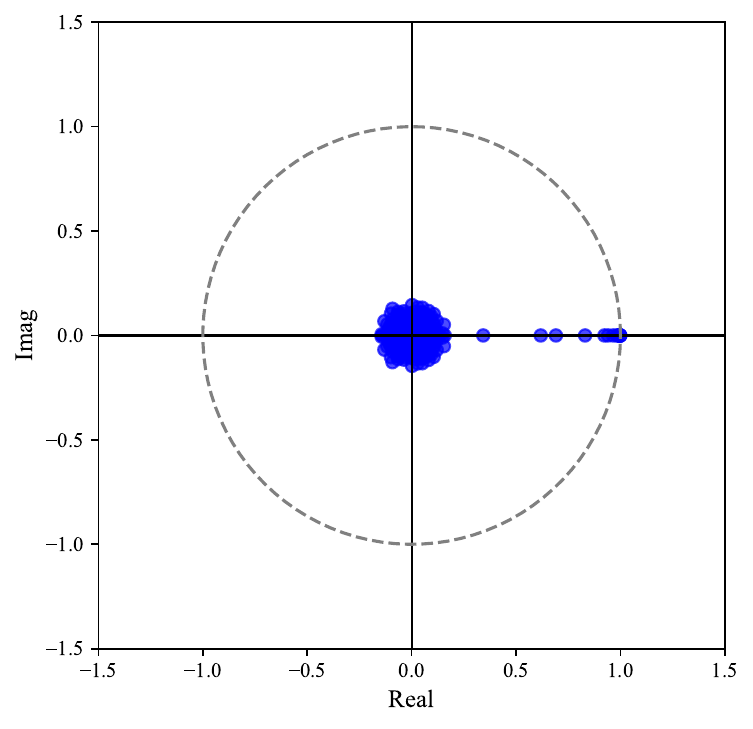}
       \caption{Standard DMD}
   \end{subfigure}
   \hspace{2em}
   \begin{subfigure}{0.29\textwidth}
       \centering
       \includegraphics[width=\textwidth]{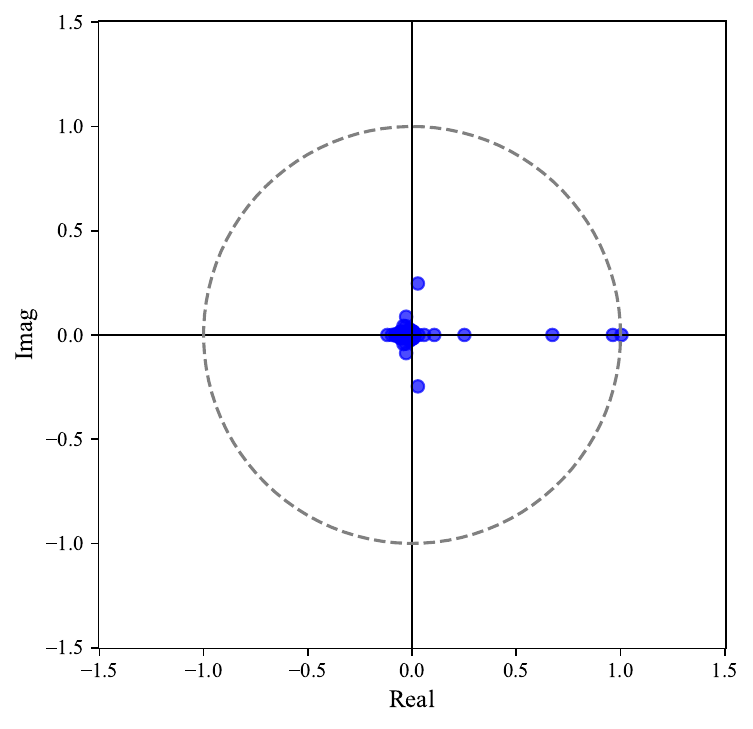}
       \caption{OGD}
   \end{subfigure}
 
   \vspace{-0.5em}
 
   \begin{subfigure}{0.29\textwidth}
       \centering
       \includegraphics[width=\textwidth]{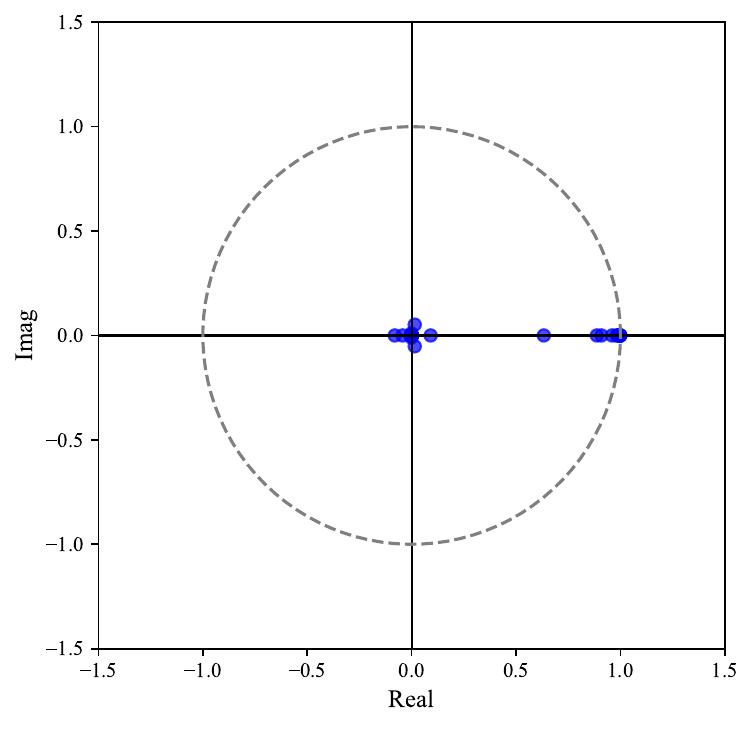}
       \caption{Exact DMD}
   \end{subfigure}
   \hspace{2em}
   \begin{subfigure}{0.29\textwidth}
       \centering
       \includegraphics[width=\textwidth]{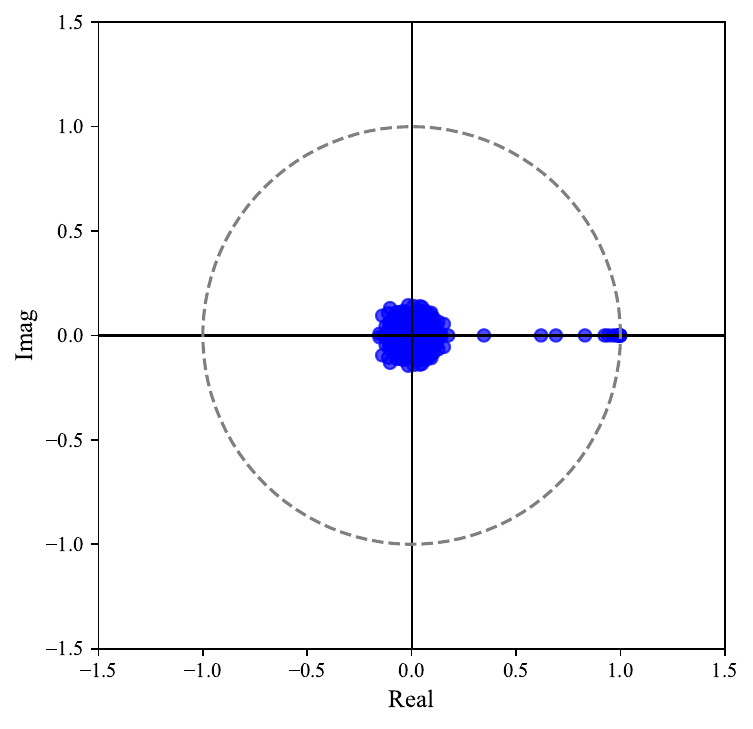}
       \caption{Online DMD}
   \end{subfigure}
   \hspace{2em}
   \begin{subfigure}{0.29\textwidth}
       \centering
       \includegraphics[width=\textwidth]{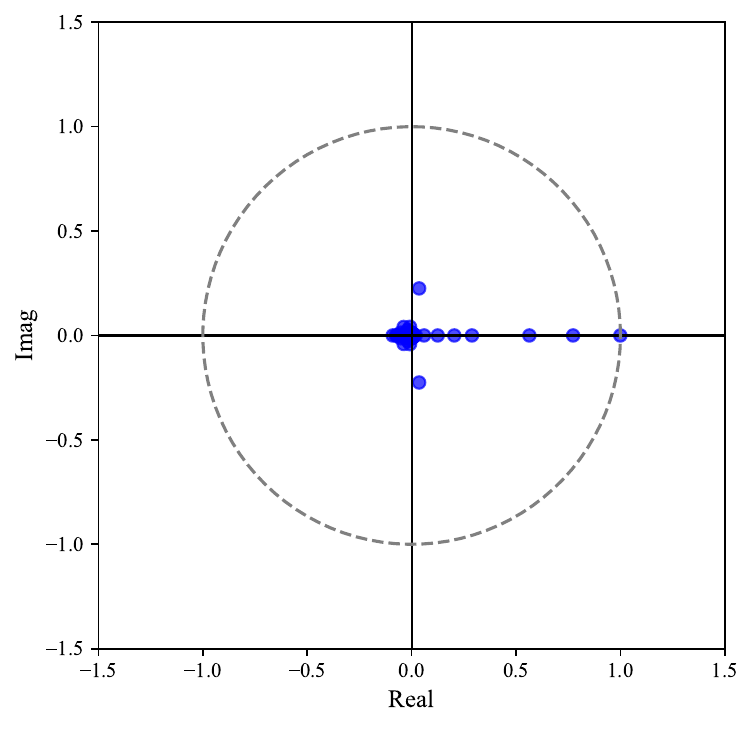}
       \caption{\( l_2 \) Norm OPIDMD}
   \end{subfigure}
   \caption{Eigenvalue Distributions}
   \label{fig:Schrodinger2}
 \end{figure}

  For this study, we use the first 9850 noisy time steps for model training, with the noise-free data at step 9850 serving as the 
  initial condition for predicting the subsequent 150 noise-free steps to evaluate each model's predictive performance. Figures 
  \ref{fig:Schrodinger1} and \ref{fig:Schrodinger2} display the DMD matrices and their corresponding eigenvalue distributions generated
   by different algorithms.
  
The true DMD matrix exhibits a certain symmetry and has a structure similar to a circulant matrix. Due to the increased number of time steps in this
 test case compared to the partial differential equation example in the main text, the results of the online DMD method show significant improvement, 
 with the matrix structure closely resembling that of the standard DMD method. However, the eigenvalues are relatively concentrated and deviate from the 
 origin, indicating that this method still has limitations in capturing the system's long-term stability. The online DMD method remains sensitive to noise 
 and may exhibit overfitting in some cases.

 Exact DMD exhibits a matrix structure similar to that of True DMD. The matrix structure generated by the OGD algorithm shows approximate low-rank characteristics, appearing smooth and sparse. Its eigenvalue distribution is concentrated near the origin, with a few eigenvalues around the unit circle, indicating good performance in terms of smoothness and noise adaptation. The matrix with \( l_2 \) regularization is similar to that of OGD, resembling a low-rank structure. Its eigenvalues are primarily clustered near zero, gradually extending towards the unit circle. In terms of prediction accuracy, the \( l_2 \)-regularized method effectively reduces the impact of noise, thereby enhancing predictive accuracy.
 
\begin{figure}[!ht]
     \centering
     \begin{subfigure}{0.29\textwidth}
         \centering
         \includegraphics[width=\textwidth]{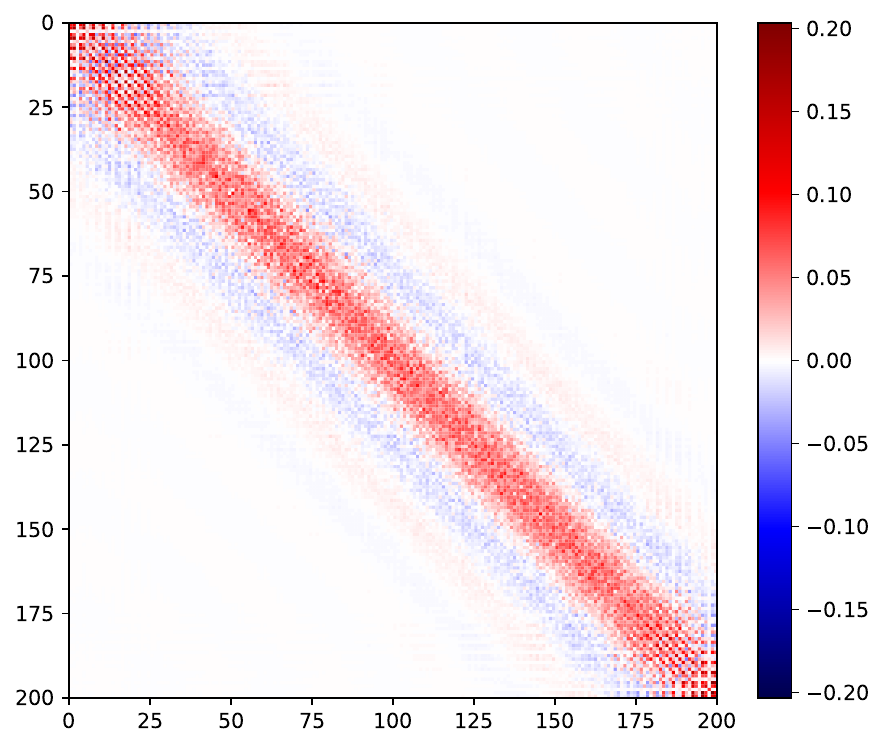}
         \caption{Symmetric piDMD}
     \end{subfigure}
     \hspace{2em}
     \begin{subfigure}{0.29\textwidth}
         \centering
         \includegraphics[width=\textwidth]{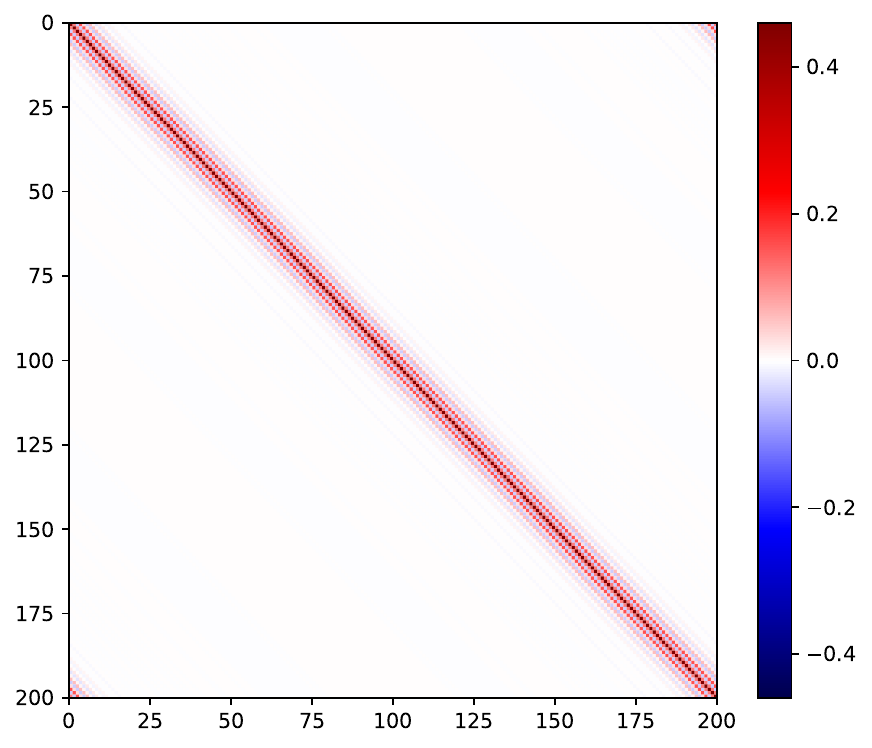}
         \caption{Circulant piDMD}
       \end{subfigure}
       \hspace{2em}
       \begin{subfigure}{0.29\textwidth}
           \centering
           \includegraphics[width=\textwidth]{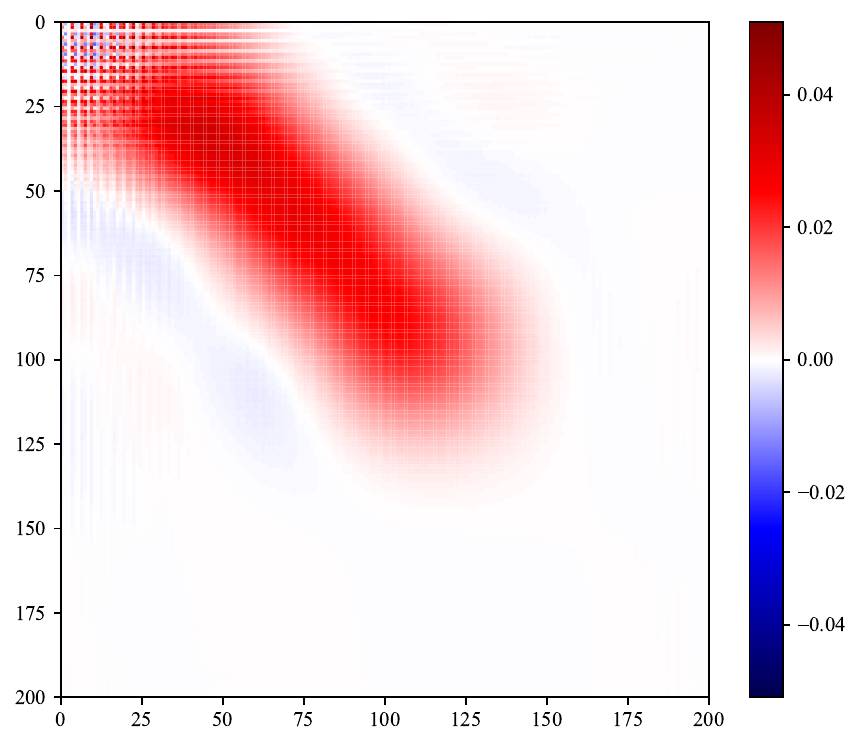}
           \caption{Low-Rank OPIDMD}
     \end{subfigure}
   
     \vspace{-0.5em}
 
     \begin{subfigure}{0.29\textwidth}
         \centering
         \includegraphics[width=\textwidth]{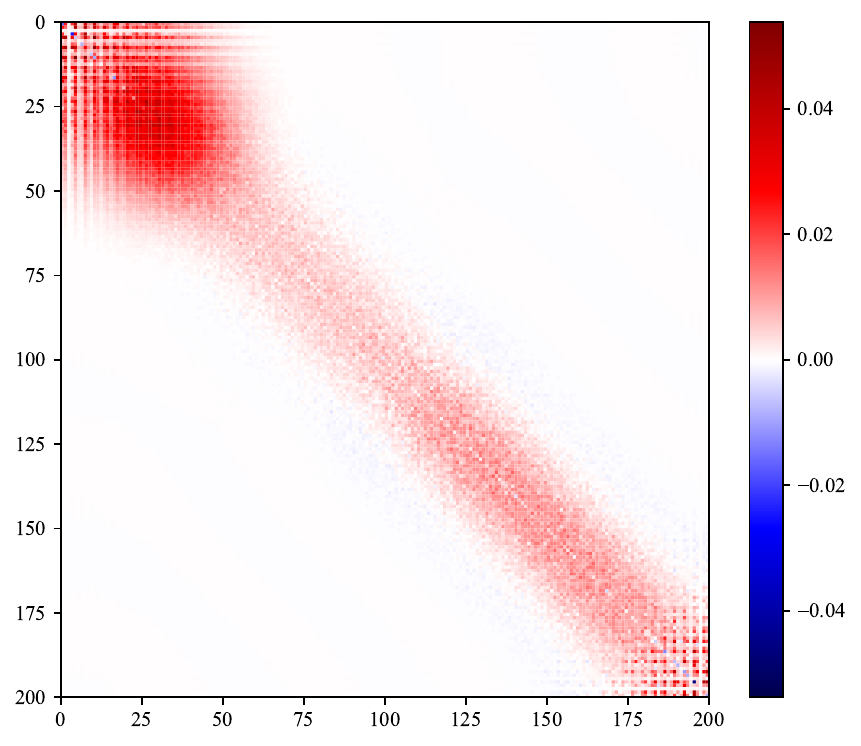}
         \caption{Symmetric OPIDMD}
     \end{subfigure}
     \hspace{2em}
     \begin{subfigure}{0.29\textwidth}
         \centering
         \includegraphics[width=\textwidth]{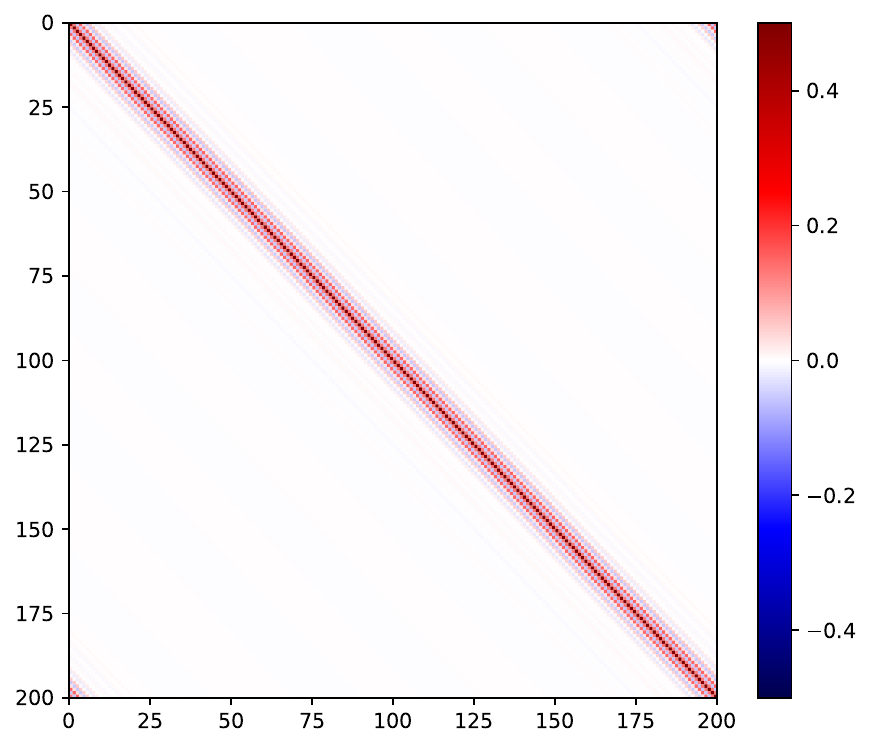}
         \caption{Circulant OPIDMD}
     \end{subfigure}
     \hspace{2em}
     \begin{subfigure}{0.29\textwidth}
       \centering
       \includegraphics[width=\textwidth]{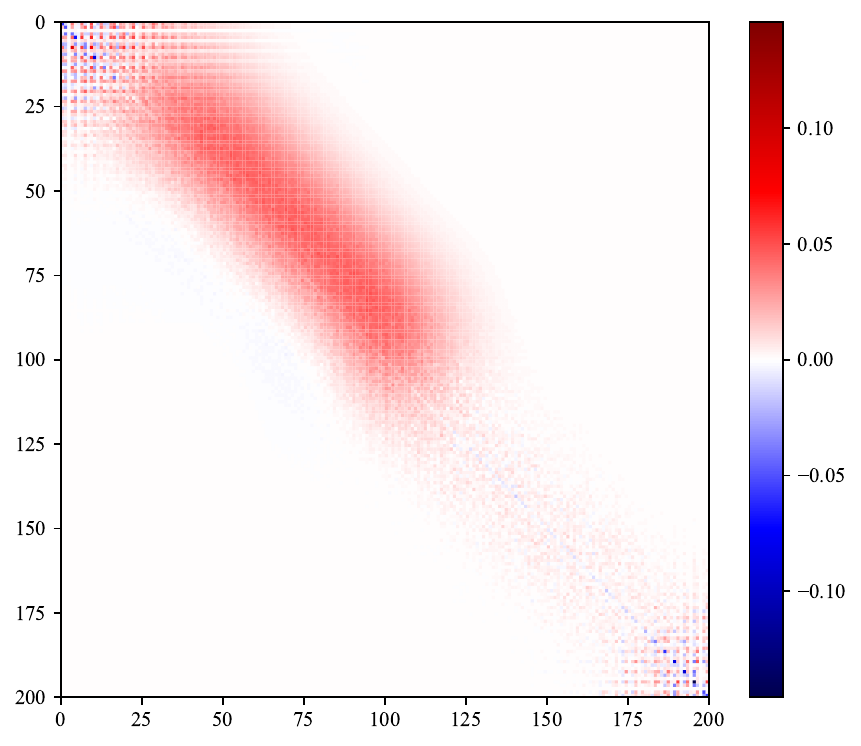}
       \caption{\( l_1 \) Norm OPIDMD}
   \end{subfigure}
   \caption{Comparison of Different DMD Matrices for the Schrödinger System}
     \label{fig:Schrodinger3}
 \end{figure}

\begin{figure}[!ht]
   \centering
   \begin{subfigure}{0.29\textwidth}
       \centering
       \includegraphics[width=\textwidth]{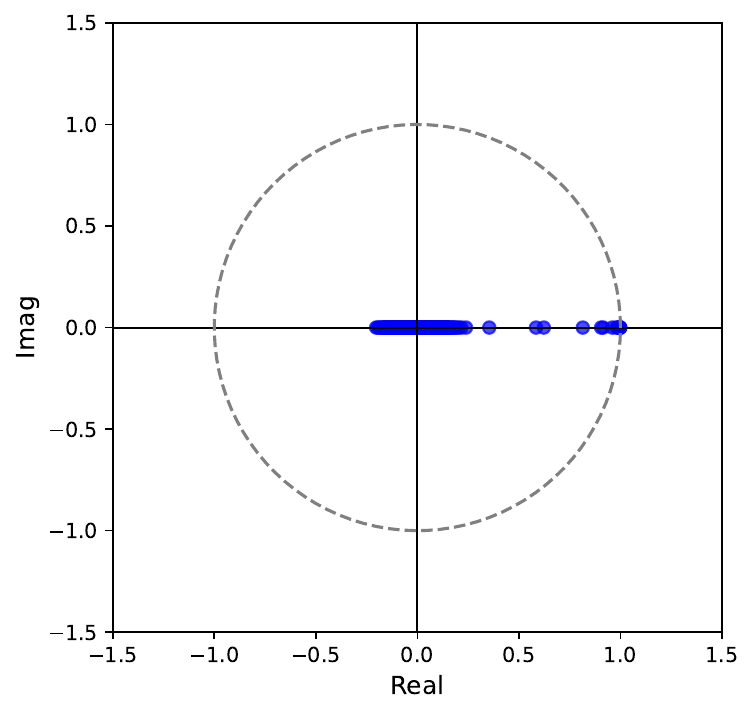}
       \caption{Symmetric piDMD}
   \end{subfigure}
   \hspace{2em}
   \begin{subfigure}{0.29\textwidth}
       \centering
       \includegraphics[width=\textwidth]{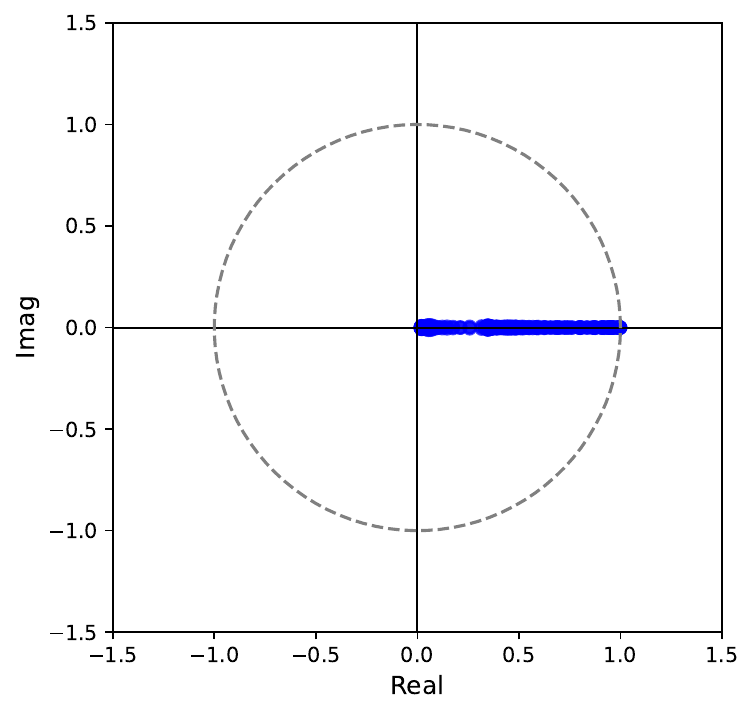}
       \caption{Circulant piDMD}
   \end{subfigure}
   \hspace{2em}
   \begin{subfigure}{0.29\textwidth}
       \centering
       \includegraphics[width=\textwidth]{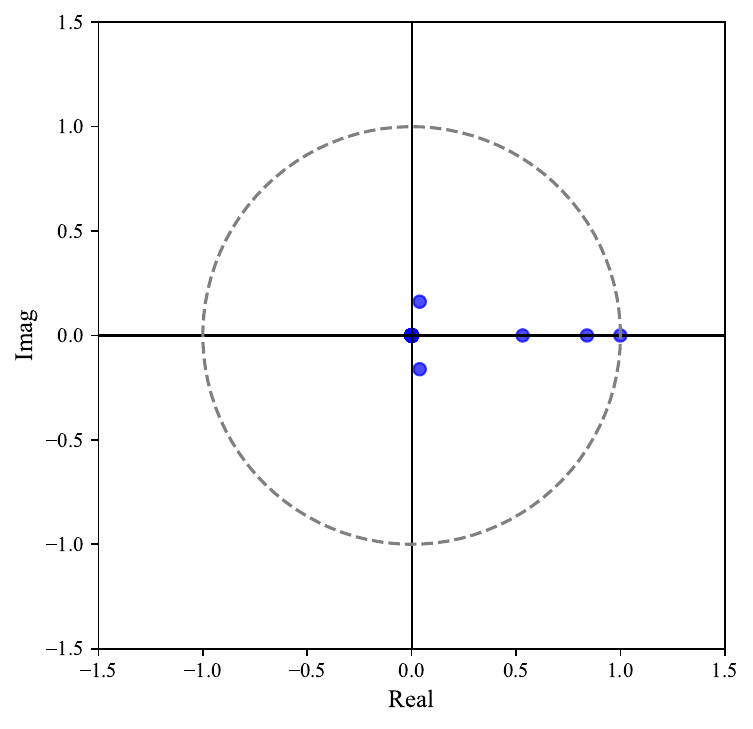}
       \caption{Low-Rank OPIDMD}
   \end{subfigure}
 
   \vspace{-0.5em}
 
   \begin{subfigure}{0.29\textwidth}
       \centering
       \includegraphics[width=\textwidth]{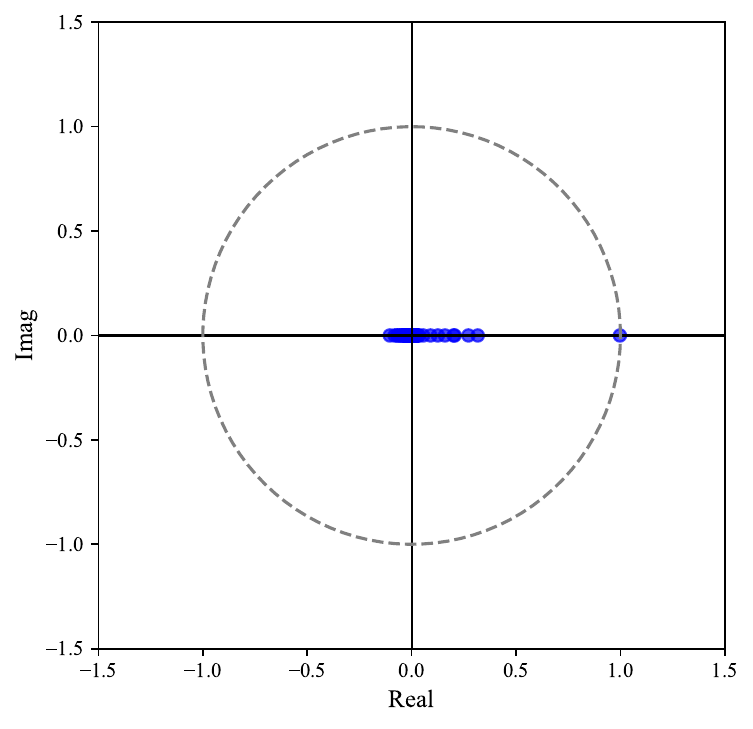}
       \caption{Symmetric OPIDMD}
   \end{subfigure}
   \hspace{2em}
   \begin{subfigure}{0.29\textwidth}
       \centering
       \includegraphics[width=\textwidth]{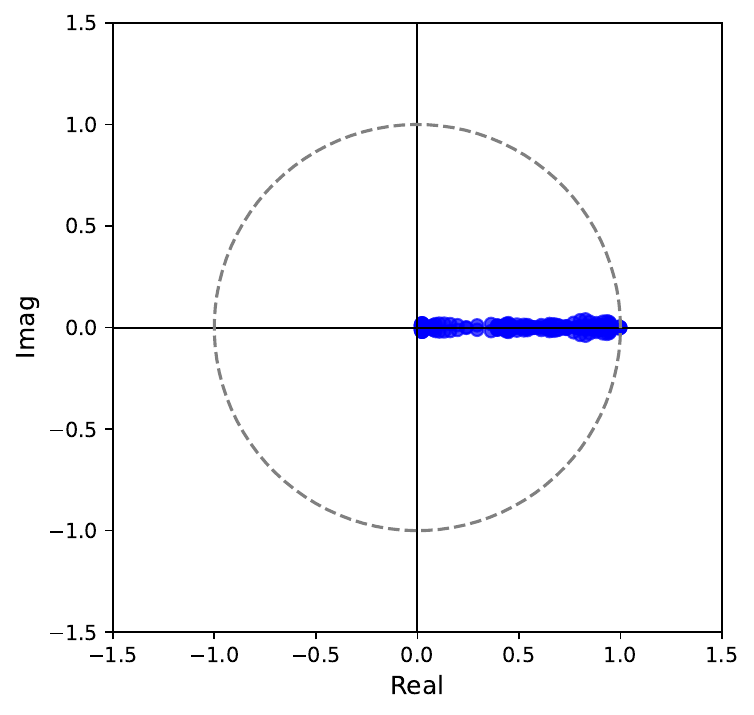}
       \caption{Circulant OPIDMD}
   \end{subfigure}
   \hspace{2em}
   \begin{subfigure}{0.29\textwidth}
       \centering
       \includegraphics[width=\textwidth]{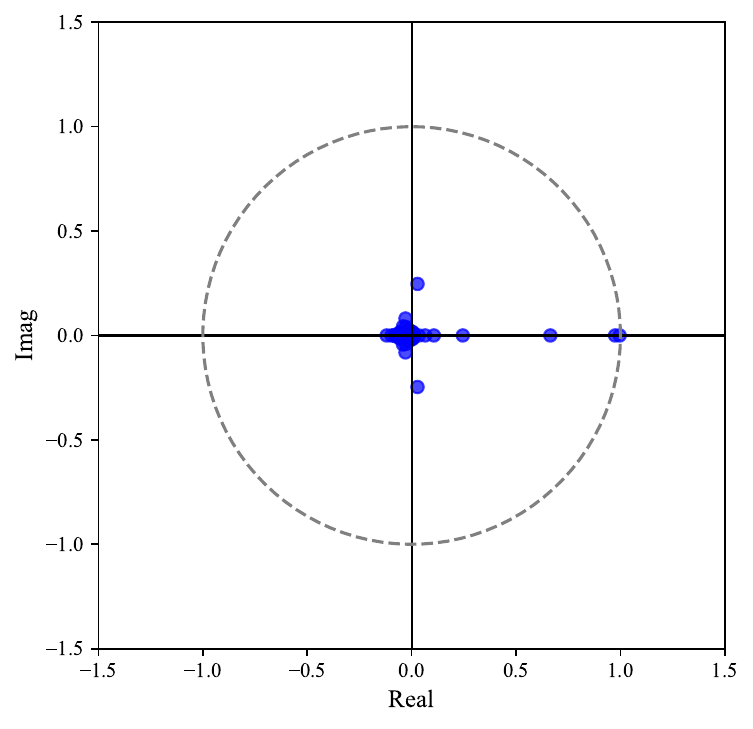}
       \caption{\( l_1 \) Norm OPIDMD}
   \end{subfigure}
 
   \caption{Eigenvalue Distributions for Different DMD Matrix Constraints in the Schrödinger System}
   \label{fig:Schrodinger4}
 \end{figure}

 Figures \ref{fig:Schrodinger3} and \ref{fig:Schrodinger4} illustrate a comparison of the matrix structures and corresponding eigenvalue distributions of piDMD and OPIDMD under various physical constraints.  First, with symmetric constraints applied, the piDMD method produces a matrix with notable symmetry along both sides of the main diagonal, resembling a circulant matrix to some extent. Additionally, the eigenvalues of the piDMD matrix are primarily concentrated near the real axis and are closely distributed, indicating stable capture of the system's dominant modes.
 
 The symmetric matrix in OPIDMD also displays a symmetric structure but is sparser compared to piDMD, lacking the circulant matrix form. Its eigenvalues are distributed along the real axis, mostly clustered near zero, which retains some characteristics of the OGD method, indicating its adaptability to smooth data features.
  
  Under the circulant matrix constraint, both piDMD and OPIDMD exhibit similar matrix structures, showing significant circulant characteristics. Their 
  eigenvalues are uniformly distributed along the real axis and are entirely within the unit circle, suggesting that this type of constraint helps maintain
   system stability and improves model accuracy for long-term predictions. The circulant matrix constraint is also the most effective physical constraint type for piDMD in terms of prediction performance.

 The OPIDMD matrix with low-rank constraints demonstrates a smooth pattern, with eigenvalues primarily concentrated near zero and only a few distributed within the unit circle. This distribution implies that the low-rank constraint effectively smooths the data, enhancing the model's robustness in noisy data scenarios.
 
 In the \( l_1 \)-regularized OPIDMD matrix, a distinct sparsity is observed, with values gradually decaying near the main diagonal. This sparse matrix structure is especially beneficial for matrix element selection. Most of its eigenvalues are clustered near the origin, with a few close to the unit circle. Overall, the situation is similar to OGD, but the prediction accuracy has improved significantly.

 \subsection{Advection-Diffusion Equation}
 
 In this section, we conduct numerical validation using the one-dimensional advection-diffusion equation. The complete partial differential equation, initial condition, and boundary conditions are given as follows:
 \begin{equation}
   \begin{cases}
     \frac{\partial u(x, t)}{\partial t} = D \frac{\partial^2 u(x, t)}{\partial x^2} - v(x) \frac{\partial u(x, t)}{\partial x}, & x \in (0, L), \, t > 0 \\[10pt]
     u(x, 0) = \sin(\pi x), & x \in [0, L] \\[10pt]
     u(0, t) = 0, \quad u(L, t) = 0, & t > 0
     \end{cases}
 \end{equation}
 Here, the diffusion coefficient is set to \( D = 0.1 \), the spatial domain length is \( L = 1.0 \), and the velocity field \( v(x) = 1 + 0.1 \cos(\pi x) \) describes how the fluid velocity varies with position. The initial condition \( u(x, 0) = \sin(\pi x) \) defines the system’s state at \( t = 0 \), while Dirichlet boundary conditions are imposed to ensure that the solution remains zero at \( x = 0 \) and \( x = L \).
 
 To construct the spatial grid, we divide the domain into uniformly spaced points, with the number of grid points set to \( N_x = 100 \), resulting in a spatial step size \( dx = L / N_x = 0.01 \). For time discretization, the total simulation time is set to \( T = 1.0 \), divided into \( N_t = 5000 \) time steps. The time step size \( dt = 0.0002 \) satisfies the CFL condition, ensuring numerical stability and accuracy. The advection term and diffusion term are discretized using upwind and central difference schemes, respectively.

\begin{figure}[!ht]
   \centering
   \begin{subfigure}{0.41\textwidth}
       \centering
       \includegraphics[width=\textwidth]{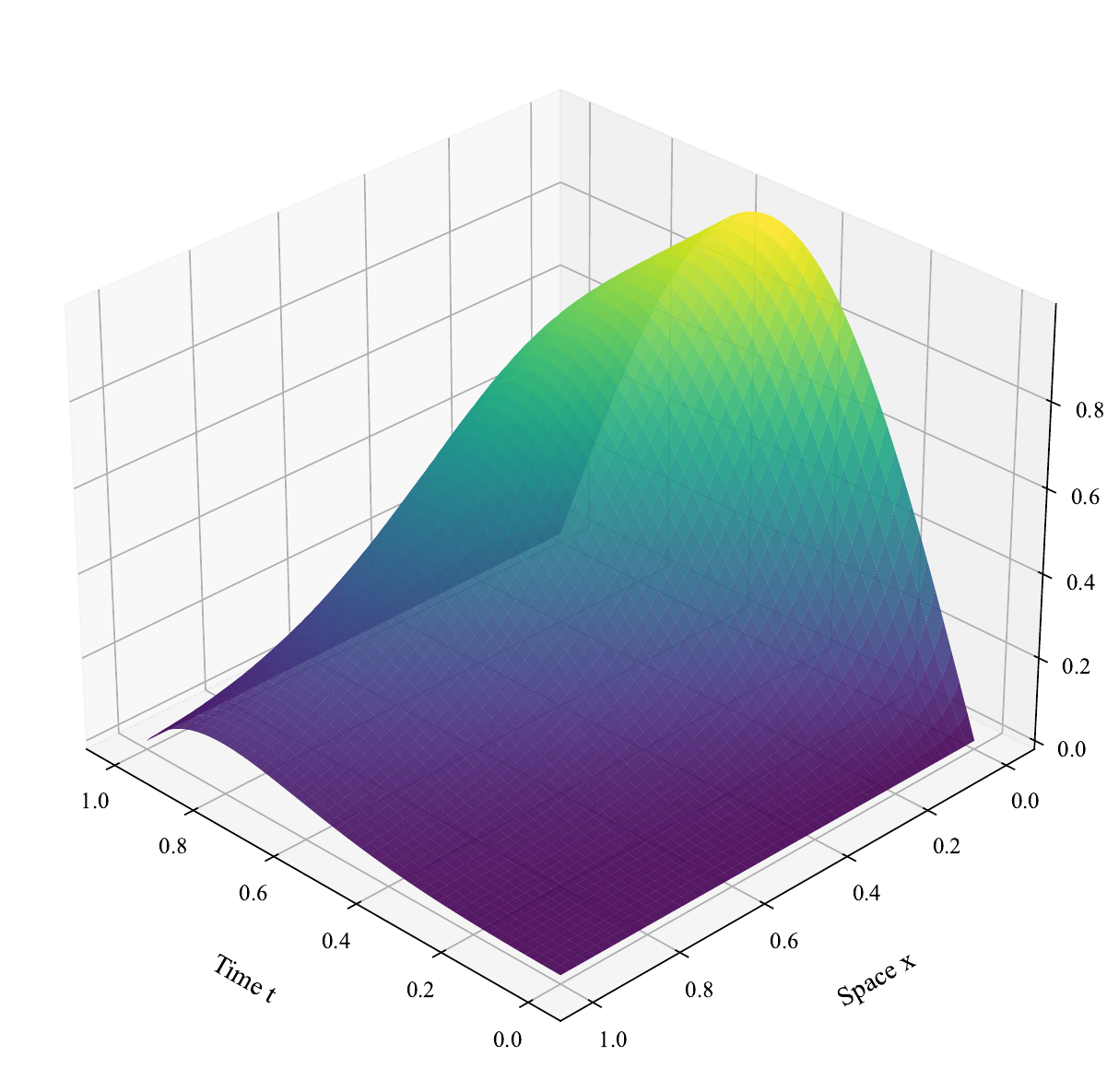}
       \caption{Without noise}
   \end{subfigure}
   \hspace{1em}
   \begin{subfigure}{0.41\textwidth}
       \centering
       \includegraphics[width=\textwidth]{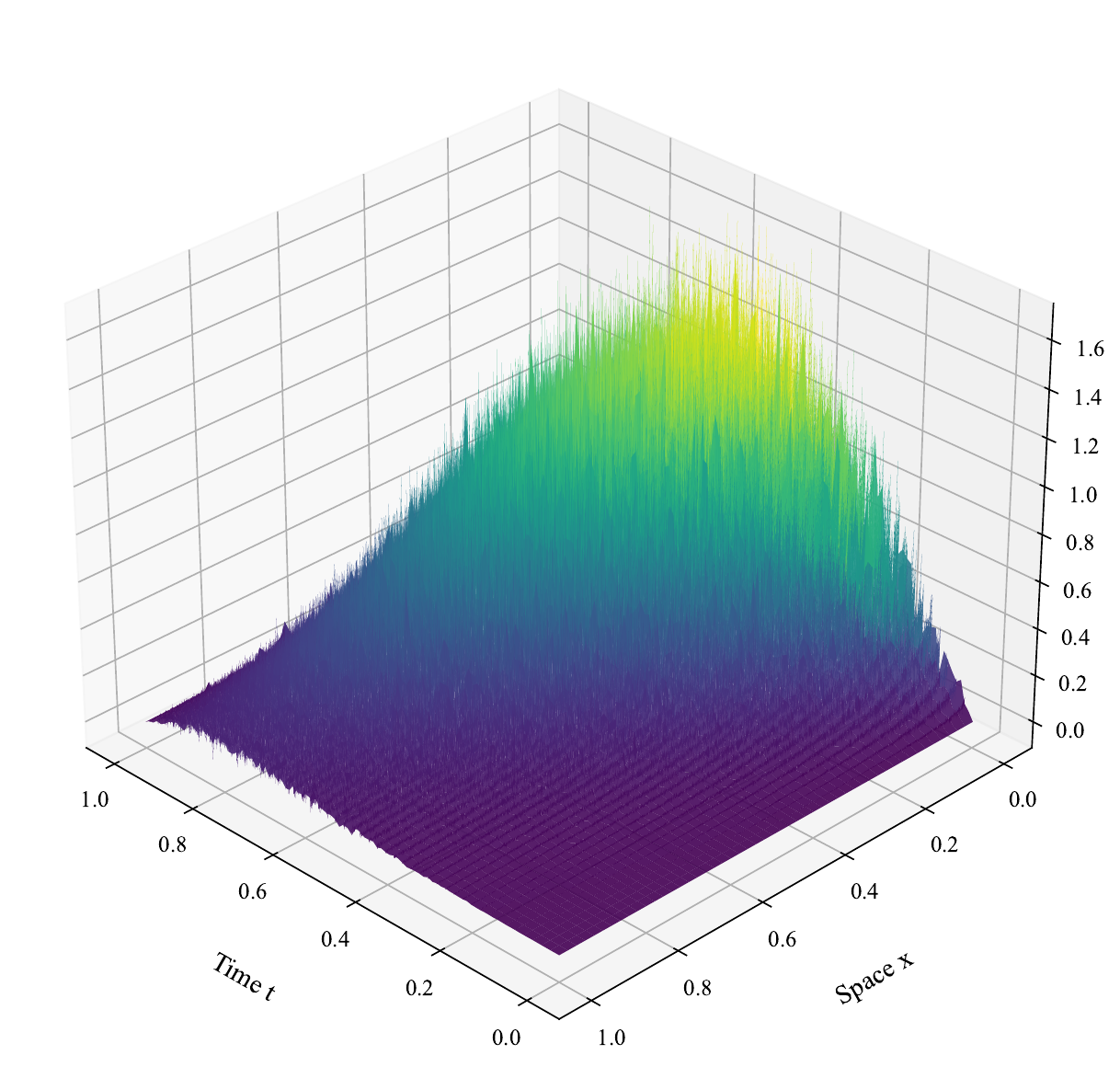}
       \caption{With noise}
   \end{subfigure}
   \caption{Solution of the one-dimensional advection-diffusion equation}
   \label{fig:convection_diffusion}
 \end{figure}

\begin{figure}[!ht]
   \centering
   \begin{subfigure}{0.29\textwidth}
       \centering
       \includegraphics[width=\textwidth]{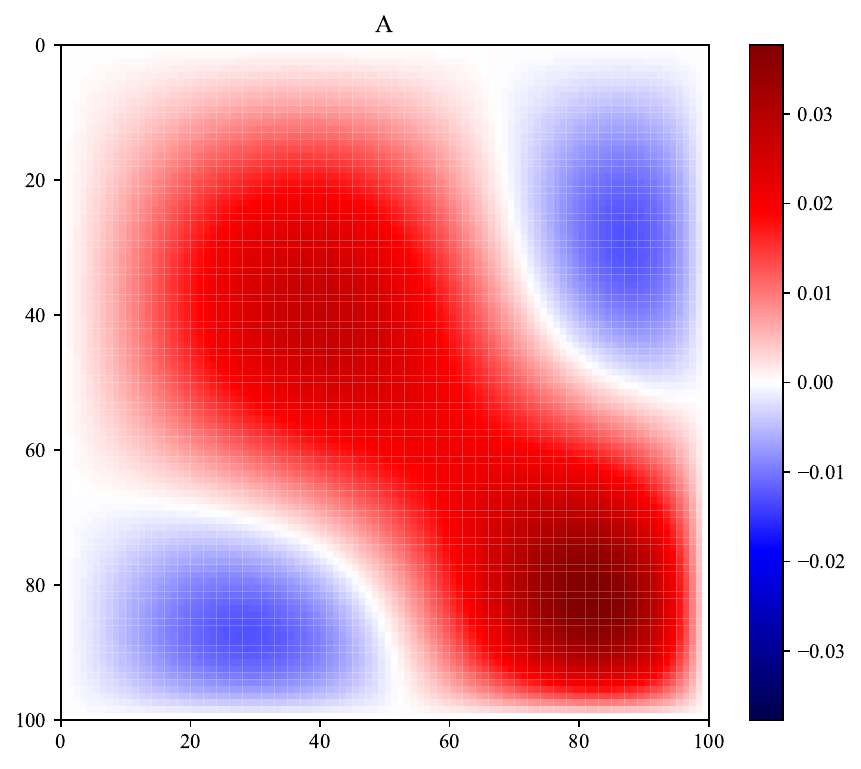}
       \caption{True DMD (rank=2)}
   \end{subfigure}
   \hspace{2em}
   \begin{subfigure}{0.29\textwidth}
       \centering
       \includegraphics[width=\textwidth]{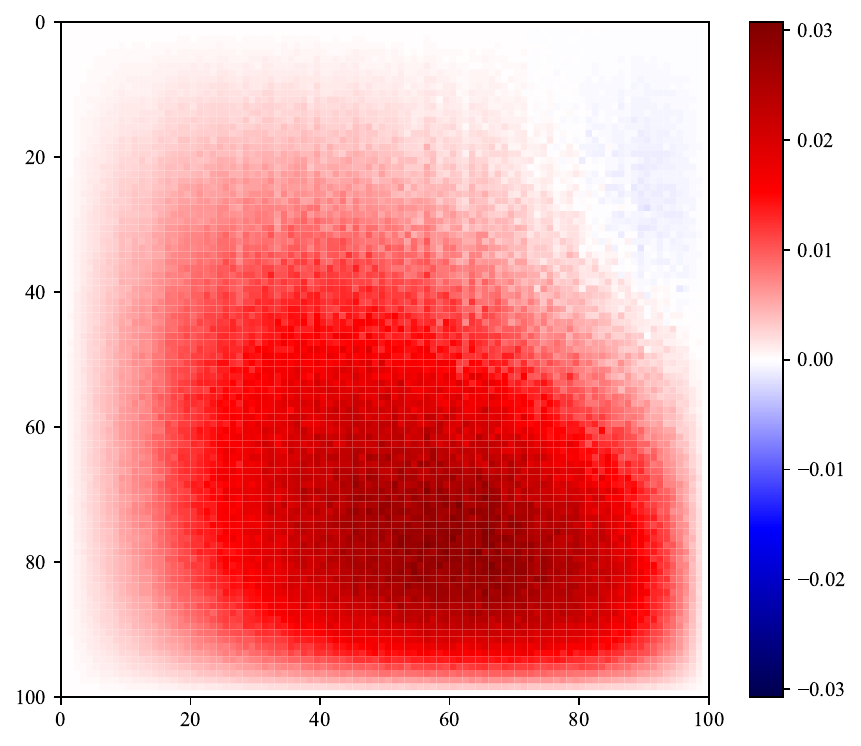}
       \caption{OGD}
   \end{subfigure}
   \hspace{2em}
   \begin{subfigure}{0.29\textwidth}
       \centering
       \includegraphics[width=\textwidth]{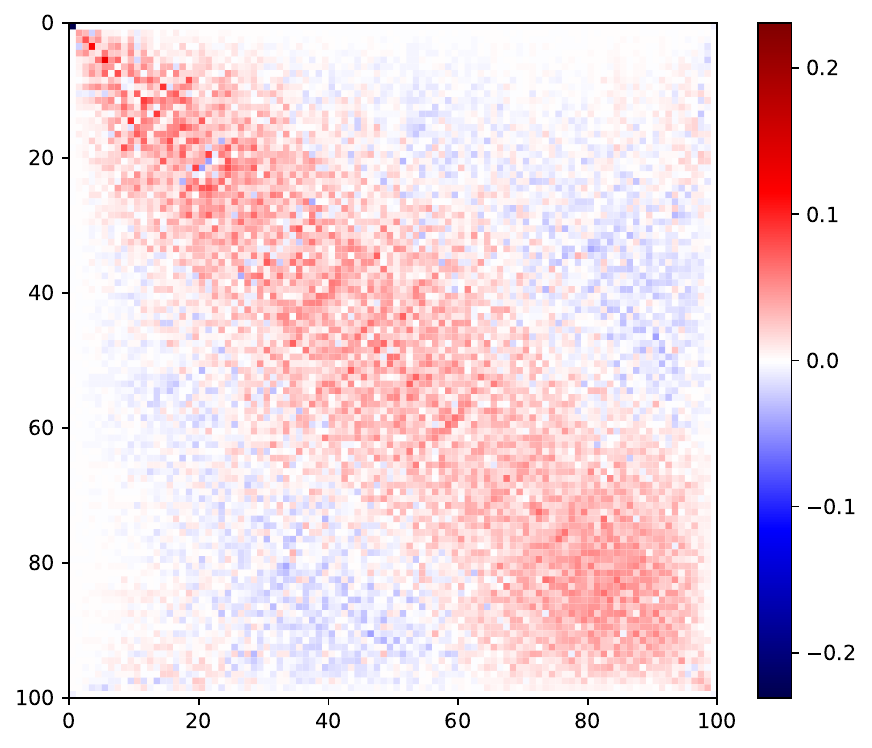}
       \caption{Symmetric piDMD}
   \end{subfigure}
 
   \vspace{1em}
 
   \begin{subfigure}{0.29\textwidth}
       \centering
       \includegraphics[width=\textwidth]{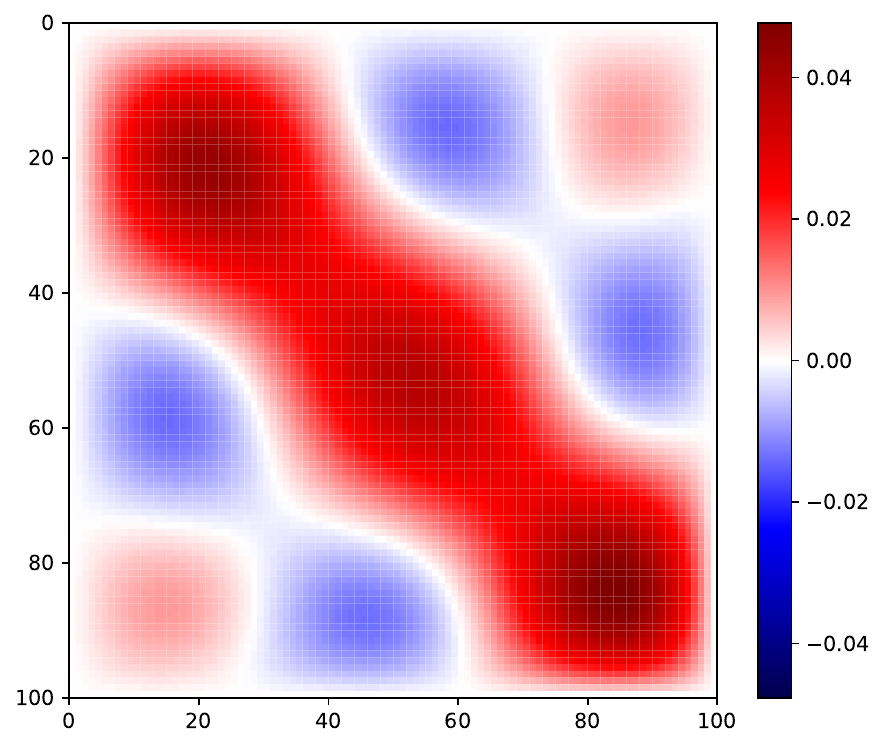}
       \caption{True DMD (rank=1)}
   \end{subfigure}
   \hspace{2em}
   \begin{subfigure}{0.29\textwidth}
       \centering
       \includegraphics[width=\textwidth]{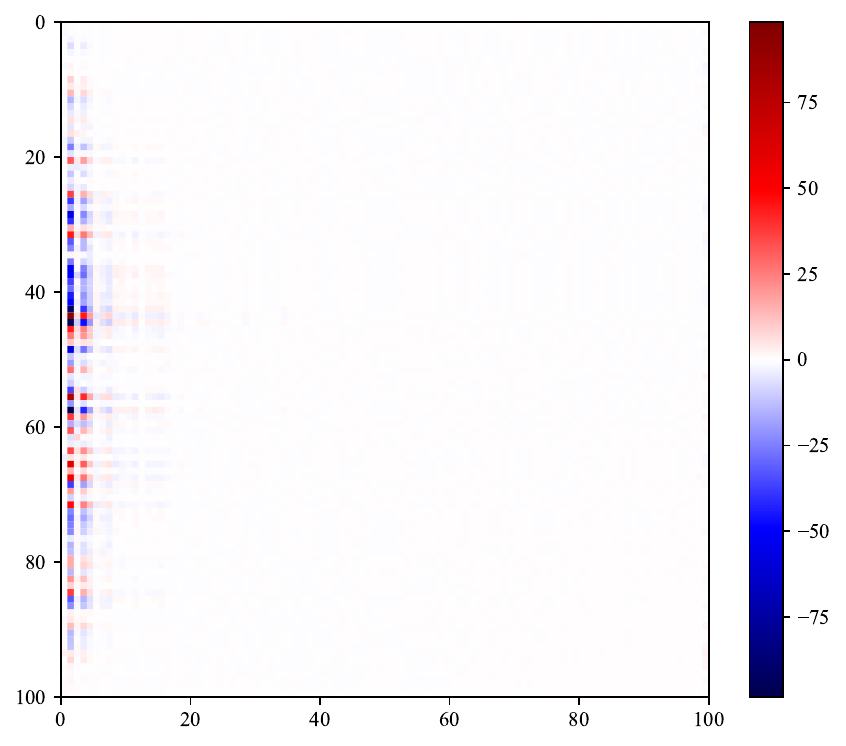}
       \caption{Online DMD}
   \end{subfigure}
   \hspace{2em}
   \begin{subfigure}{0.29\textwidth}
       \centering
       \includegraphics[width=\textwidth]{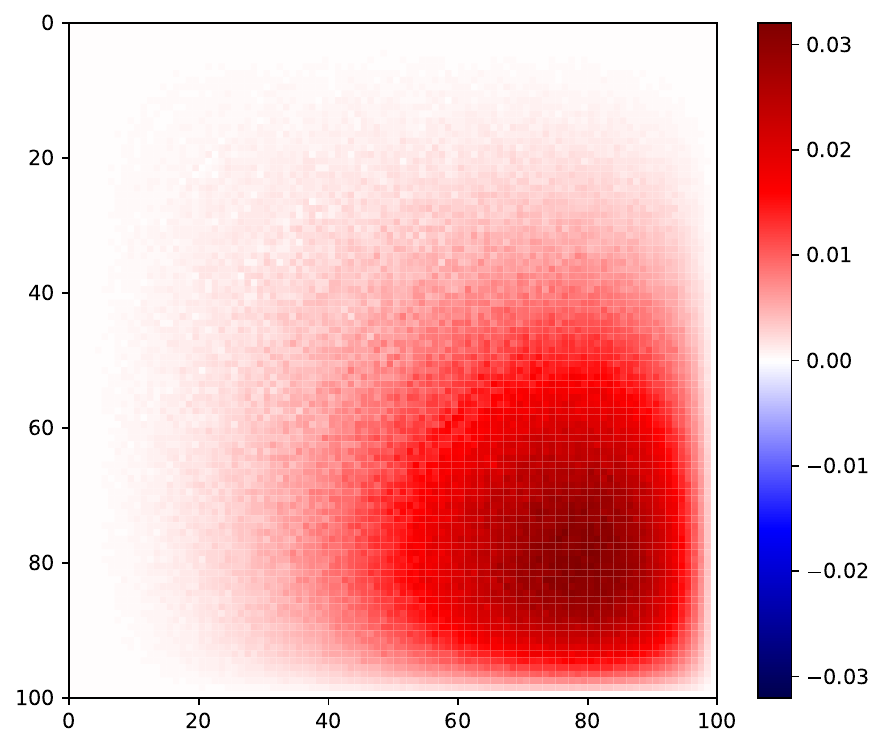}
       \caption{Symmetric OPIDMD}
   \end{subfigure}
   \caption{Comparison of DMD Matrices}
   \label{fig:advection1}
 \end{figure}

\begin{figure}[!ht]
   \centering
   \begin{subfigure}{0.29\textwidth}
       \centering
       \includegraphics[width=\textwidth]{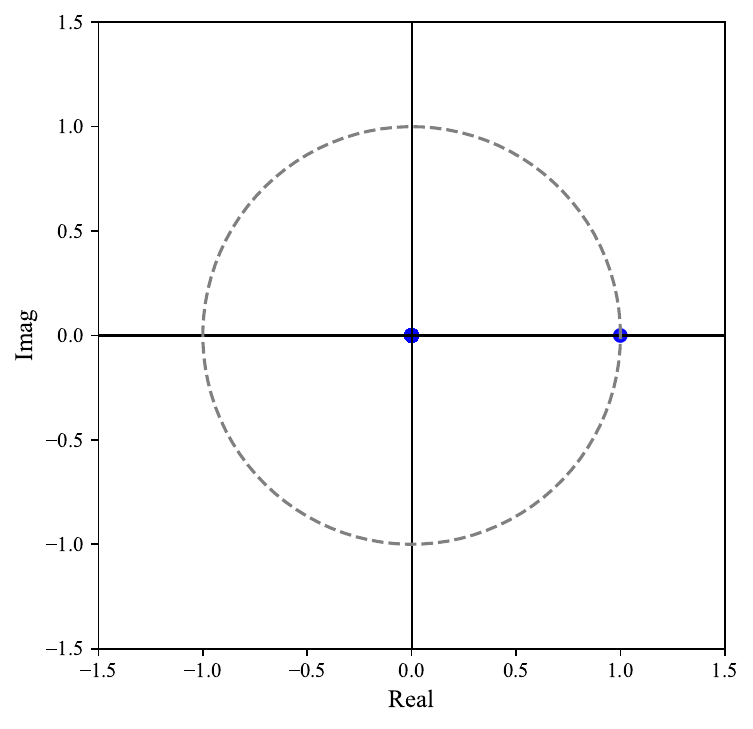}
       \caption{True DMD (rank=2)}
   \end{subfigure}
   \hspace{2em}
   \begin{subfigure}{0.29\textwidth}
       \centering
       \includegraphics[width=\textwidth]{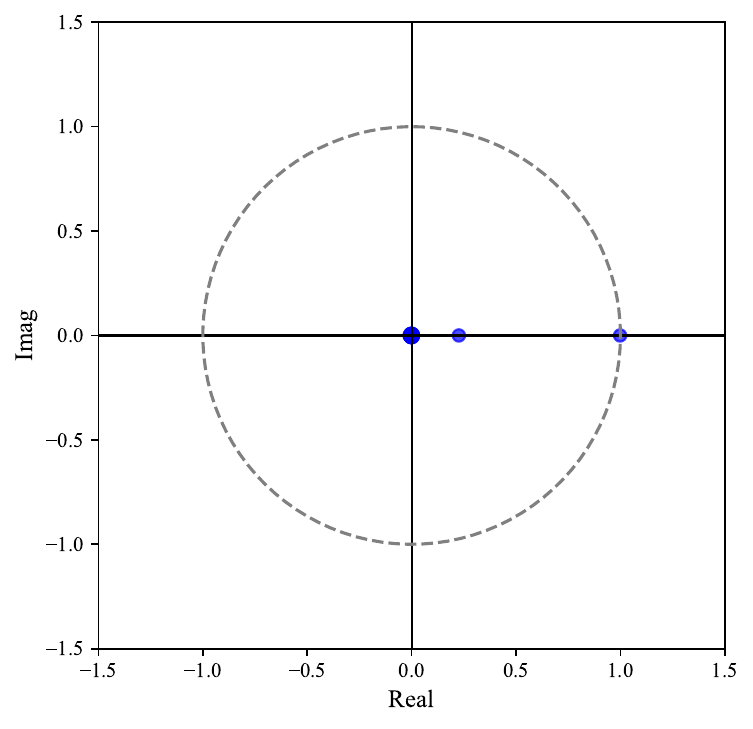}
       \caption{OGD}
   \end{subfigure}
   \hspace{2em}
   \begin{subfigure}{0.29\textwidth}
       \centering
       \includegraphics[width=\textwidth]{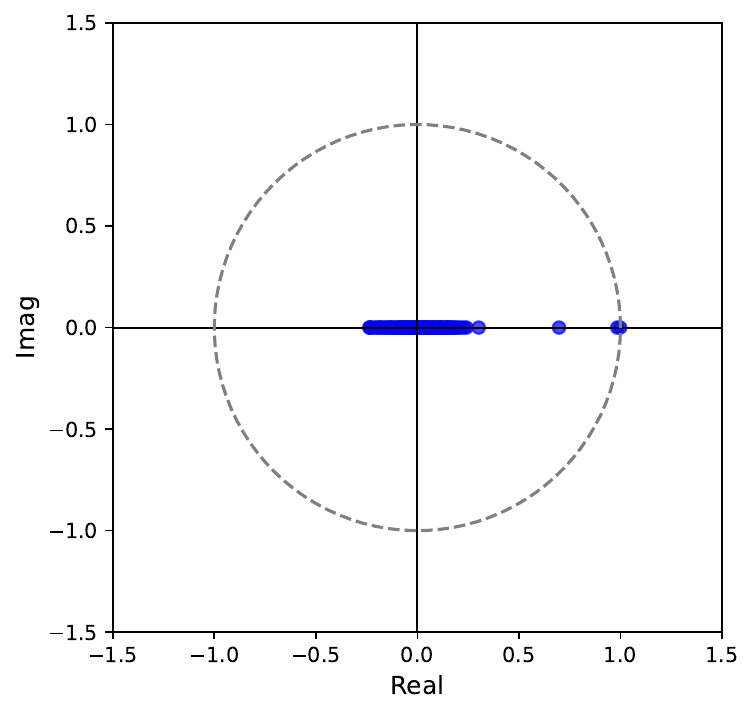}
       \caption{Symmetric piDMD}
   \end{subfigure}
 
   \vspace{1em}
 
   \begin{subfigure}{0.29\textwidth}
       \centering
       \includegraphics[width=\textwidth]{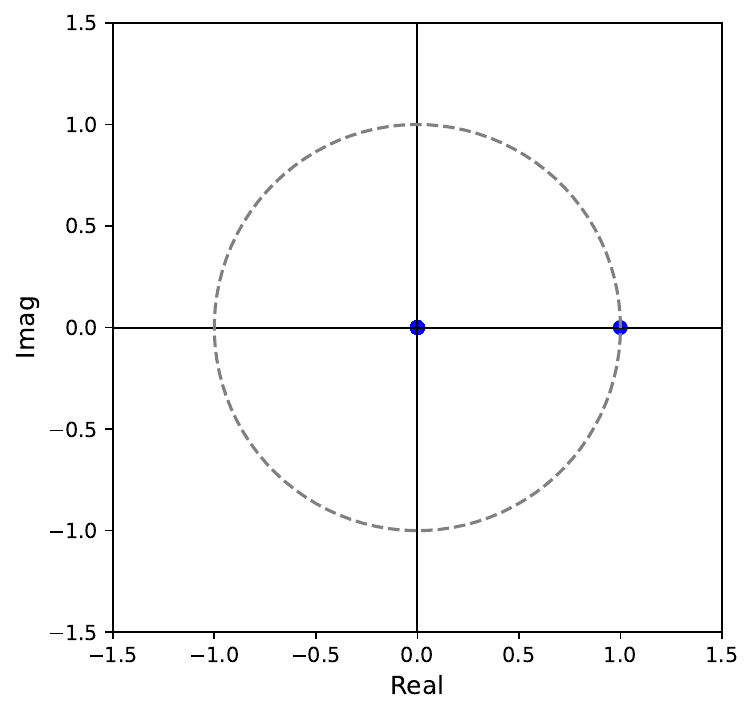}
       \caption{True DMD (rank=1)}
   \end{subfigure}
   \hspace{2em}
   \begin{subfigure}{0.29\textwidth}
       \centering
       \includegraphics[width=\textwidth]{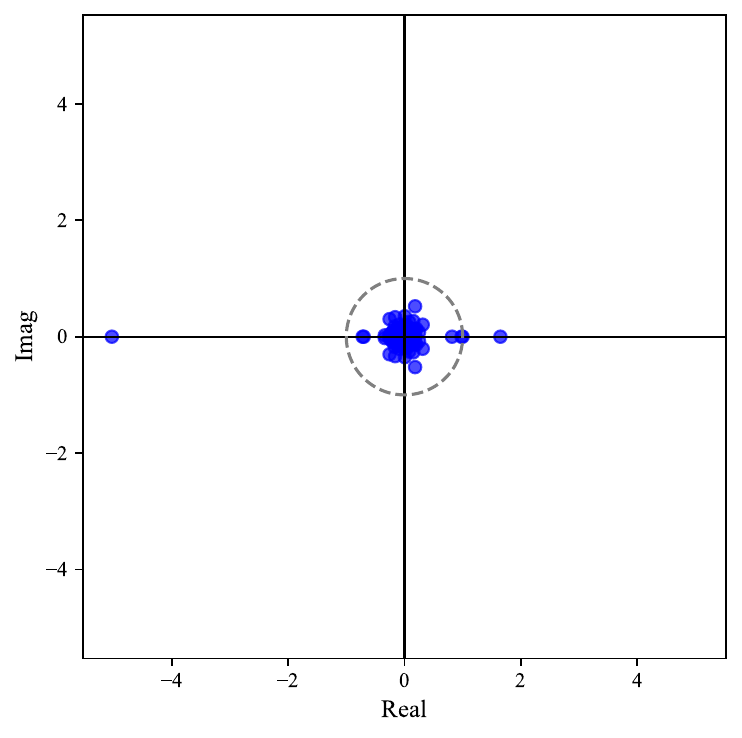}
       \caption{Online DMD}
   \end{subfigure}
   \hspace{2em}
   \begin{subfigure}{0.29\textwidth}
       \centering
       \includegraphics[width=\textwidth]{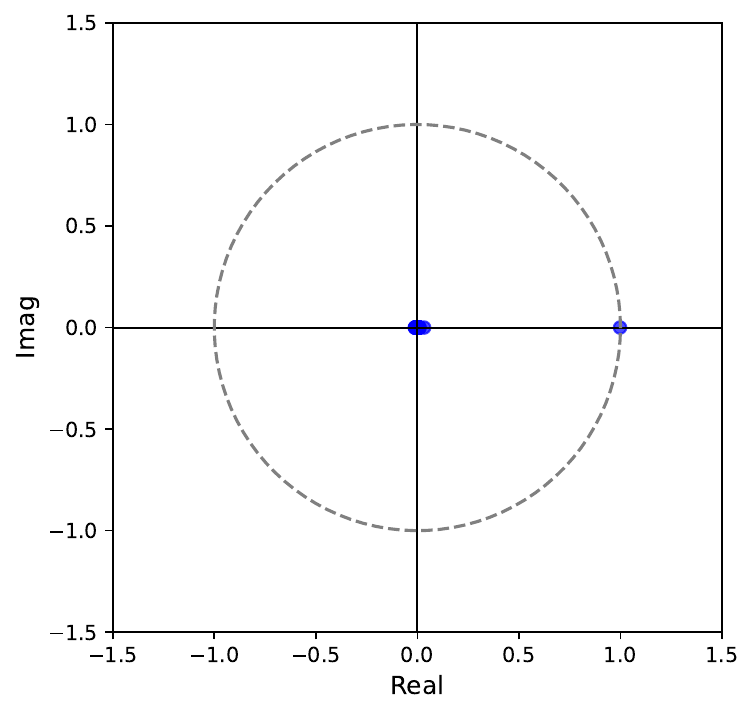}
       \caption{Symmetric OPIDMD}
   \end{subfigure}
   \caption{Eigenvalue Distributions}
   \label{fig:advection2}
 \end{figure}

 The solution, as shown in Figure \ref{fig:convection_diffusion}, displays the system's state over time during the simulation.  We use the first 4850 noisy time steps to train the model, and the noise-free data at step 4850 is used as the initial condition for predicting the final 150 noise-free steps to evaluate model performance. Figures \ref{fig:advection1} and \ref{fig:advection2} show the DMD matrices and the corresponding eigenvalues obtained by different algorithms. 
 
 The OGD method appears to favor a simpler matrix structure. Specifically, it exhibits characteristics similar to a rank-1 DMD matrix, suggesting that this method may prioritize capturing the system’s dominant mode while ignoring more complex dynamics. This tendency allows the OGD method to generate a relatively simple model.
 The OPIDMD algorithm, with symmetric constraints, also retains a similar simplification. This algorithm displays balanced symmetry in the matrix structure, with eigenvalues mostly concentrated at the origin, with only one eigenvalue located on the unit circle.
 
 The matrix structure of the online DMD method appears to be significantly affected by noise. The resulting matrix shows noticeable
  noise interference, making its structure more diffuse and lacking clear patterns. This influence of noise is also reflected in the
   eigenvalue distribution of Online DMD, where some eigenvalues deviate from the unit circle. Such deviations may suggest overfitting
    or instability issues when handling noisy data.
 
 The piDMD method exhibits structural characteristics similar to a rank-2 DMD matrix, though this similarity is not particularly strong. Due to the symmetric constraint, the eigenvalues of the piDMD method are primarily concentrated along the real axis; however, this approach may also capture some spurious modes.
 Given that the time steps in this case are only 4850, the Online DMD method is significantly influenced by noise, resulting in unstable eigenvalues and a peculiar matrix structure.

 \subsection{Five degrees of freedom system}
\begin{figure}[htb]
   \centering
   \includegraphics[width=0.6\linewidth]{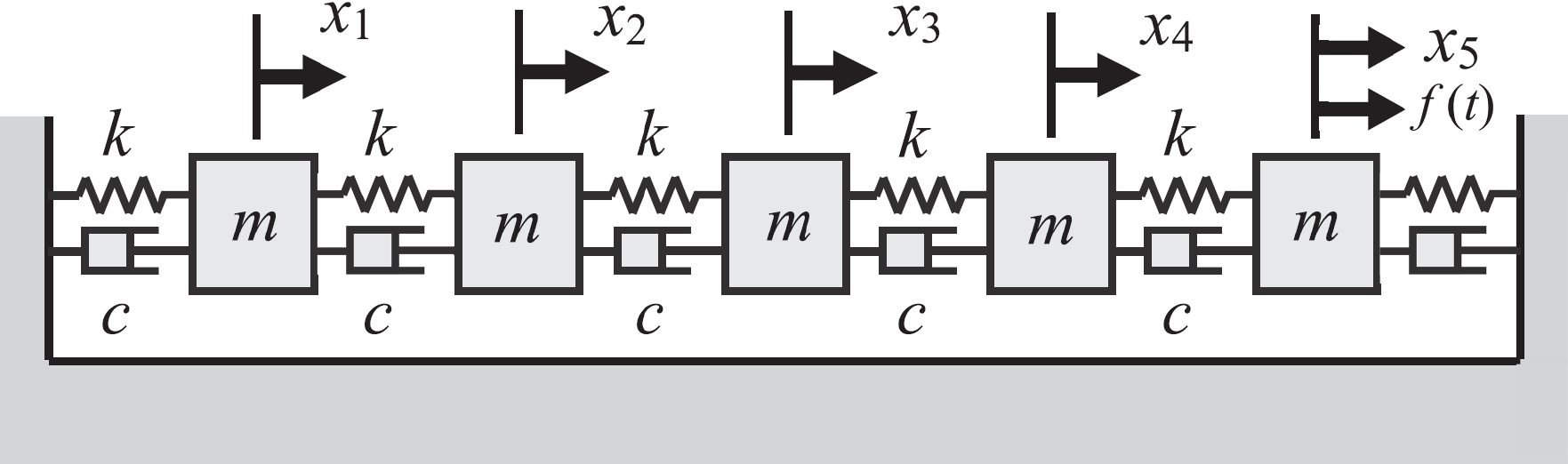}
   \caption{Five degree of freedom system}\label{fig20}
 \end{figure}

 The previous sections provided examples of short-term predictions. This part presents an example where physical information is used to make the OPIDMD matrix structure closer to that of True DMD. In structural dynamics, understanding the behavior of systems subjected to dynamic loads is crucial.
 In this section, a five-degree-of-freedom (DOF) mass-spring-damper system is considered.
 The original second-order differential equation governing the motion is given by:
 \begin{equation}\label{eqc_{28}}
   M\ddot{x} + C\dot{x} + Kx = F(t)
 \end{equation}
 where \(F(t)\) represents external excitation. \( M \) is
  the mass matrix, a diagonal matrix where each element equals \( m \), representing the mass. The stiffness matrix,
   \( K \), is structured to reflect the connectivity and stiffness of each element in the system, given by:
 \begin{equation}
   K = \begin{bmatrix}
     2k & -k & 0 & 0 & 0 \\
     -k & 2k & -k & 0 & 0 \\
     0 & -k & 2k & -k & 0 \\
     0 & 0 & -k & 2k & -k \\
     0 & 0 & 0 & -k & 2k
     \end{bmatrix},
 \end{equation}
 where \( k \) is the stiffness coefficient. The damping matrix, \( C \), is proportionally related to the stiffness 
 matrix, defined as \( C = (c/k)K \), with \( c \) being the damping coefficient.
 For the purposes of numerical example generation, the parameters are set to \( m = 1 \), \( c = 0.06 \), and \( k = 1 \). Assuming that at the initial moment, there is a random excitation at mass 5, the system then undergoes free vibration.The diagram of the five-degree-of-freedom  mass-spring-damper system is presented in Figure \ref{fig20}.
 
 We implement the Newmark-beta numerical method, detailed in \cite{chopra2007dynamics}, to solve the equation denoted as \eqref{eqc_{28}}.
  This method is highly regarded in the field of dynamic analysis for its ability to finely tune the balance between accuracy and numerical 
  stability, utilizing its two main parameters, $\gamma$ and $\beta$.
  We set $\gamma=0.5$ for unconditional stability, ensuring consistent acceleration, and $\beta=0.25$ to introduce numerical damping, 
  allowing a clear acceleration-time relationship and improving the simulation's dynamic response.
 
\begin{figure}[!ht]
   \centering
   \begin{subfigure}{0.45\textwidth}
       \centering
       \includegraphics[width=\textwidth]{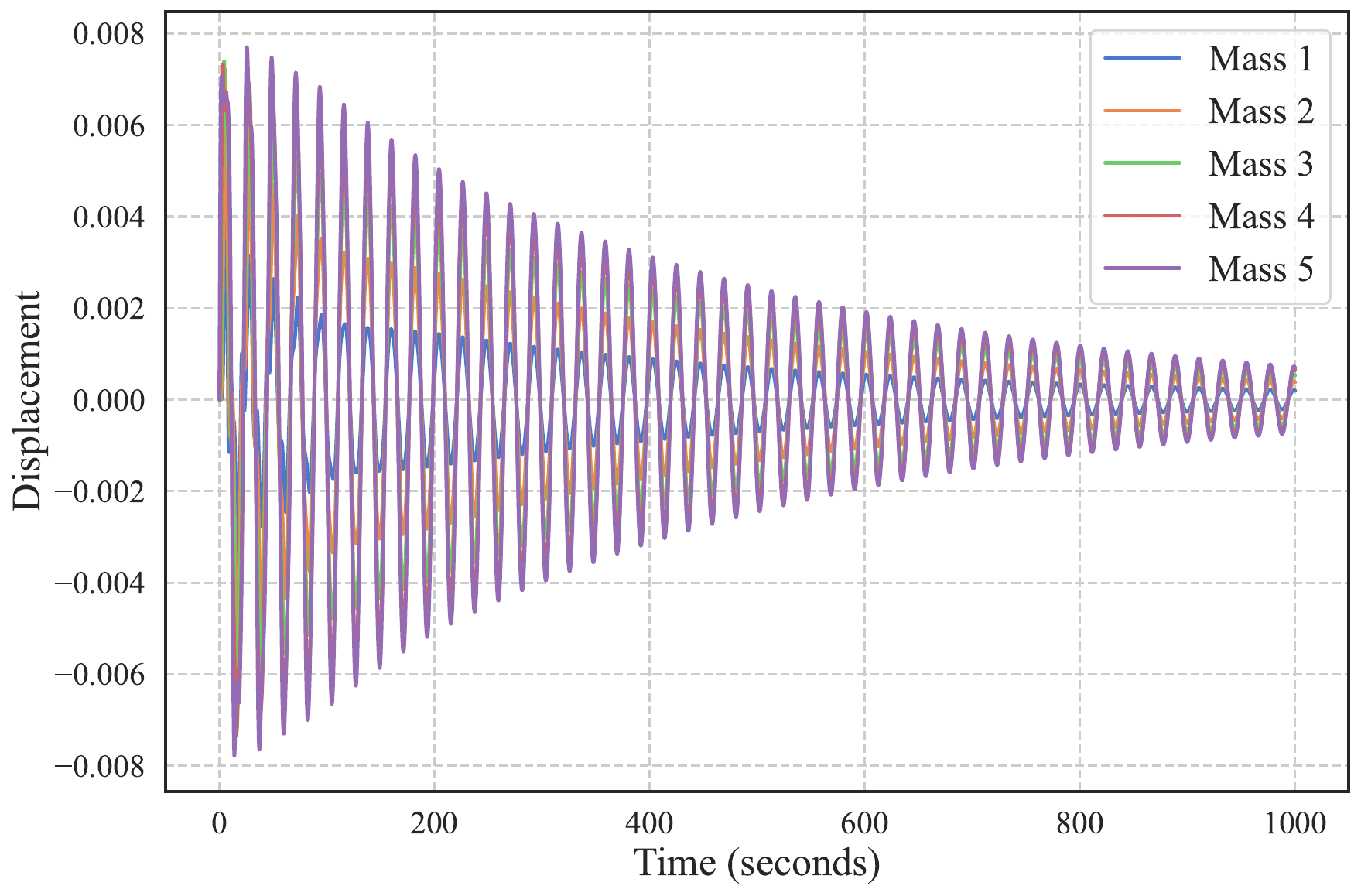}
       \caption{Time response of   displacements}
   \end{subfigure}
   \hspace{1em}
   \begin{subfigure}{0.45\textwidth}
       \centering
       \includegraphics[width=\textwidth]{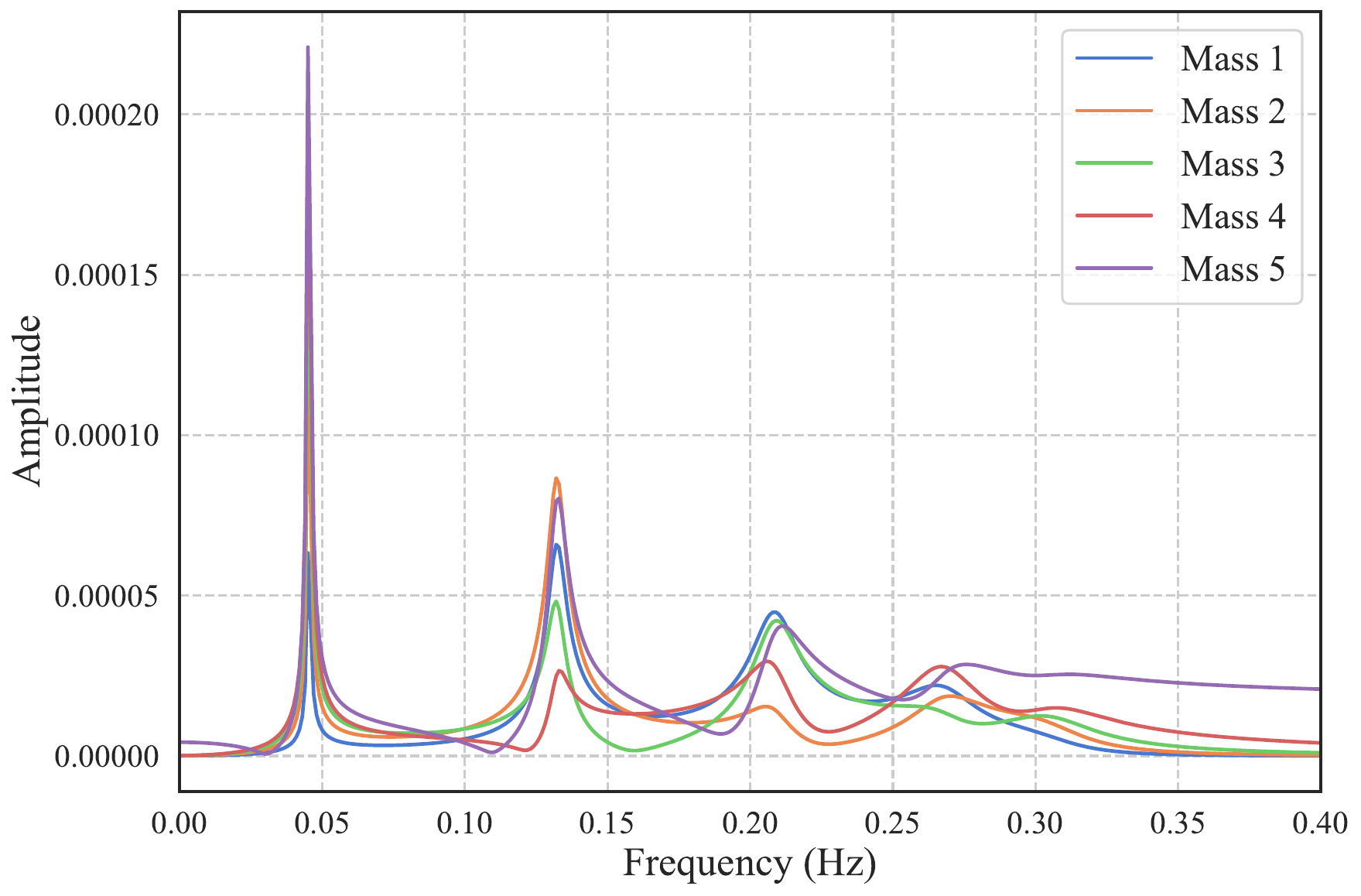}
       \caption{Frequency response of accelerations}
   \end{subfigure}
   \caption{Time response of Five degrees of freedom system}
   \label{fig:12}
 \end{figure}

  In the numerical example, an initial excitation is applied to the fifth node. The analysis proceeds with a time step of 0.005 seconds, 
  covering an extensive duration of 1000 seconds to thoroughly assess the system's dynamic response. The model coordinate time responses, 
  captured over the interval $0 \le t \le 1000s$, are meticulously illustrated in Figure \ref{fig:12}. 
  The figure illustrates the displacement of each mass over time and the frequency spectrum of accelerations for each mass, used to demonstrate the structure's natural frequencies.
 
\begin{figure}[!ht]
   \centering
   \begin{subfigure}{0.29\textwidth}
       \centering
       \includegraphics[width=\textwidth]{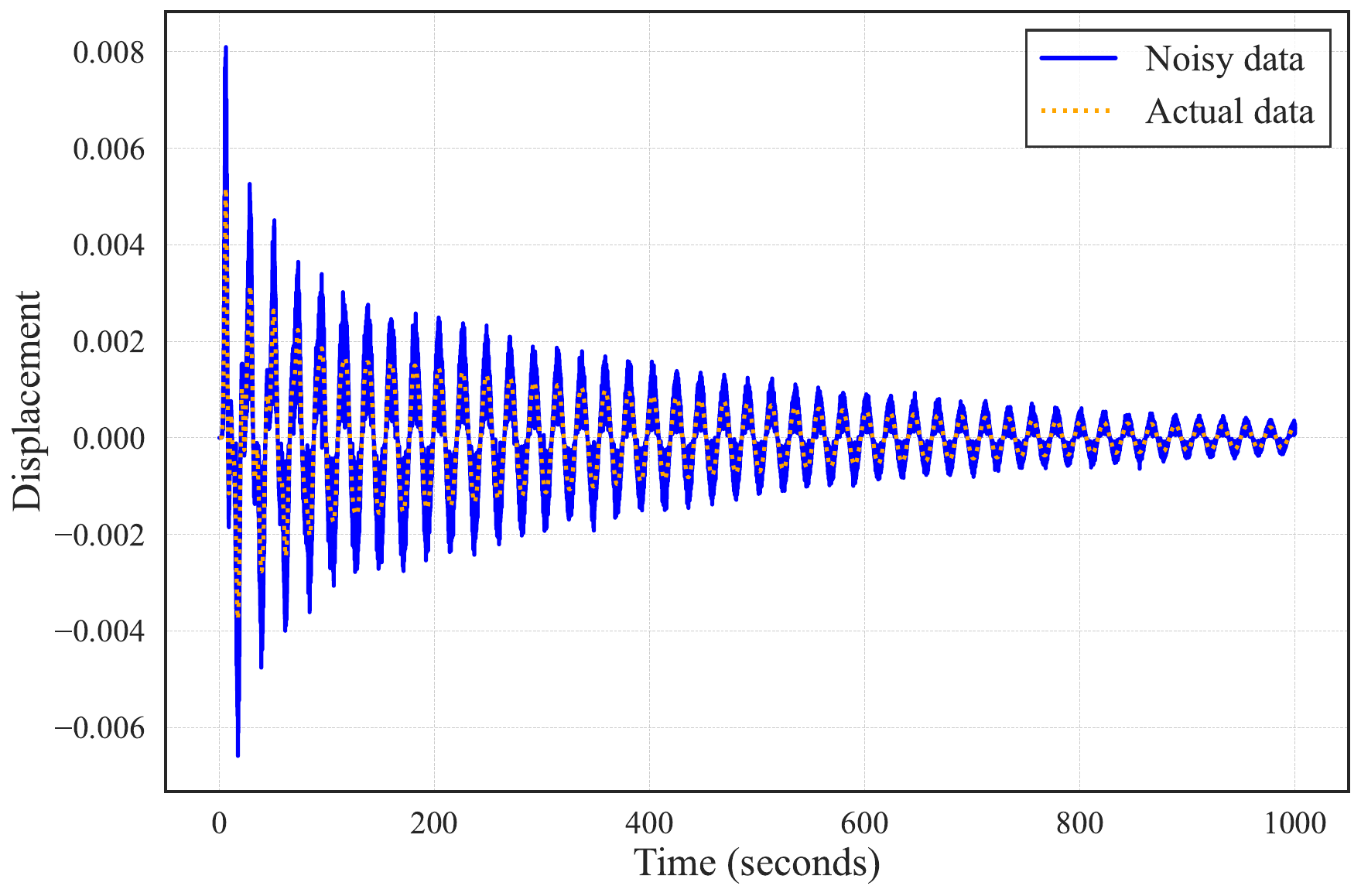}
       \caption{ $x_1(t)$ Displacement}
   \end{subfigure}
   \hspace{2em}
   \begin{subfigure}{0.29\textwidth}
       \centering
       \includegraphics[width=\textwidth]{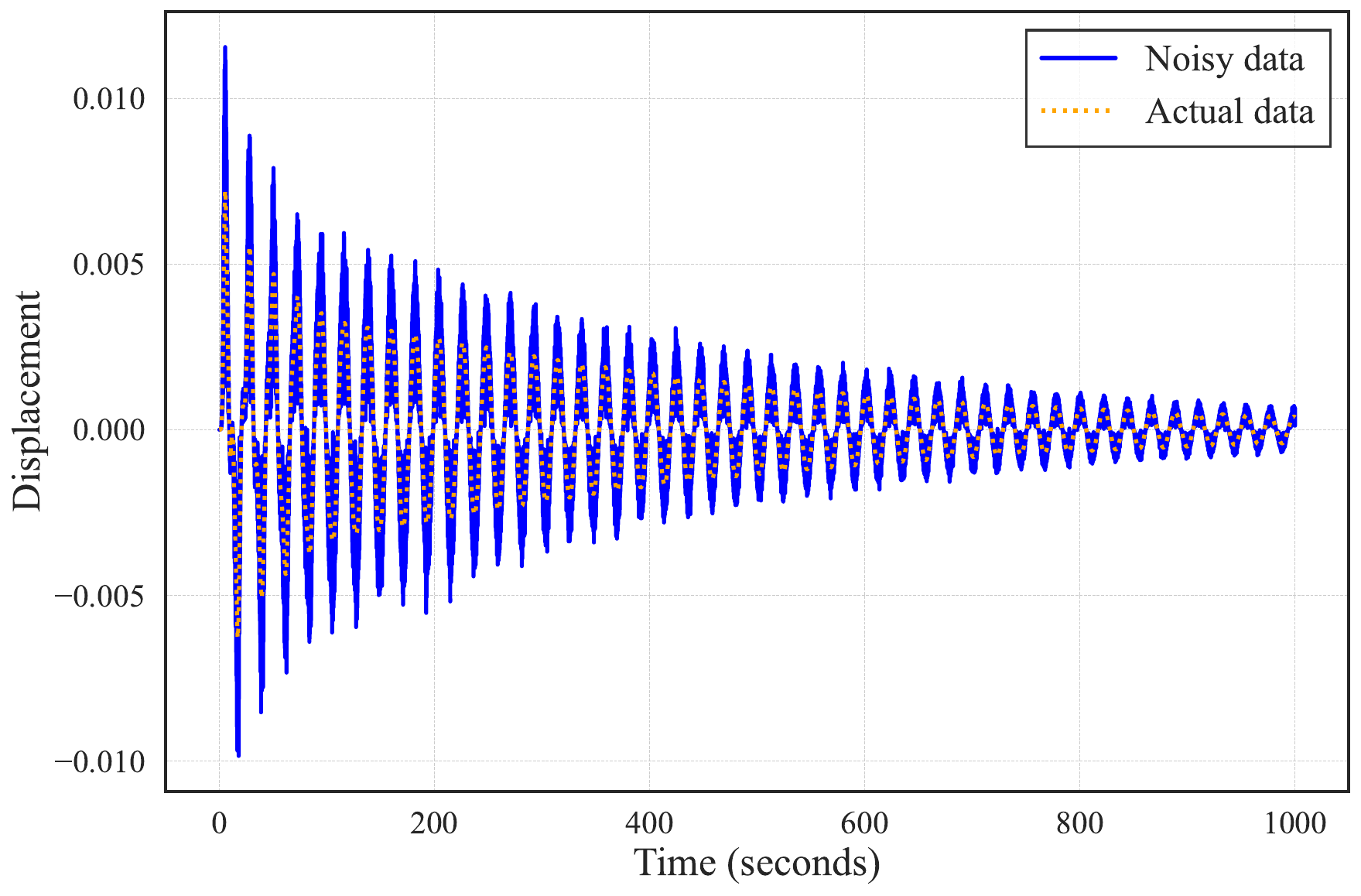}
       \caption{$x_2(t)$ Displacement}
     \end{subfigure}
     \hspace{2em}
     \begin{subfigure}{0.29\textwidth}
         \centering
         \includegraphics[width=\textwidth]{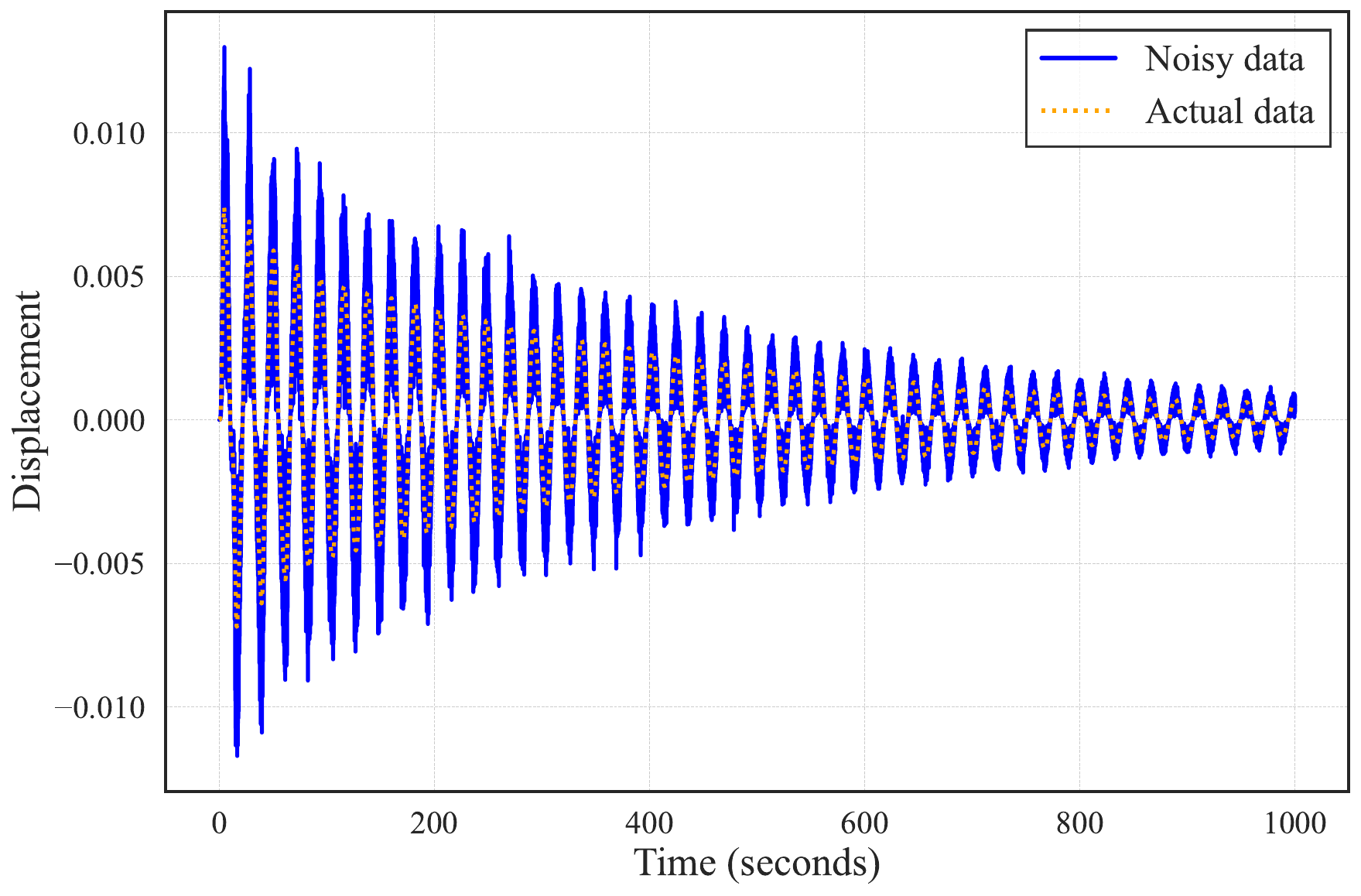}
         \caption{$x_3(t)$ Displacement}
   \end{subfigure}
 
   \vspace{1em}
   \begin{subfigure}{0.29\textwidth}
       \centering
       \includegraphics[width=\textwidth]{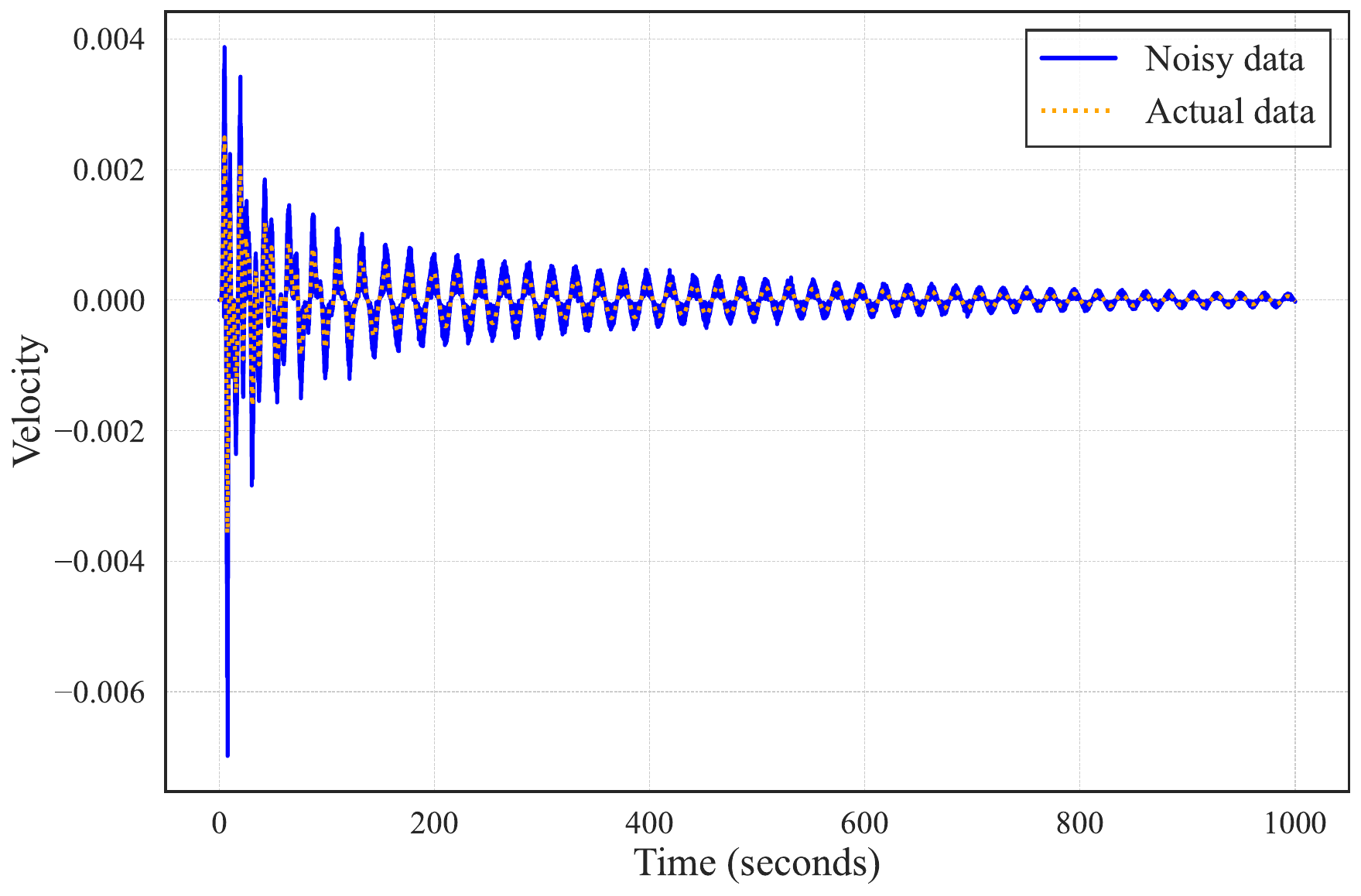}
       \caption{ $ \dot{x}_1(t)$ Velocity}
   \end{subfigure}
   \hspace{2em}
   \begin{subfigure}{0.29\textwidth}
       \centering
       \includegraphics[width=\textwidth]{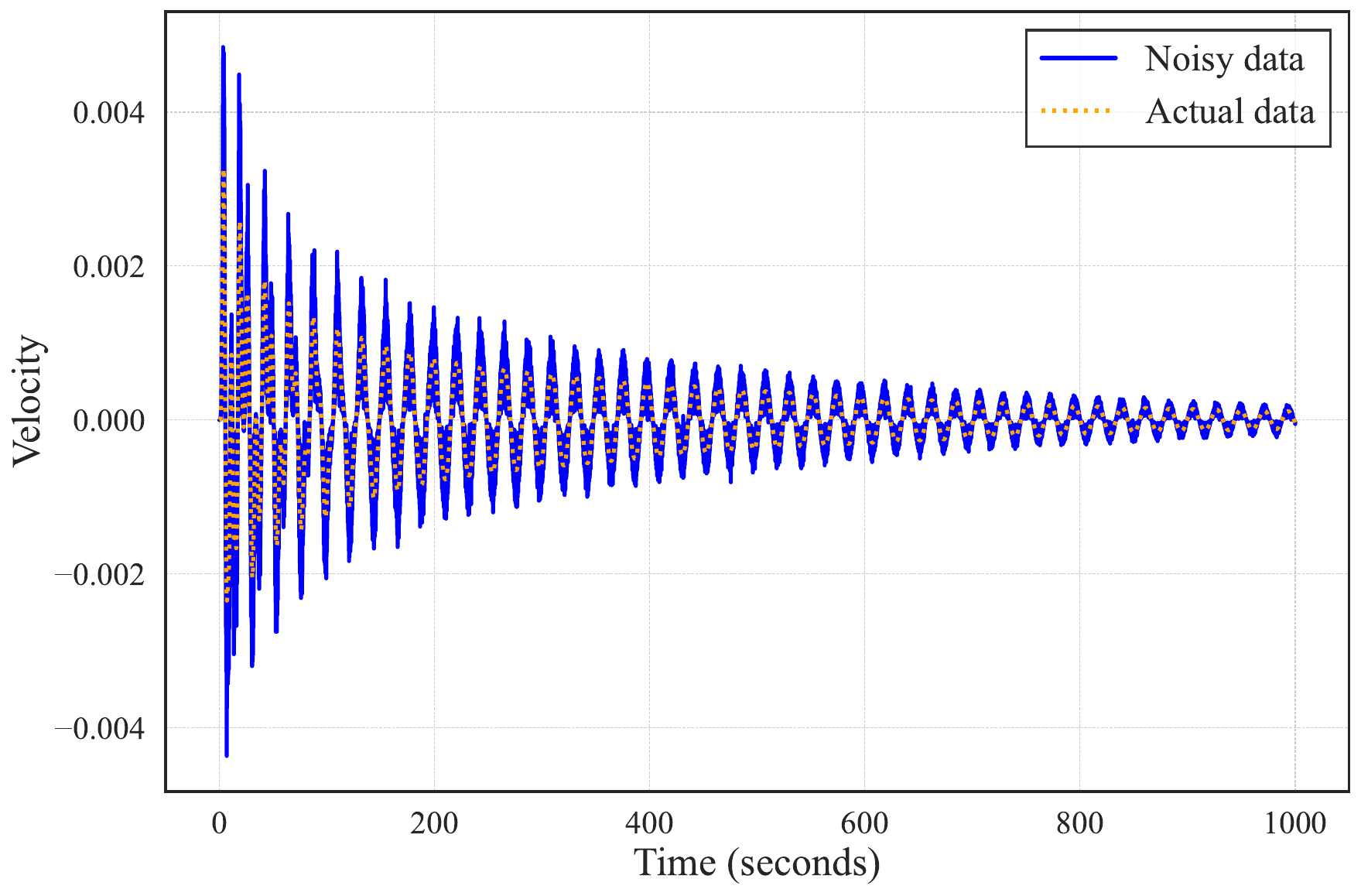}
       \caption{$ \dot{x}_2(t)$  Velocity}
   \end{subfigure}
   \hspace{2em}
   \begin{subfigure}{0.29\textwidth}
     \centering
     \includegraphics[width=\textwidth]{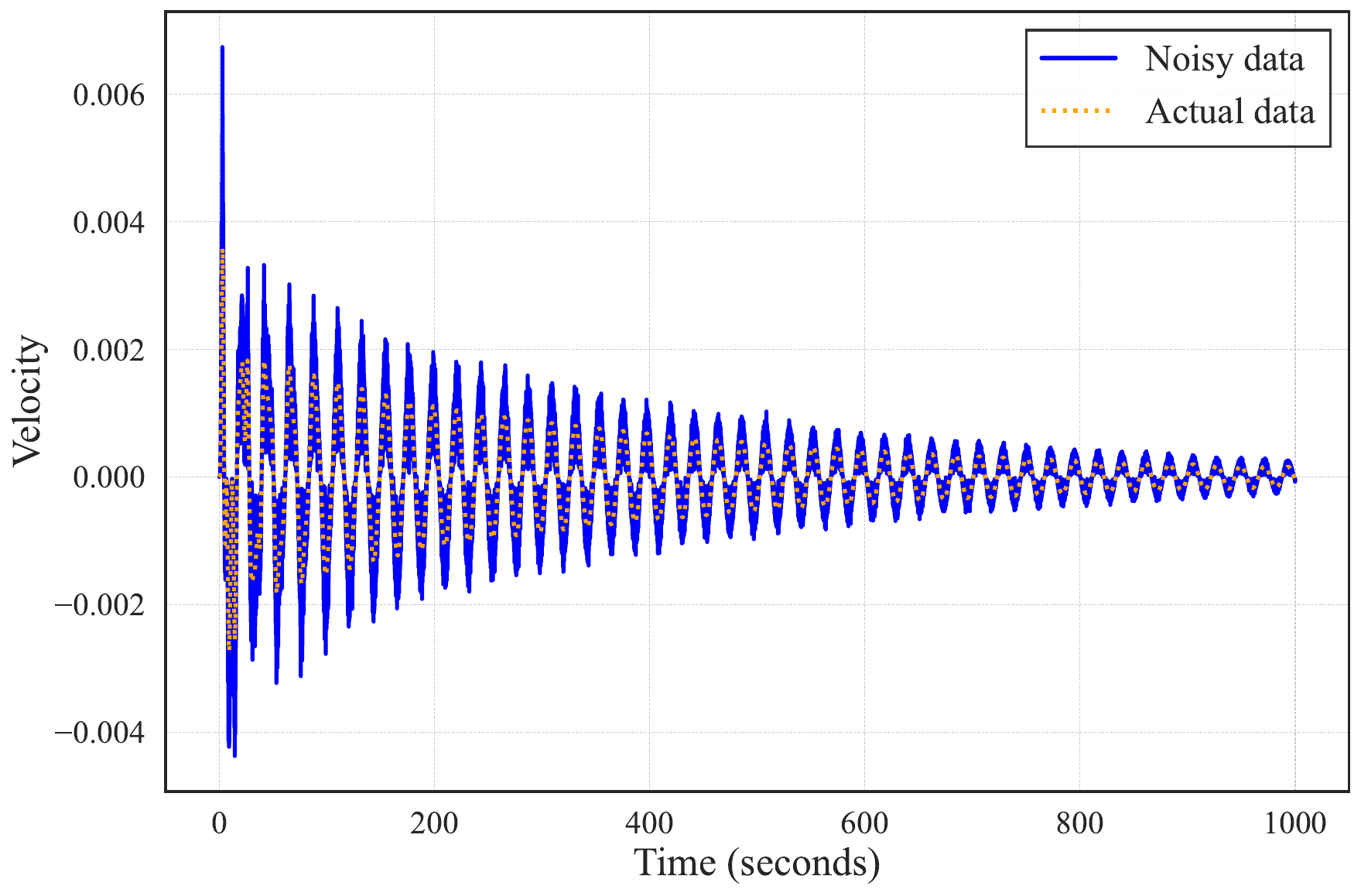}
     \caption{$ \dot{x}_3(t)$  Velocity}
 \end{subfigure}
 \caption{Time response of physical DOF $x(t)$ with noise}
   \label{fig:13}
 \end{figure}

 The analysis incorporates white noise proportional to signal amplitude, with each data point increased by noise amounting to 25\% of its amplitude. 
 Figure \ref{fig:13} illustrates the impact of this noise addition on the first three mass blocks, with the approach to noise addition for remaining mass blocks being analogous.
 The examined physical system does not strictly match the constraints specified in piDMD\cite{baddooPhysicsinformedDynamicMode2023} or adhere to the matrix
  structure assumptions addressed earlier in this paper. Nonetheless, it serves as an accurate example of a linear shift-invariant causal system, 
  aligning with the system categories of causal and shift-invariant as identified in the \cite{baddooPhysicsinformedDynamicMode2023}. Subsequent examples demonstrate that by implementing stringent physics-based regularization, the impact of noise on the dataset can be effectively reduced. This ensures that the analysis maintains its integrity and relevance, even in the presence of significant noise interference.

 In the numerical solution of this example, a time step of 0.005 seconds was used, covering a duration of 1000 seconds and generating a total of 200,000 
 data points. Each data point includes displacement and velocity information, represented as \( z_k \in \mathbb{R}^{10} \) for \( k = 1 \) to 200,000. 
 However, given the low dimensionality of the matrix in this case, with only 100 parameters, testing showed that the OPIDMD algorithm, with the addition
  of physical constraints, achieves satisfactory results using only the first 800 data points. Therefore, only the initial 800 data points were used in
   the paper to train the model.
 
\begin{figure}[!ht]
     \centering
     \begin{subfigure}{0.29\textwidth}
         \centering
         \includegraphics[width=\textwidth]{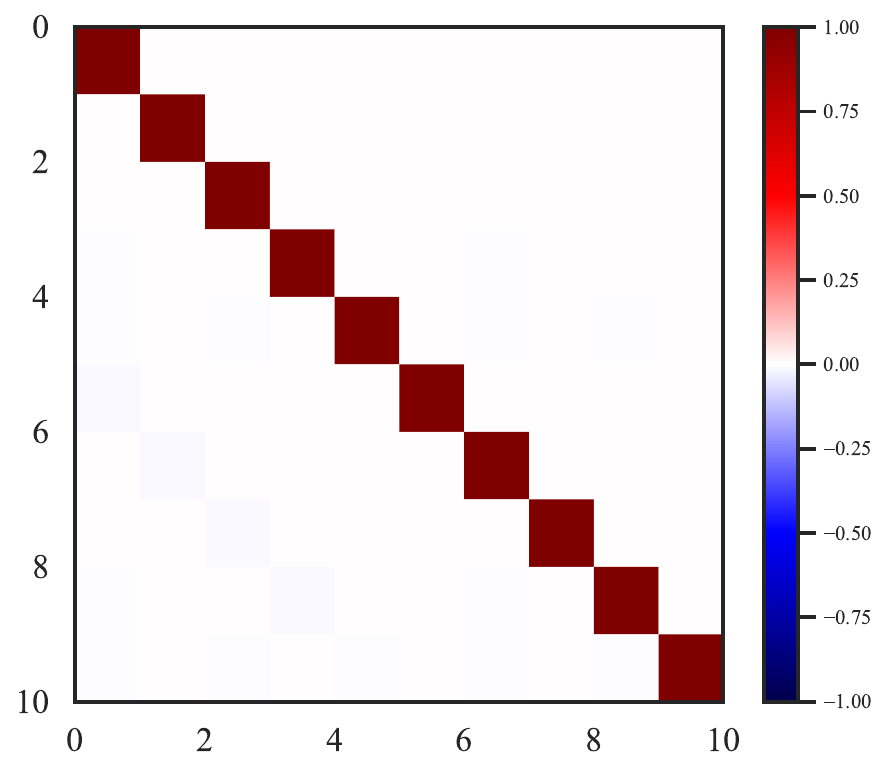}
         \caption{Original DMD}
     \end{subfigure}
     \hspace{2em}
     \begin{subfigure}{0.29\textwidth}
         \centering
         \includegraphics[width=\textwidth]{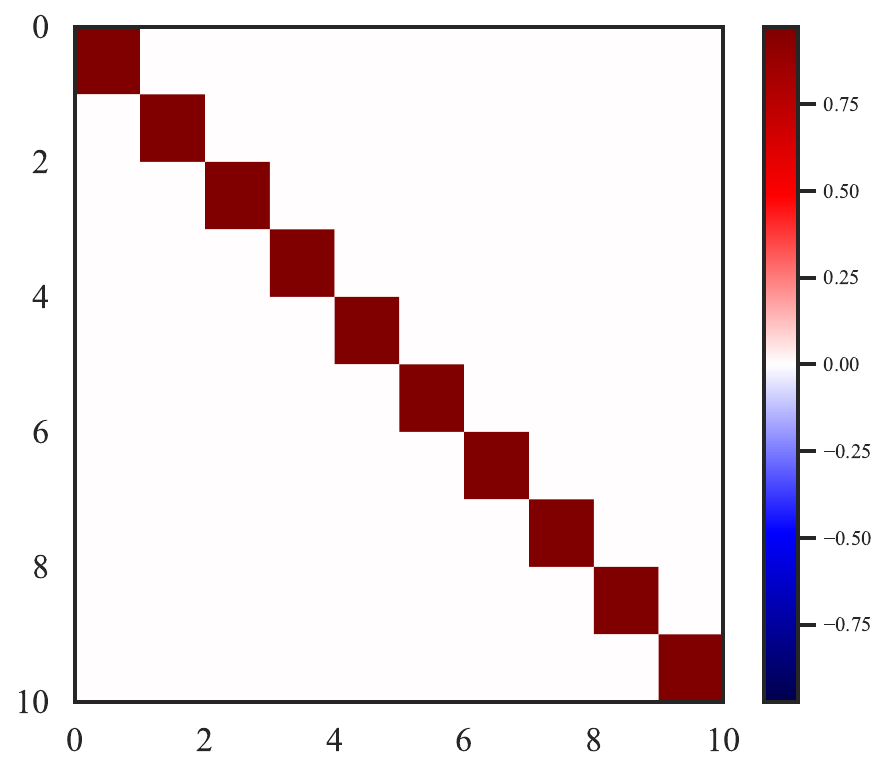}
         \caption{OPIDMD }
       \end{subfigure}
       \hspace{2em}
       \begin{subfigure}{0.29\textwidth}
           \centering
           \includegraphics[width=\textwidth]{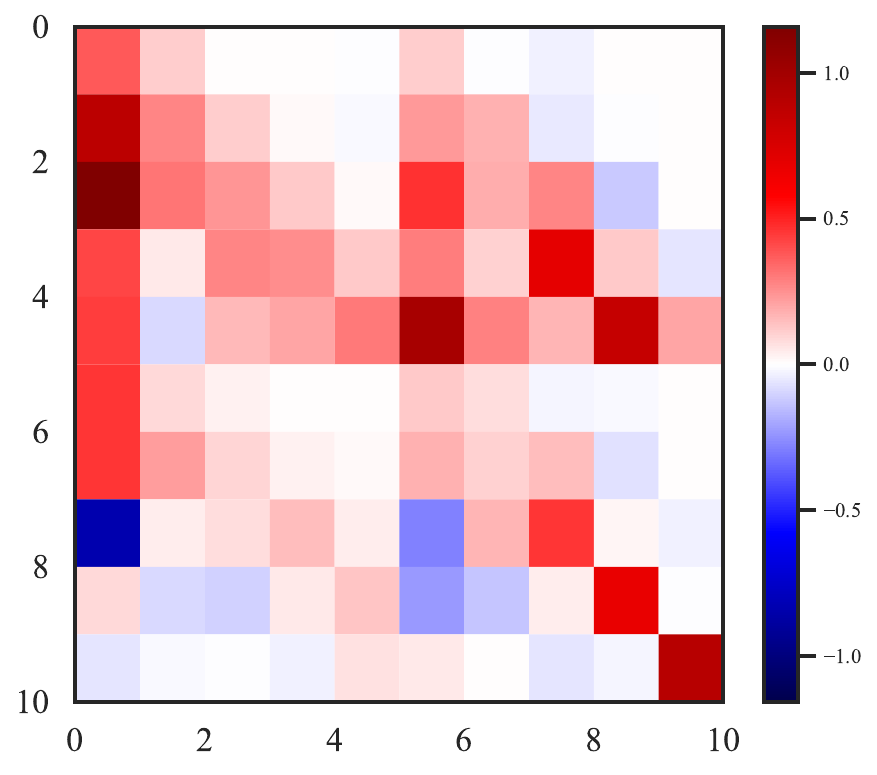}
           \caption{Online DMD}
     \end{subfigure}
   
     \vspace{1em}
     \begin{subfigure}{0.29\textwidth}
         \centering
         \includegraphics[width=\textwidth]{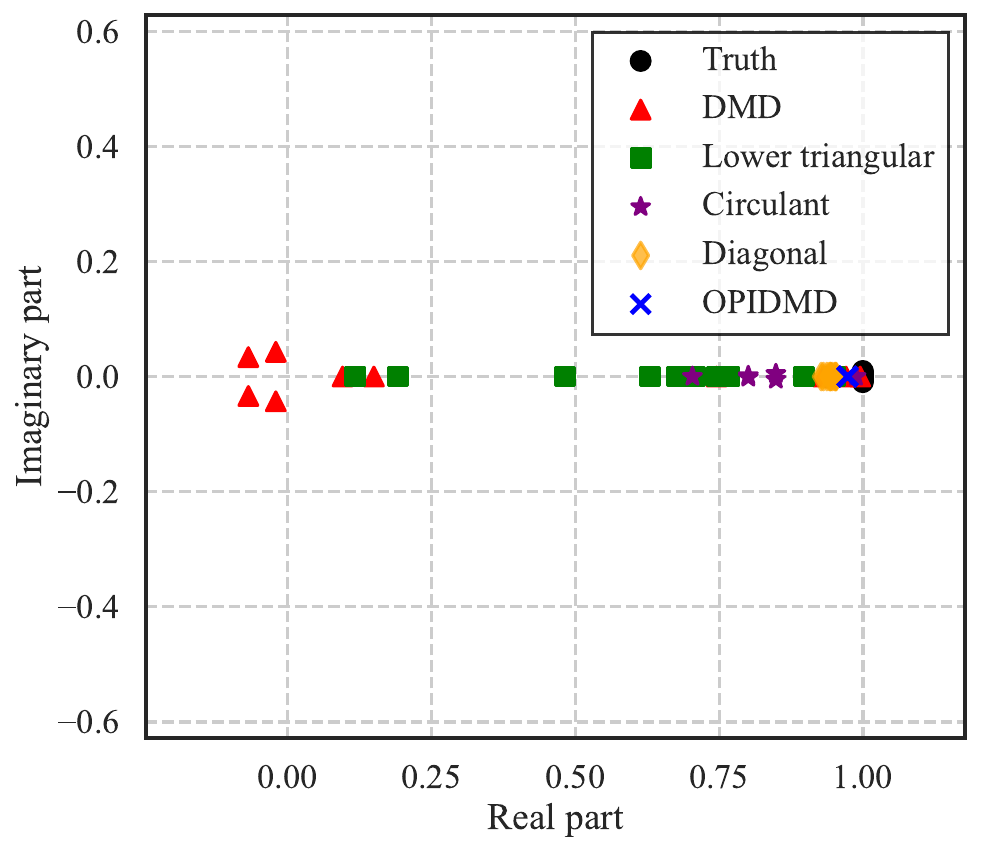}
         \caption{Eigenvalue Spectrum}
     \end{subfigure}
     \hspace{2em}
     \begin{subfigure}{0.29\textwidth}
         \centering
         \includegraphics[width=\textwidth]{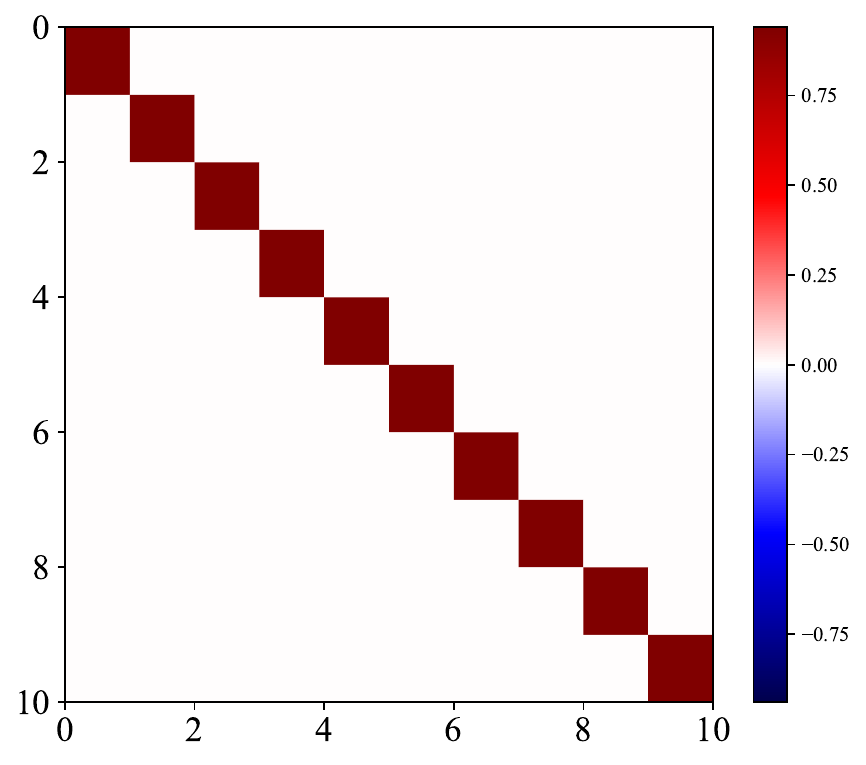}
         \caption{OPIDMD  Final}
     \end{subfigure}
     \hspace{2em}
     \begin{subfigure}{0.29\textwidth}
       \centering
       \includegraphics[width=\textwidth]{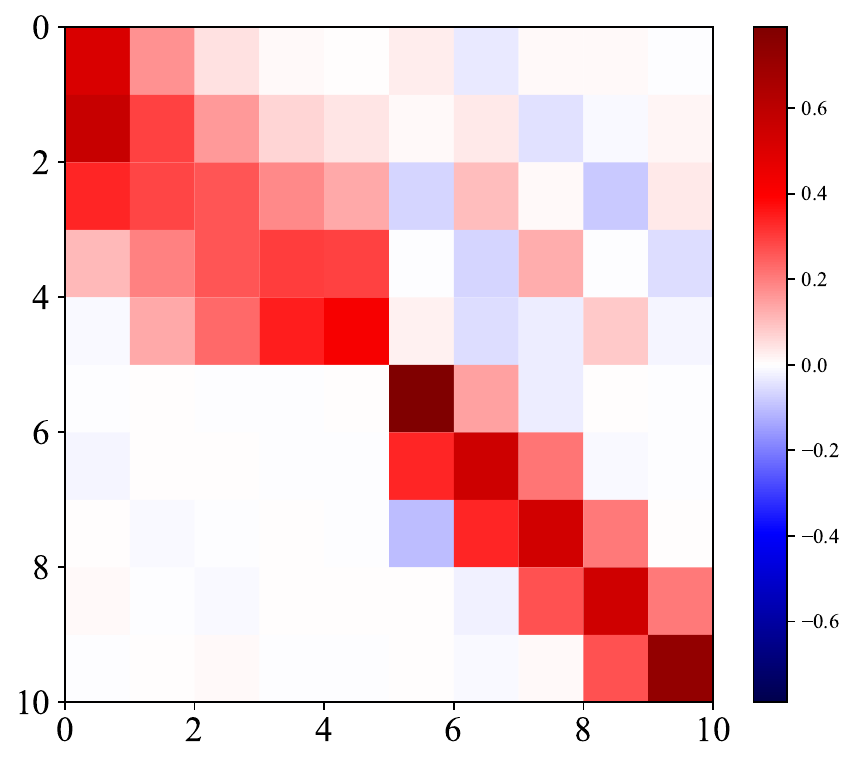}
       \caption{Online DMD Final}
   \end{subfigure}
   \caption{Comparison of DMD Matrices and Spectral Analysis}
     \label{fig:15}
   \end{figure}

 By fine-tuning the learning rate, OPIDMD achieves effective reconstruction with just 800 data points. We enhance the input data by concatenating displacement
  and velocity vectors to form \( z = [x, \dot{x}]^T \), which serves as the column vectors of the matrix \( X_{800} \) in the DMD algorithm. This allows 
  us to accurately determine the system matrix \( A_{800} \) using the OPIDMD algorithm. The entire dataset of 200,000 samples was used to validate the algorithm's effectiveness. Figure \ref{fig:15}(a) illustrates the true DMD matrix
  derived using the standard DMD algorithm on noise-free data. In contrast, \ref{fig:15} (b) and (c)  present the matrices obtained from
   OPIDMD and Online DMD, respectively, both generated from noisy data. 
 
\begin{figure}[!ht]
    \centering
    \begin{subfigure}{0.29\textwidth}
        \centering
        \includegraphics[width=\textwidth]{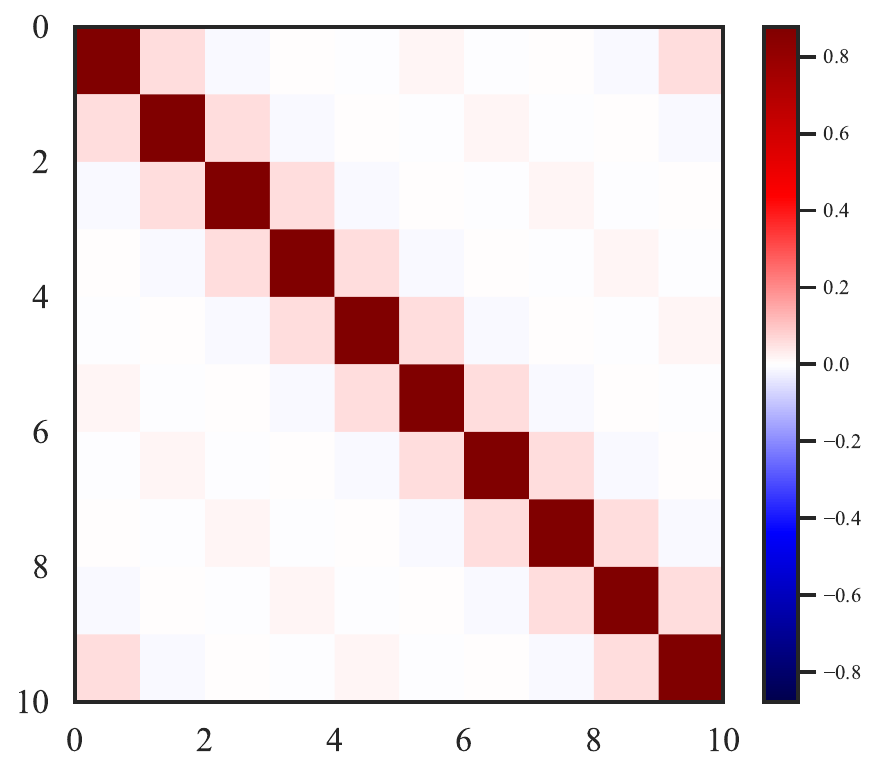}
        \caption{Circulant }
    \end{subfigure}
    \hspace{2em}
    \begin{subfigure}{0.29\textwidth}
        \centering
        \includegraphics[width=\textwidth]{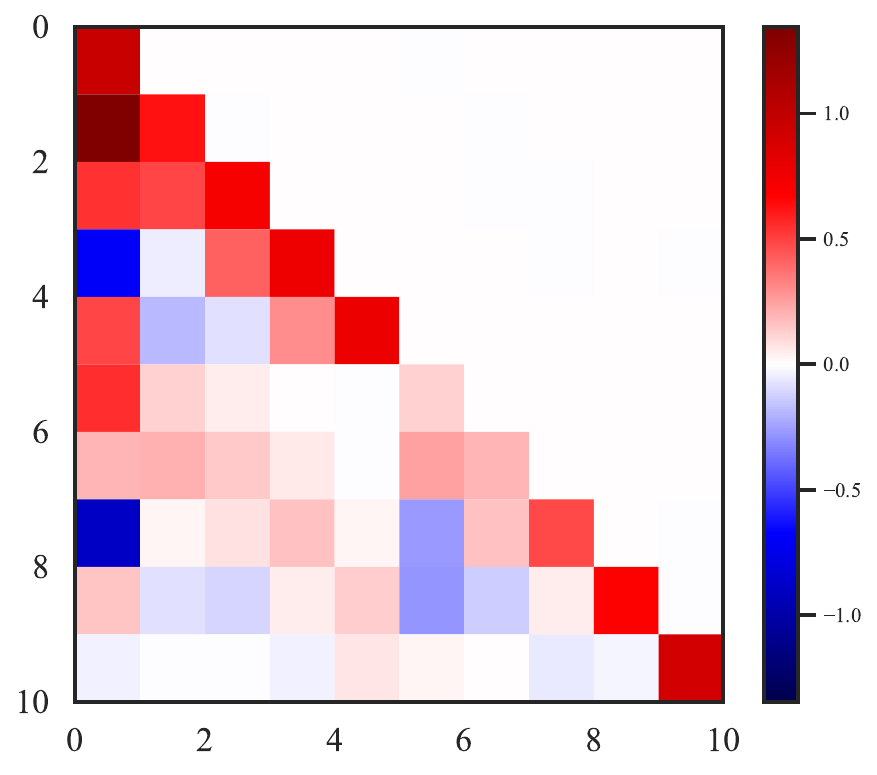}
        \caption{Lower Triangular }
      \end{subfigure}
      \hspace{2em}
      \begin{subfigure}{0.29\textwidth}
          \centering
          \includegraphics[width=\textwidth]{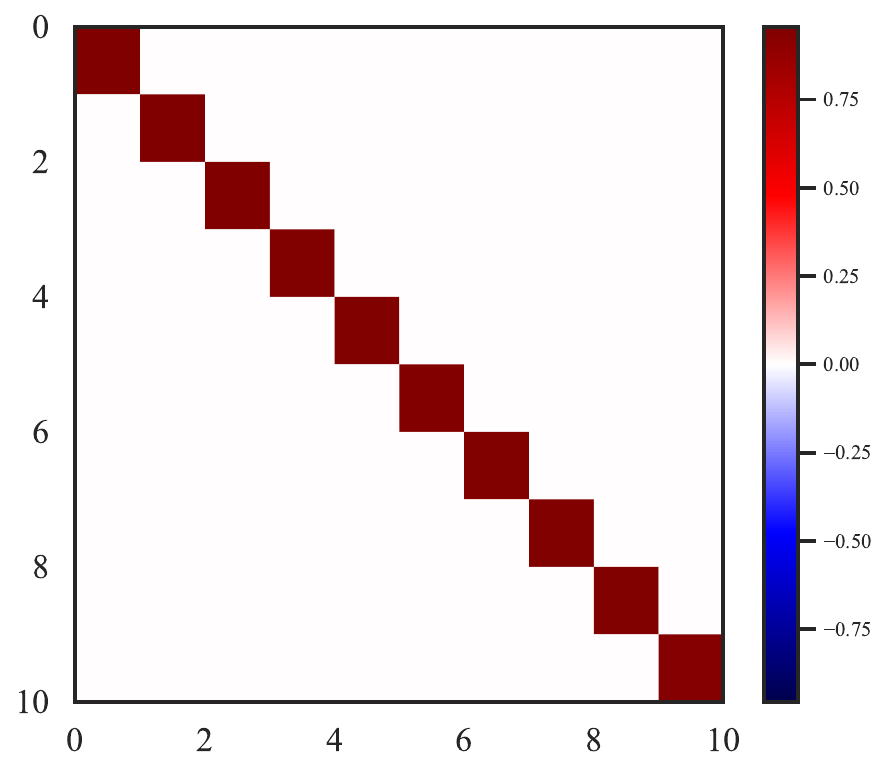}
          \caption{Diagonal }
    \end{subfigure}
    \caption{Visualizations of Different Matrix Constraints Used in piDMD}
    \label{fig:17}
  \end{figure}

 The eigenvalue comparison in Figure \ref{fig:15} (d) is based on the initial noisy 800 data sets, with findings consistent across subsequent full dataset 
 analysis. These visual comparisons highlight that the matrices from OPIDMD more closely mirror the actual situation compared to those from Online DMD,
  indicating the latter's increased sensitivity to noise. Figure \ref{fig:15} (e) and (f) represent the models trained with the full noisy dataset 
  using OPIDMD and Online DMD, respectively. 
  Additionally, using the same 800 noisy data sets, Figure \ref{fig:17} presents visualizations 
  of matrices in piDMD with constraints on circulant matrices, lower triangular matrices, and diagonal matrices, further exploring different structural 
  influences within the same dataset.
 
\begin{figure}[!ht]
   \centering
   \begin{subfigure}{0.29\textwidth}
       \centering
       \includegraphics[width=\textwidth]{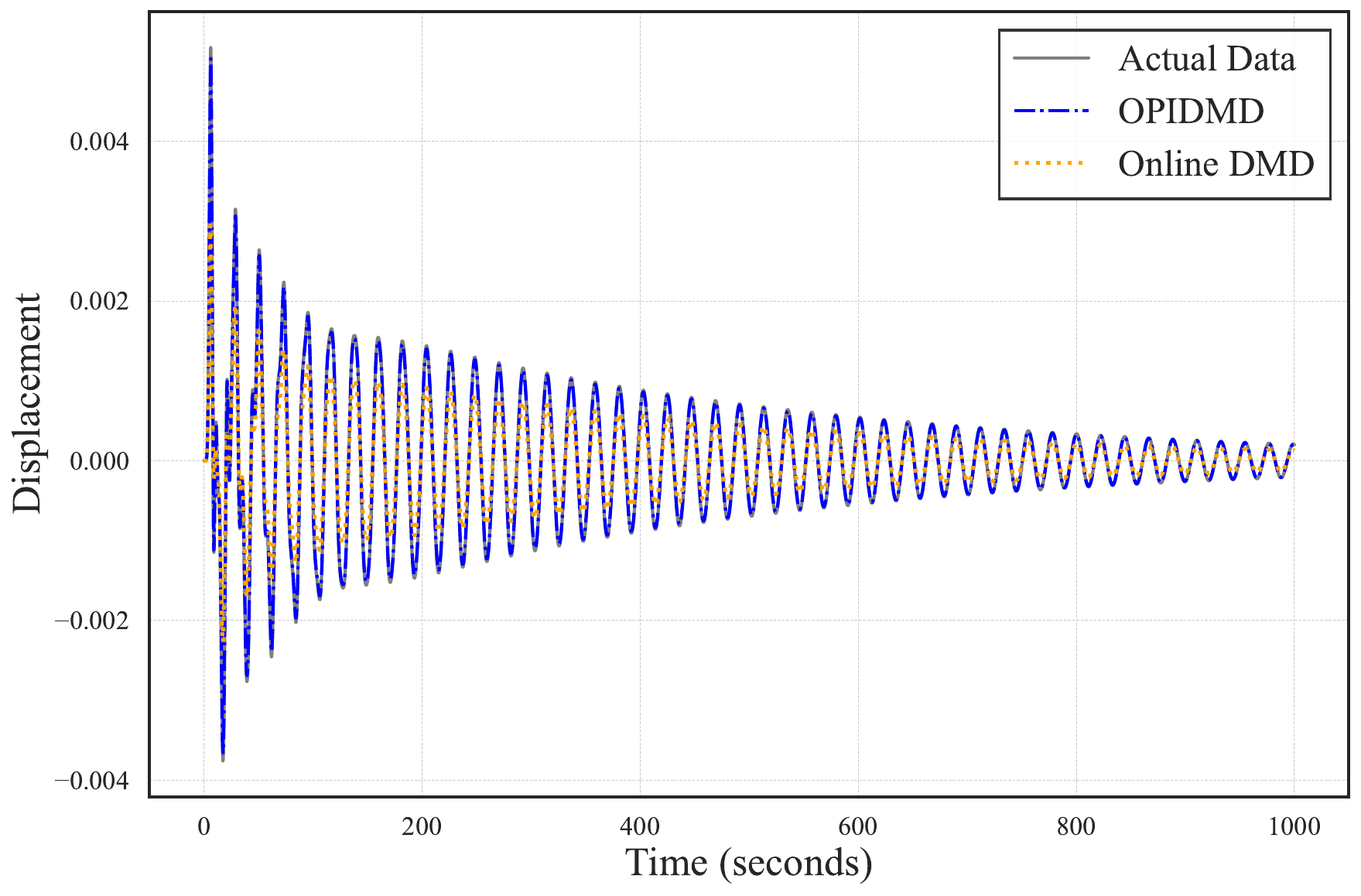}
       \caption{\( x_1(t) \) Displacement}
   \end{subfigure}
   \hspace{2em}
   \begin{subfigure}{0.29\textwidth}
       \centering
       \includegraphics[width=\textwidth]{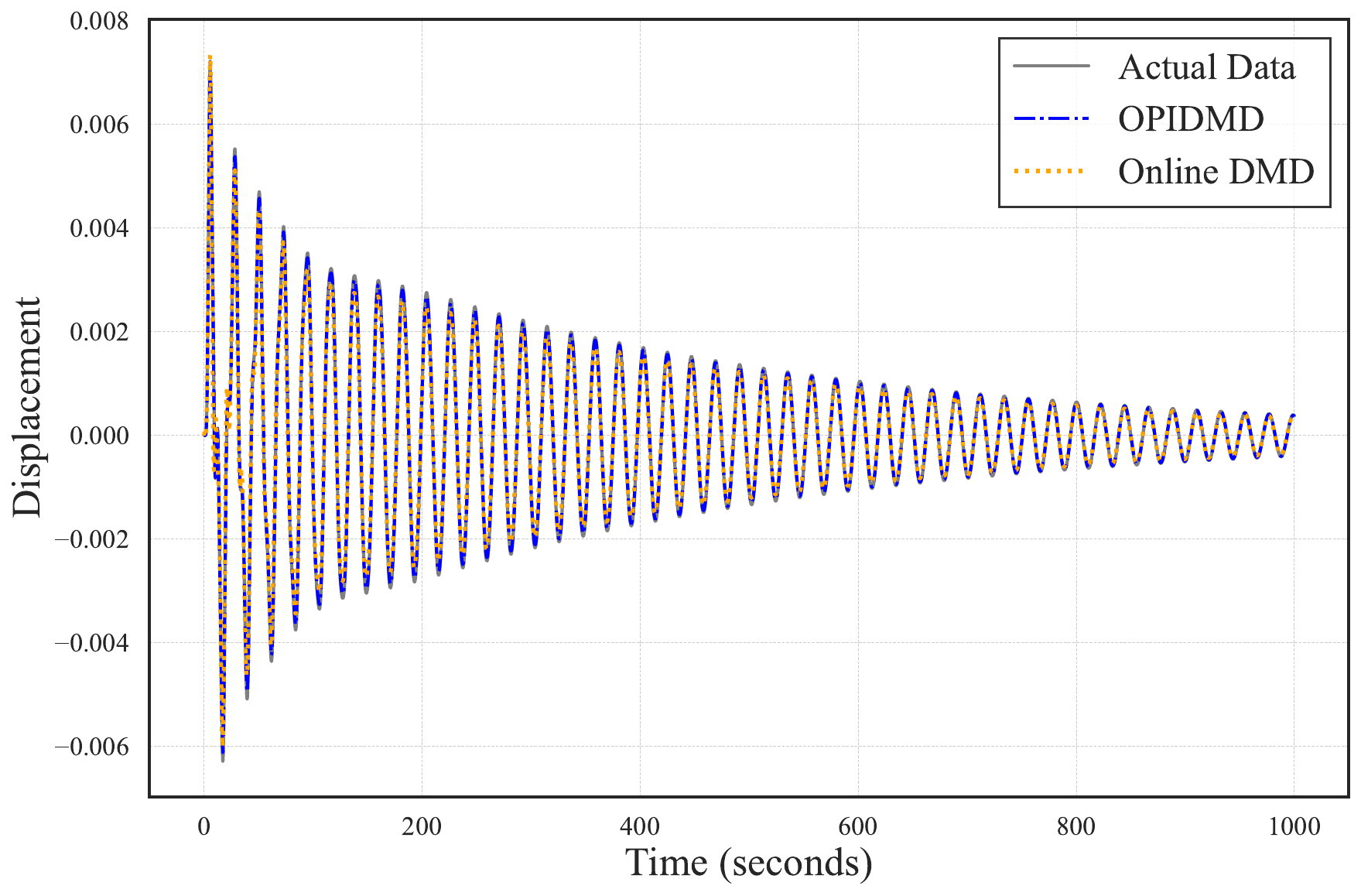}
       \caption{\( x_2(t) \) Displacement}
     \end{subfigure}
     \hspace{2em}
     \begin{subfigure}{0.29\textwidth}
         \centering
         \includegraphics[width=\textwidth]{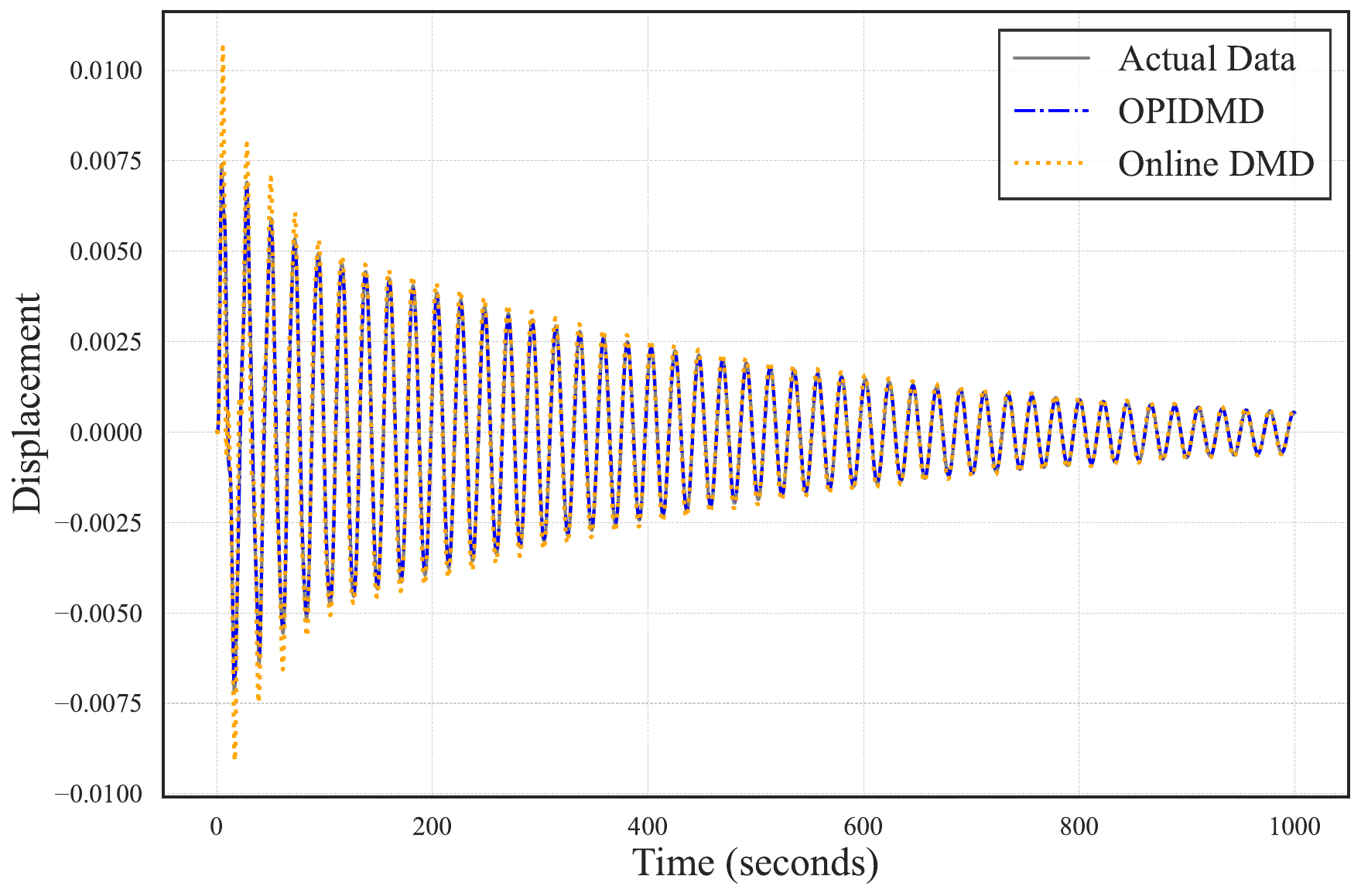}
         \caption{\( x_3(t) \) Displacement}
   \end{subfigure}
 
   \vspace{1em}
   \begin{subfigure}{0.29\textwidth}
       \centering
       \includegraphics[width=\textwidth]{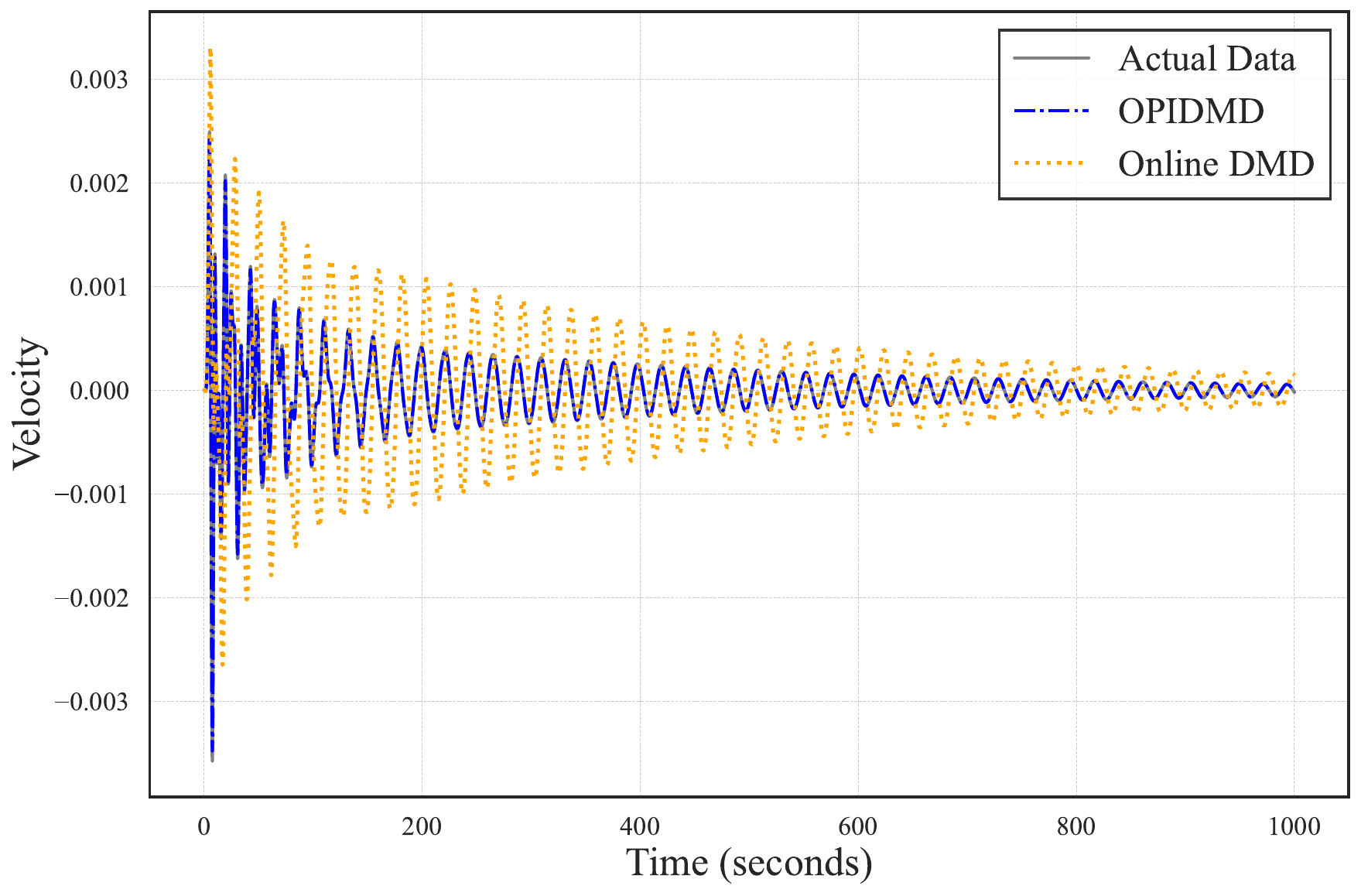}
       \caption{$ \dot{x}_1(t)$  Velocity}
   \end{subfigure}
   \hspace{2em}
   \begin{subfigure}{0.29\textwidth}
       \centering
       \includegraphics[width=\textwidth]{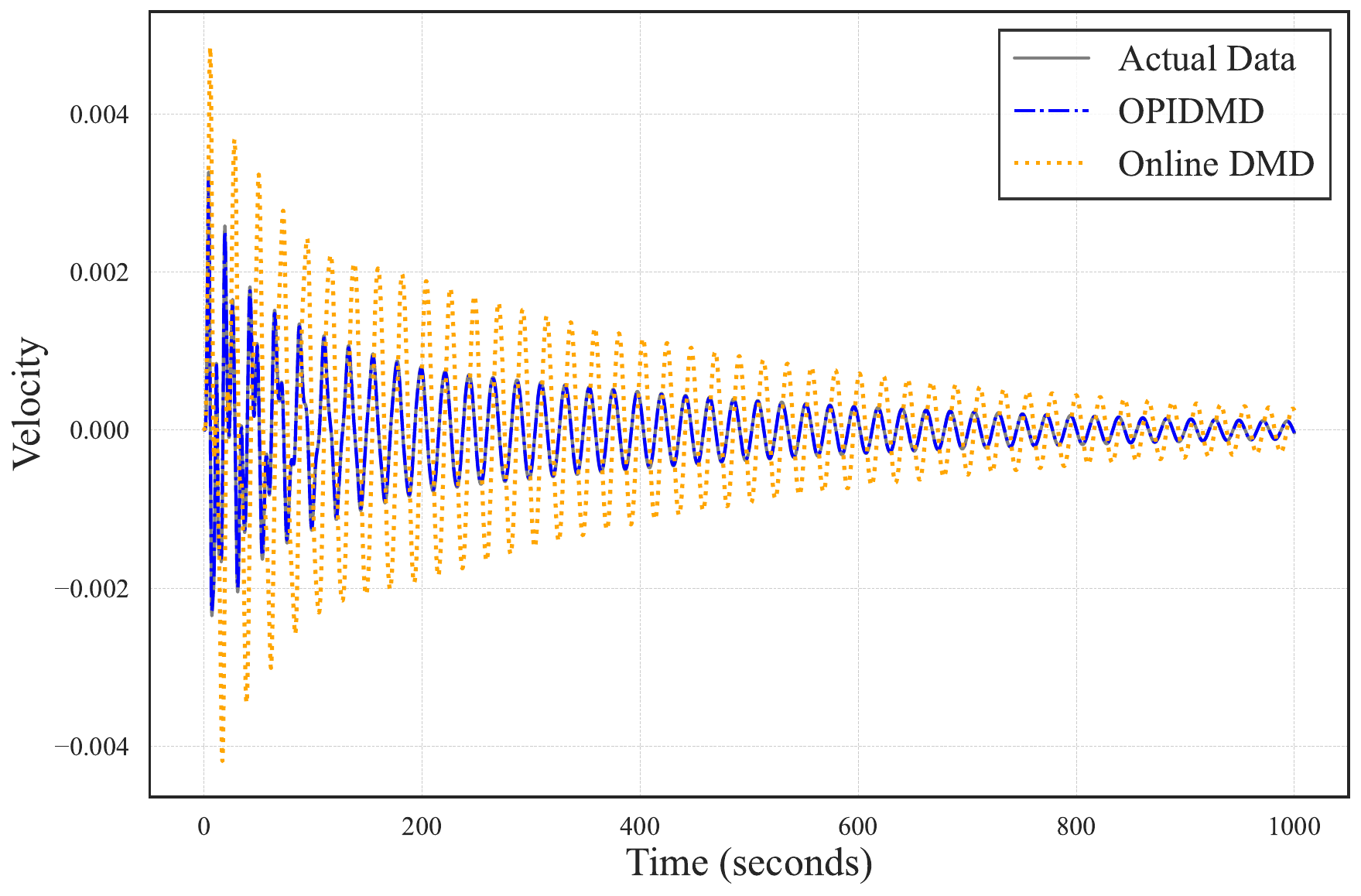}
       \caption{$ \dot{x}_2(t)$ Velocity}
   \end{subfigure}
   \hspace{2em}
   \begin{subfigure}{0.29\textwidth}
     \centering
     \includegraphics[width=\textwidth]{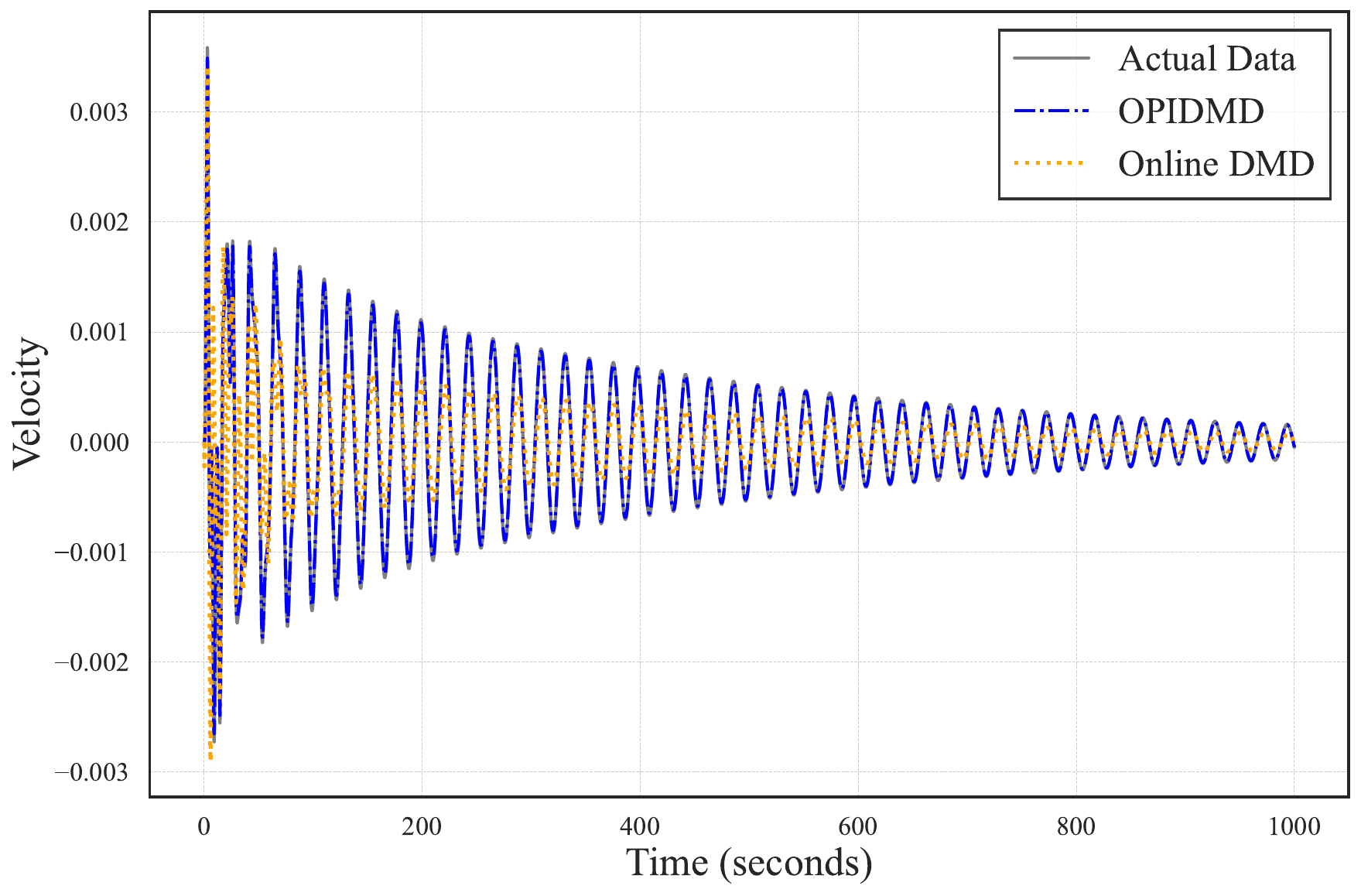}
     \caption{$ \dot{x}_3(t)$ Velocity}
 \end{subfigure}
 \caption{Comparative Displacement and Velocity Responses of DOF \( x(t) \) }
   \label{fig:14}
 \end{figure}

 The initial reconstruction outcomes from noisy 800 data sets are depicted in Figure \ref{fig:14}, where OPIDMD's performance notably exceeds that 
 of online DMD. OPIDMD's advantage is largely due to the integration of physical constraints, which enhance its resistance to noise and accelerate 
 its convergence speed. At this stage, OPIDMD achieves an \(R^2\) score of 0.998, significantly surpassing the 0.194 score obtained by online DMD. 
  Since piDMD is not an online algorithm, its reconstruction effects are not depicted. Instead, the \(R^2\) scores for reconstructions using constraints on 
  circulant matrices, lower triangular matrices, and diagonal matrices are calculated as 0.761, 0.992, and 0.992, respectively. These results are displayed 
  in Figure \ref{fig:17}.
  
  At first glance, the superior reconstruction results of OPIDMD compared to piDMD may seem surprising, given that piDMD can utilize all noisy 800 data sets 
  simultaneously, while OPIDMD processes data sequentially. This difference primarily arises from the unique mechanisms inherent in each algorithm. By 
  finely adjusting the learning rate, OPIDMD often converges rapidly to the optimal solution. Over time, as more data are incorporated, the differences 
  between the reconstruction results of these methods diminish, ultimately showing minimal variance.
  
 Within this paper's theoretical framework, we can expand our analysis to include a variety of noise types beyond the traditional 
 Gaussian assumption, better reflecting real-world scenarios. This expansion is supported by the use of MLE, enabling the creation of an 
 optimized likelihood function that accurately matches the specific noise characteristics of the dataset.
 To augment the model's fidelity, physical constraints are integrated, enhancing solution accuracy by steering the Online proximal gradient 
 descent algorithm toward outcomes that are both mathematically robust and physically valid. 
 
 However, this nuanced approach reveals a shortfall in the Online DMD algorithm, which lacks the flexibility to adjust to non-Gaussian noise 
 types and integrate physical constraints as our method does. Consequently, while the online proximal gradient descent algorithm significantly 
 gains from these advancements, offering more precise and applicable solutions.
 
 \section{Statistical Analysis of DMD}

 \subsection{Bias and Variance Analysis in DMD Statistical Inference}\label{Bias_and_Variance}
 From the frequentist perspective, the estimator $A(x_1, x_2, \ldots, x_{m+1})$ serves as a function based on the sample data, 
 aiming to approximate the true value $A^*$. This approach is employed to deduce the unknown parameters 
 $A^*$ from observed data ${x_1, x_2, \ldots, x_{m+1}}$. To understand how physical constraints affect the estimator $A$, it is crucial to examine
  both the bias and variance of the estimated matrix $A$ in comparison to the actual parameter matrix $A^*$. 
 The bias of the estimator $A$ is defined as the expected deviation from the true parameter matrix:
 $$
  \text{Bias}(A) = \mathbb{E}[A - A^*].
 $$
 The estimator $A$ is unbiased if it holds that $\mathbb{E}[A - A^*] = 0$. The variance of the estimator, denoted
 $$
   \text{Var}(A) = \mathbb{E}\left[(A - \mathbb{E}[A])(A - \mathbb{E}[A])^T\right]
 $$
 measures the spread of the estimator values around their expected value,  which does not directly depend on the true parameter $A^*$. 
 For analytical convenience, we also consider the vectorized form of \( A \), represented as \( \hat{A} \), and the true 
 vectorized parameter \( \hat{A}^* \).
 $$
   \text{Bias}(\hat{A}) = \mathbb{E}[\hat{A} - \hat{A}^*], \quad \text{Var}(\hat{A}) = \text{Cov}[\hat{A}]
 $$
 Under the mean squared error (MSE) framework, the bias and variance of the estimator $A$ are connected through the following relationship:
 \begin{align*}
   \text{MSE}(A) &= \mathbb{E} \left[ \|A - A^*\|_F^2 \right] \\
   &=\mathbb{E} \left [ \|A - \mathbb{E}[A]\|_F^2 + \|\mathbb{E}[A] - A^*\|_F^2  \right ]\\
   &=\text{tr}\left [ \text{Var}\left ( A \right )  \right ] +\left \| \text{Bias}\left ( A \right )  \right \|_F^2 \\
   &=\text{tr}\left [ \text{Var}\left ( \hat{A} \right )  \right ] +\left \| \text{Bias}\left ( \hat{A} \right )  \right \|_2^2 
 \end{align*}
 
 The trade-off between bias and variance is a fundamental aspect of statistical learning, where reducing variance may increase bias, 
 and vice versa. This trade-off is particularly evident in regularized versions of the piDMD algorithm.
 The regularization(physics-constrained) term imposes a penalty on the structure of $A$, which can influence the estimator's bias and variance in meaningful ways.

 From the perspective of machine learning, the regularizer $R(A)$ , such as $\left \| A \right \|_F^2$ and $\|A\|_1$, is chosen not only to 
 incorporate physical constraints but also to limit the complexity of the model. 
 By incorporating the $l_2$ norm into the optimization, we extend the standard DMD algorithm with ridge regression, which can be expressed as:
 \begin{align}\label{eqc_{19}}
   A &= \underset{A}{\mathrm{argmin}}\left [ \left\|Y - AX\right\|_F^2+\lambda \left \| A \right \|_F^2\right ]  \\
   &=\underset{A}{\mathrm{argmin}}\left [\text{tr}(X^TA^TAX+\lambda A^TA-2Y^TAX)\right ]\label{eqc_{20}}
 \end{align}
 The simplicity of the $l_2$ norm renders this form of regularization particularly amenable to quantitative analysis.
  By incorporating this $l_2$ regularization, we are equipped to systematically evaluate the interplay between model 
  complexity—governed by the regularization term—and model fit—dictated by the reconstruction loss—and their respective 
  effects on the estimator's bias and variance. To optimize Equation \eqref{eqc_{20}}, we calculate the derivatives with respect to $A$
 and set these derivatives equal to zero\cite{petersen2008matrix}, resulting in the solution:
 $$
     A=YX^T(XX^T+\lambda I)^{-1}
 $$
 Even when $X$ (and hence $X^TX$) is not full rank, ($X^TX + \lambda I$) is always symmetric and positive definite  since
 we are adding $\lambda > 0$ to all the diagonal elements of $X^TX$. Therefore $A$ always exists and is unique.
 
 To analyze the bias and variance of the estimator \(A\) in standard DMD with the \(l_2\) norm, we revisit the probabilistic interpretation of DMD. 
 This assumes a relationship $y_i=A^*x_i+\epsilon_i$ where each $\epsilon_i \sim N(0,\sigma^2I)$ being independently and identically distributed.
  We assume $X$ is given, and hence
 constant. For notational simplicity let $\epsilon \sim \mathcal{MN}_{n\times m}(0,I,\sigma^2I)$ where $\epsilon_i \sim \mathcal{N}(0,\sigma^2I)$.
 So, $Y = A^*X + \epsilon$. Consider the singular value decomposition of matrix $X$ :
 $$
   X=U\Sigma V^T=
   \left[\begin{array}{ccc}
     \mid & & \mid \\
     u_1 & \cdots & u_n \\
     \mid & & \mid
     \end{array}\right]
     \left[\begin{array}{ccc|c}
       \sigma_1 & & & \\
       & \ddots & & 0 \\
       & & \sigma_r & \\
       \hline & 0 & & 0
       \end{array}\right]
     \left[\begin{array}{ccc}
       \mid & & \mid \\
       v_1 & \cdots & v_m \\
       \mid & & \mid
       \end{array}\right]^T
 $$
 where \( \Sigma \in \mathbb{R}^{n \times m} \) contains the singular values \( \sigma_1, \dots, \sigma_r \) and the columns of \(U \in \mathbb{R}^{n \times n}  \) and \(V  \in \mathbb{R}^{m \times m}  \) represent the left and right singular vectors \(u_i\) and \(v_i\), respectively. This implies
 $$
   (XX^T+ \lambda I)^{-1}=U\left[\begin{array}{ccc}
     \frac{1}{\sigma_1^2+\lambda}  & \ldots & 0 \\
     \vdots & \ddots & \vdots \\
     0 & \ldots & \frac{1}{\sigma_n^2+\lambda} 
     \end{array}\right] U^T
 $$
 Here, \(\sigma_1, \dots, \sigma_r\) are the non-zero singular values, with the assumption that \(\sigma_{r+1}^2, \dots, \sigma_n^2\) are 0, 
 thus facilitating the inversion process. Now we analyze the Bias of $A$  and the Variance of $\hat{A}$. We start with the expression for the estimator
 \begin{align*}
   A &=YX^T(XX^T+\lambda I)^{-1}  \\
   &=A^*XX^T(XX^T+\lambda I)^{-1}+\epsilon X^T(XX^T+\lambda I)^{-1}
 \end{align*}
 To compute the Bias of this model, we take the expectation of the above and
 observe that:
 \begin{align*}
   \mathbb{E} \left [ A \right ] &=\mathbb{E} \left[ A^*XX^T(XX^T+\lambda I)^{-1}+\epsilon X^T(XX^T+\lambda I)^{-1} \right]\\
   &=A^*\left[ XX^T(XX^T+\lambda I)^{-1}\right]+\mathbb{E}\left[\epsilon\right] X^T(XX^T+\lambda I)^{-1}  \\
   &= A^*U\left[\begin{array}{ccc}
         \frac{\sigma_1^2}{\sigma_1^2+\lambda}  & \ldots & 0 \\
         \vdots & \ddots & \vdots \\
         0 & \ldots & \frac{\sigma_n^2}{\sigma_n^2+\lambda} 
         \end{array}\right] U^T
 \end{align*}
 
 This result demonstrates the estimator $A$ experiences a shrinkage effect. This effect is quantified by the reduction 
 in the eigenvalues of the matrix, each becoming less than unity. The magnitude of shrinkage is directly proportional to the
 value of $\lambda$; higher values of $\lambda$ induce stronger shrinkage, leading to a significant reduction in the eigenvalues,
 thereby pushing $A$ towards zero. This phenomenon illustrates that the $l_2$-regularized DMD algorithm's estimator is biased,
 particularly towards zero.

 While the introduction of regularization imparts a bias to the estimator, it concurrently yields a reduction in variance. 
 This trade-off is a critical aspect in statistical learning and predictive modeling. To facilitate a more detailed analysis 
 of the estimator's variance, we employ the vectorized forms of $A$ and $\epsilon$, and leverage properties of the matrix
 Gaussian distribution. Specifically, we utilize the following property\cite{gupta2018matrix}:
 $$
   X \sim \mathcal{MN}_{n\times m}(M, E, F)  \quad XA \sim \mathcal{MN}_{n\times m}(MA, E, FAA^T) 
 $$
 $$
  \mathrm{vec}(X) \sim \mathcal{N}_{nm}(\mathrm{vec}(M), F \otimes E) \quad \mathrm{vec}(XA) \sim \mathcal{N}_{nm}(\mathrm{vec}(M), FA A^T \otimes E)
 $$
 This gives us
 \begin{align*}
   \text{Var}[\hat{A}] &=\text{Cov} \left [ \mathrm{vec}(A^*XX^T(XX^T+\lambda I)^{-1}+\epsilon X^T(XX^T+\lambda I)^{-1})\right ] \\
   &=\sigma^2I \left ( X^T(XX^T+\lambda I)^{-1} \right ) \left ( X^T(XX^T+\lambda I)^{-1} \right ) ^T \otimes I\\
   &=V\left[\begin{array}{ccc}
     \frac{\sigma_1^2\sigma^2}{(\sigma_1^2+\lambda)^2}  & \ldots & 0 \\
     \vdots & \ddots & \vdots \\
     0 & \ldots & \frac{\sigma_m^2\sigma^2}{(\sigma_m^2+\lambda)^2} 
     \end{array}\right] V^T \otimes I 
 \end{align*}
 where  $\sigma_1,\dots ,\sigma_r$ are the singular values, where $\sigma_{r+1}^2,\dots ,\sigma_m^2$ are 0. Because
 $$
 \text{tr}(A \otimes B) = \text{tr}(A)\text{tr}(B) = \text{tr}({\Lambda}_A \otimes \Lambda_{B})
 $$
 Where \({\Lambda}_A\) and \({\Lambda}_B\) denote the diagonal matrices containing the eigenvalues of \(A\) and \(B\), respectively \cite{petersen2008matrix}. This implies
 \begin{align*}
   \text{tr}\left [ \text{Var}\left ( A \right )  \right ]&=\text{tr}\left [ \text{Var}[\hat{A}] \right ] \\
   &=\text{tr}\left [ \sigma^2I  X^T(XX^T+\lambda I)^{-1}  (XX^T+\lambda I)^{-1}X \right ]   \cdot \text{tr} \left [ I \right ] \\
   &=n  \left ( \sum_{i=1}^{r} \frac{\sigma_i^2\sigma^2}{(\sigma_i^2+\lambda)^2}  \right )
 \end{align*}
 This analysis demonstrates that an increase in $\lambda$ leads to a decrease in the eigenvalues of $\hat{A}$'s covariance matrix, 
 which in turn reduces $\text{tr}\left [ \text{Var}\left ( \hat{A} \right ) \right ]$. Through regularization by adjusting $\lambda$, we
  effectively lower variance, thereby rendering the estimate less susceptible to data's random fluctuations. Nonetheless, this benefit 
  of variance reduction is offset by an increase in bias; a higher $\lambda$ makes the estimator diverge from the true parameter matrix
   $A^*$, thereby infusing a systematic error.
 
 The implications of this trade-off extend beyond this mathematical formulations into the realm of practical application in 
 statistical learning. Regularization serves as a mechanism to prevent overfitting by penalizing the complexity of the
  model, effectively compressing the spread of the estimator's distribution. This is particularly valuable in dynamic 
  systems analysis where the data might be prone to overfitting due to high dimensionality or collinearity among variables.
 
 Similarly, the integration of physical constraints within the DMD framework parallels regularization's role. Enforcing established
  physical laws or behaviors on the modeled system inherently constrains the model's solution space, much like regularization's complexity 
  penalty.  Nonetheless, overly stringent physical constraints can also introduce bias, potentially oversimplifying the model and thus failing to fully capture the 
  system's dynamics.
 
 \subsection{Bias and Variance in Prediction}
 In predictive settings, our objective diverges from that of standard statistical inference. Rather than estimating an unknown parameter $A$, 
 the goal is to learn a function $f: \mathbb{R}^n \rightarrow \mathbb{R}^n  $ capable of accurately predicting $y_i$ given $x_i$. This shift in focus underscores the importance of choosing
 an appropriate $\lambda$ that optimizes the estimator's performance for prediction, balancing the need for accuracy with the ability to generalize 
 effectively to new data.
 
 In advancing our understanding of statistical learning,  we commence with a fundamental assumption: $y = f(x)+ {\epsilon }' $, 
 where ${\epsilon }' \in \mathbb{R}^n$ satisfies $\mathbb{E}[{\epsilon }' ] = 0$ and $\text{Var}[{\epsilon }' ] = \tau ^2I$. It is important to note that ${\epsilon }' $ need 
 not necessarily follow a Gaussian distribution. Furthermore, we define the true function $f$ as\cite{bishop2006pattern}:
 $$
   f(x')=\mathbb{E} \left [ y|x=x' \right ]
 $$
 Let us consider $\hat{f}$ as the estimator obtained through a unspecified training process
 over the data sets $X$ and $Y$. Given a new, unseen example pair $(x, y)$, we aim to evaluate the 
 corresponding generalization error.
 This error is defined with respect to the variability inherent in $\epsilon$ within the test example and the trained estimator $\hat{f}$\cite{bishop2006pattern}:
 \begin{align*}
   \text{MSE}(\hat{f}) &=\mathbb{E} \left[ \|y-\hat{f}(x)\|_2^2 \right] \\
   &=n\tau^2+\left \| \text{Bias}\left ( \hat{f}(x) \right )  \right \|_2^2 + \text{tr}\left [ \text{Var}\left ( \hat{f}(x) \right ) \right ]   
 \end{align*}
 where \(\text{Bias}( \hat{f}(x)) \) is defined as \(\mathbb{E} [ \hat{f}(x) - f(x) ]\), and the \(\text{Var}( \hat{f}(x) ) \) is 
 given by \(\text{Cov}[\hat{f}(x)]\). This decomposition reveals parallels with the statistical inference setting. The term $n\tau^2$ denotes the 
 irreducible error, while the Bias and Variance terms represent the centered first and second moments of $\hat{f}$, respectively. Linking these 
 concepts to the $l_2$-regularized DMD algorithm, we can discern the relationship between the Bias and
 Variance in both prediction and inference frameworks:
 \begin{align*}
   \left \| \text{Bias}\left ( \hat{f} \right )  \right \|_2^2 &=\left \| \mathbb{E} \left[  \hat{f}(x)-f(x)\right] \right \|_2^2  \\
   &=  \left \| \mathbb{E} \left[ Ax-A^*x\right] \right \|_2^2 \\
   &=\left \| \text{Bias}\left ( A \right )x\right \|_2^2 
 \end{align*}
 and
 \begin{align*}
   \text{tr}\left [ \text{Var}\left ( \hat{f}(x) \right ) \right ]   &=\text{tr}\left [ \mathbb{E}\left ( \left ( \hat{f}(x)-f(x)  \right )\left (  \hat{f}(x) -f(x) \right )^T \right )     \right ]   \\
   &=\text{tr}\left [ \mathbb{E}\left ( \left ( A-A^*  \right )xx^T\left ( A-A^* \right )^T \right ) \right ]  \\
   &=\text{tr}\left [ xx^T\mathbb{E}\left ( \left ( A-A^*  \right )\left ( A-A^* \right )^T \right ) \right ]\\
   &=\text{tr}\left [ xx^T\text{Var}\left ( A \right ) \right ]
 \end{align*}
 In predictive settings, irreducible error arises from inherent noise in test data, which influences the variance observed in training data and manifests as an irreducible error term in testing. This error, rooted in the dataset’s intrinsic noise, cannot be minimized through modeling efforts. Therefore, the focus shifts towards optimizing the Bias-Variance trade-off to enhance model performance. In \(l_2\)-regularized DMD algorithms, the regularization parameter \(\lambda\) is key for balancing Bias and Variance. Increasing \(\lambda\) typically lowers Variance at the expense of increasing Bias, impacting model precision on unseen data. The optimal \(\lambda\) is found by minimizing the squared or generalization error through cross-validation, which evaluates model performance across various subsets of the dataset.
 
 Effectively managing the Bias-Variance trade-off involves simultaneous assessment of training and cross-validation errors. This approach allows for a detailed analysis of the components affecting model performance, enabling targeted adjustments to either Bias or Variance based on their impact on accuracy.
 
 \subsection{Bias and Variance Analysis in Physics-Informed Dynamic Mode Decomposition}
 The DMD method is commonly used for high-dimensional systems, where the dimension of the state variables \( n \) is much larger 
 than the amount of data \( m \) (i.e., \( n \gg m \)). In such cases, the dimension of the linear operator learned by DMD is \( n \times n \), while the actual
  dataset dimension is only \( n \times m \). This situation, where the number of parameters far exceeds the amount of data, can lead to model overfitting if no
   constraints are imposed on the linear operator, thereby reducing its ability to accurately capture the system's core dynamical characteristics.
 
   To address this, Exact DMD introduces a low-rank constraint to improve performance in noisy data scenarios\cite{tuDynamicModeDecomposition2014}. Low-rank constraints not 
 only effectively remove noise but also enhance DMD's accuracy in modal analysis and prediction. It is worth noting that even under ideal noise-free conditions, 
 the high dimensionality of the data and the limited sample size can result in the fitted Standard DMD matrix containing spurious modes. These spurious modes 
 can adversely affect analysis results, causing the quality of modal analysis and prediction to significantly deviate from the system's true dynamical characteristics.
 
 The role of low-rank constraints extends beyond noise suppression; they also simplify the system’s complexity, making the DMD matrix more inclined to capture 
 the principal features of the data, thereby enhancing model robustness and reliability. From a bias-variance perspective, low-rank 
 constraints can be seen as a variance reduction strategy.  Although introducing low-rank constraints adds some bias, it significantly reduces the system’s 
 degrees of freedom, making the linear 
  operator less reliant on the data alone and selectively ignoring minor details. This focuses the DMD matrix on capturing the dominant dynamics of the data, 
  ultimately improving its ability to describe the essential behavior of the system.
 
 The piDMD algorithm extends the constraints in Exact DMD, enabling the selection of constraints more appropriate than low-rank constraints based on the 
 characteristics of different physical systems. However, these algorithms are primarily designed for cases where \( n \gg m \), making them particularly 
 suitable for scenarios involving simulated data of physical systems. In practice, physical systems are often monitored through sensor networks. Due to
  a limited number of sensors and the need for a high sampling frequency to accurately capture the system’s 
 dynamics, the state variable dimension \( n \) is relatively low, while the sample count \( m \) is large. Directly applying the above methods in this 
 context significantly increases computation time.
 
 To address this issue, this paper proposes the OPIDMD algorithm to improve real-time computational performance. It employs the online proximal gradient descent
  algorithm to add physical constraints, thus requiring the OGD algorithm as the foundational optimization step. When certain properties 
  of the linear operator (e.g., sparsity or specific symmetry) are known, these physical constraints can be introduced via a proximal operator to further
   enhance algorithm performance. Therefore, regardless of the problem type, OPIDMD's prediction accuracy should generally exceed that of a standalone
    OGD algorithm. Although the proximal operator is applicable only to 
    linear operators, the OPIDMD algorithm’s linear operator learned in nonlinear systems has time-varying characteristics, allowing each step's linear 
    model to serve as a local linear approximation of the nonlinear dynamical system based on the Taylor expansion principle. This feature makes OPIDMD 
    more effective than existing DMD methods for accurately capturing the system's dynamics in short-term predictions.

    \section{Computational Time}
    \subsection{Comparison with the Online Algorithm}
    Finally, we evaluate the computational time of the OPIDMD algorithm and other algorithms, providing a comprehensive analysis of its performance compared to existing methodologies. All the experiments were conducted in Python on a PC equipped with a 3.4 GHz Intel Core i7-14700K processor.
    Our comparative analysis begins by evaluating various algorithms in terms of the state dimension \( n \) and the number of snapshots \( k \), as shown in Table \ref{table:characteristics_dmd_algorithms}. The table illustrates the computational complexity of performing the proximal gradient descent method under different physical constraint conditions.
    \begin{table}[!h]
    \centering
    \renewcommand{\arraystretch}{1.3} 
    \caption{Computational Efficiency of DMD Algorithms}
    \label{table:characteristics_dmd_algorithms}
    \begin{tabular}{lllll}
    \hline
    Aspect & Comp. Time & Past Snapshots & Real-Time & Exact piDMD \\
    \hline
    Implicit Constraint  & \( O(n^2) \)  & No & Yes & Yes \\
    Shift-Invariant  & \( O(n^2) \)  & No & Yes & Yes \\
    Self-Adjoint    & \( O(n^2) \) & No & Yes & Yes \\
    Causal          & \( O(n^2) \) & No & Yes & Yes \\
    Local           & \( O(n^2) \)  & No & Yes & Yes \\
    Low-rank        & \( O(n^3) \)  & No & Yes & Yes \\
    Sparse          & \( O(n^2) \)  & No & Yes & Yes \\
    Norm Shrinkage & \( O(n^2) \)  & No & Yes & Yes \\
    Online DMD         & \( O(n^2) \)  & No & Yes & Yes \\
    Streaming DMD      & \( O(r^2n) \) & No & Yes & No \\
    Standard DMD       & \( O(kn^2) \) & Yes & No & Yes \\
    piDMD        & Case-by-Case & Yes & No & Yes \\
    \hline
    \end{tabular}
    \vspace*{-4pt}
    \end{table}

    Given the limited availability of online algorithms specifically designed for piDMD, our investigation contrasts computational times with those 
    of online DMD and streaming DMD algorithms, employing OGD for comparison. For streaming DMD, operating under a fixed rank $r$ of 100.
    \begin{figure}[!ht]
      \centering
      \begin{subfigure}{0.41\textwidth}
          \centering
          \includegraphics[width=\textwidth]{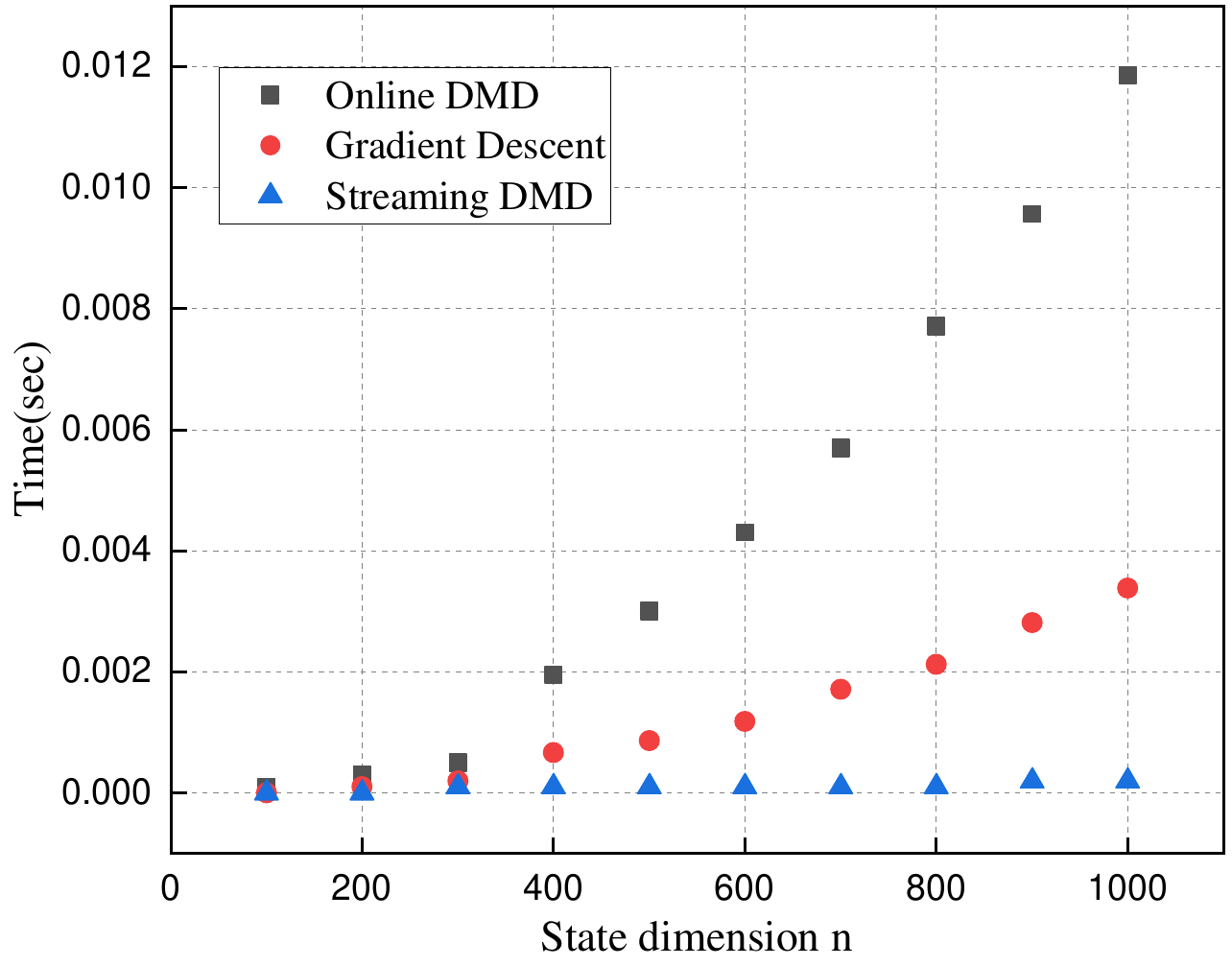}
          \caption{Compute DMD matrix at each step}
      \end{subfigure}
      \hspace{2em}
      \begin{subfigure}{0.4\textwidth}
          \centering
          \includegraphics[width=\textwidth]{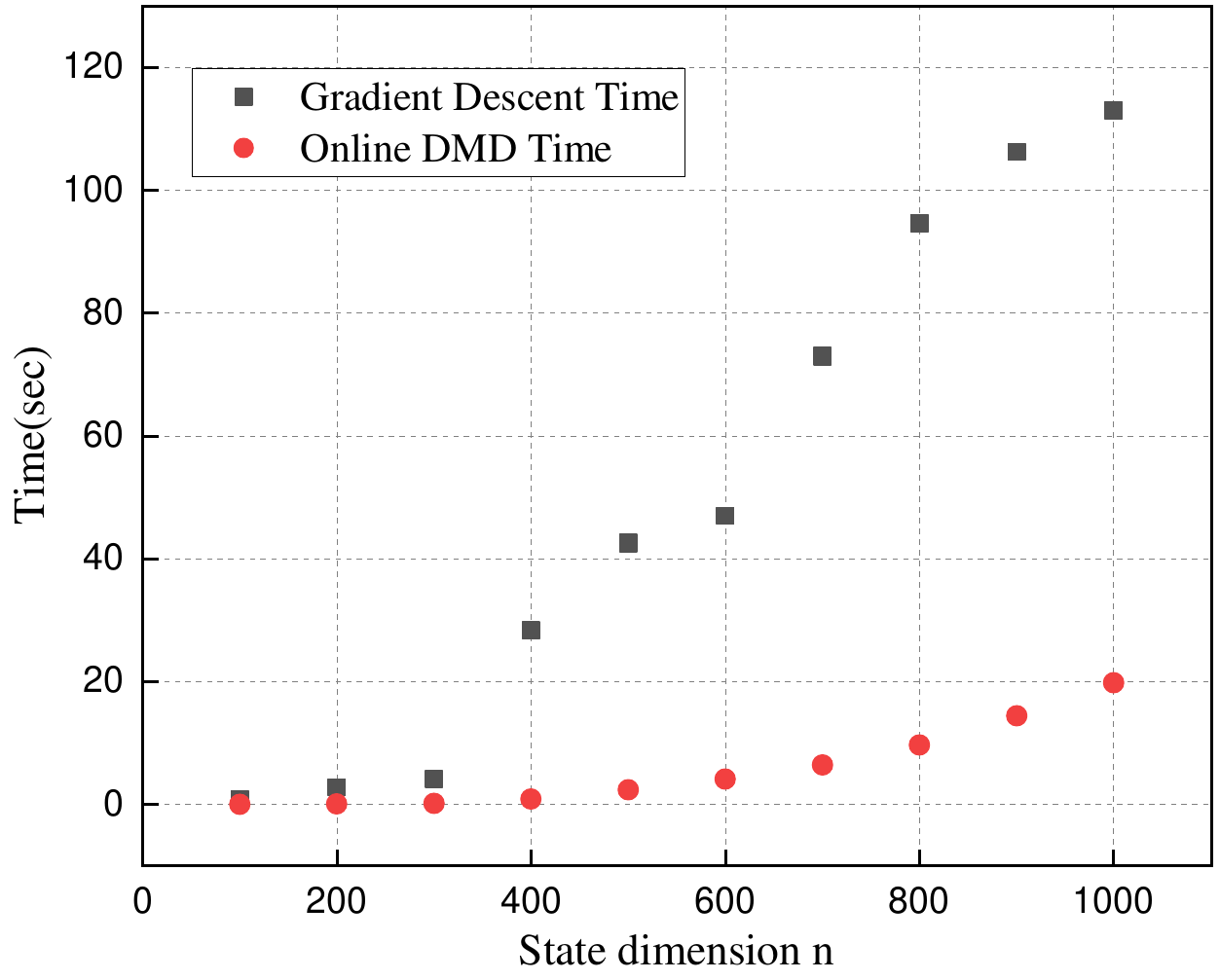}
          \caption{Compute DMD matrix at final step}
      \end{subfigure}
      \caption{Performance of the different Online DMD algorithms}
      \label{fig9}
    \end{figure}
    The comparison reveals that in scenarios with expansive state dimensions, streaming DMD emerges as the most expedient algorithm due to 
    its linear scalability with the state dimension—unlike the quadratic scaling observed with alternative algorithms. It is crucial, however,
    to recognize that streaming DMD approximates by projecting onto a subspace of dimension $r$, thus not yielding the exact DMD matrix. In contrast, 
    online DMD and OGD algorithms are capable of computing the full DMD matrix accurately, without resorting to approximations,
    offering a comprehensive view of the different DMD algorithms' performance on benchmark cases as depicted in Figure \ref{fig9}.
    
    As shown in Figure \ref{fig9}(a), the OGD algorithm demonstrates a computational time advantage in a single run compared to online DMD. However, in part (b) of Figure \ref{fig9}, when considering the time required for both algorithms to converge to the accurate matrix \( A_k \) through multiple iterations, online DMD shows a greater advantage. 
    Although initial comparisons indicate that gradient descent has a longer convergence time than online DMD, the examples provided in the text reveal that when the underlying laws of the physical system are known and the data contains noise, OPIDMD often converges faster than online DMD. Additionally, OPIDMD is less susceptible to noise interference.
    
    \subsection{Comparison with piDMD}
    The analysis in this paper is based on a common scenario: the system evolves slowly over time, and data arrives continuously (e.g., as in real-world sensor systems that provide a constant stream of data). In this context, the model makes incremental updates at each time step based on newly received data, aiming to achieve a more accurate representation of the system.

    When a new set of data arrives, the OPIDMD method can perform a single-step update to refine the model, ensuring an accurate system model at the current time step. In contrast, piDMD cannot incrementally adjust parameters based on the existing model; instead, it requires re-training with the entire dataset to reconstruct the model with each update.
    
\begin{figure}[!ht]
      \centering
      \begin{subfigure}{0.375\textwidth}
          \centering
          \includegraphics[width=\textwidth]{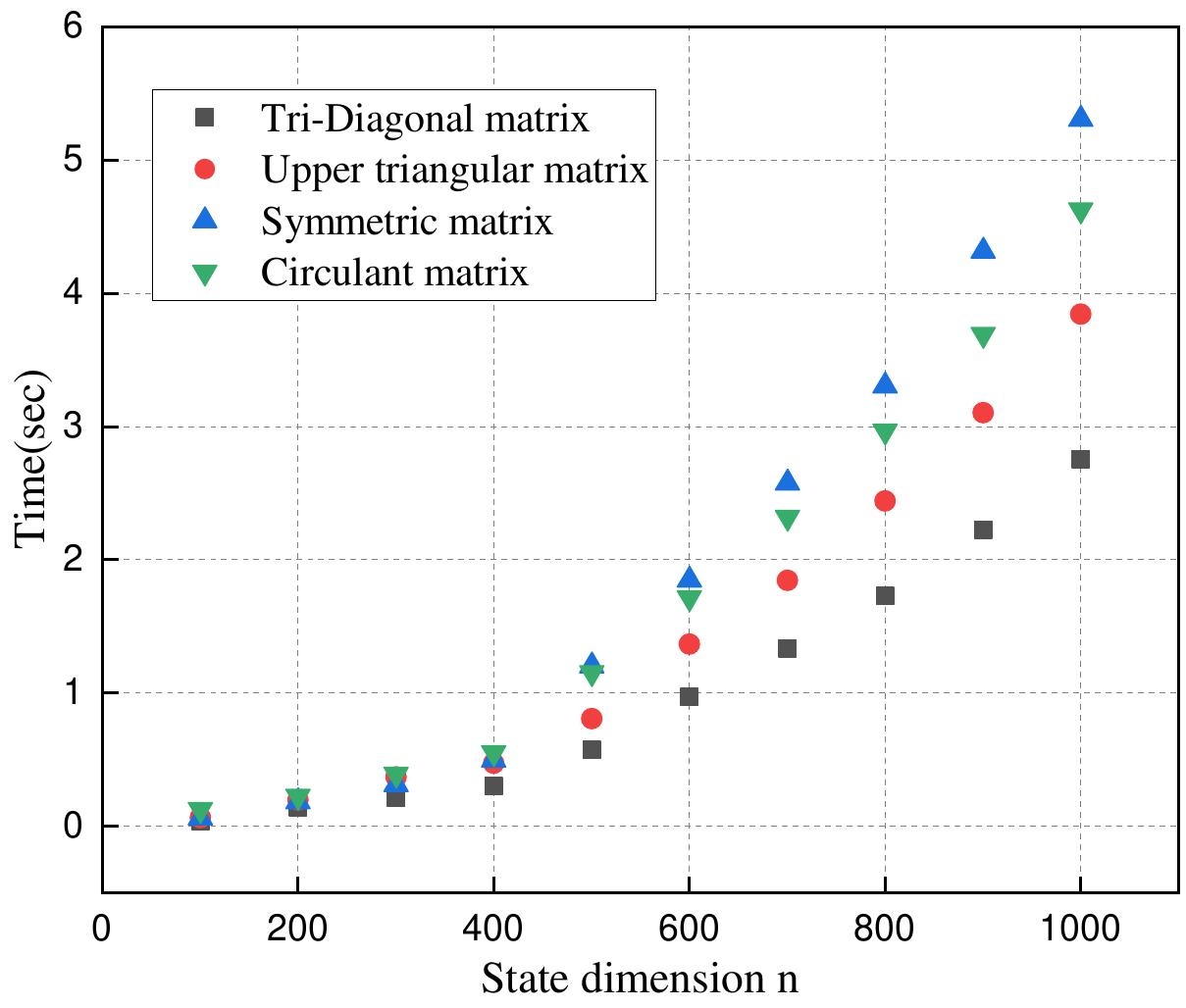}
          \caption{piDMD matrix at final step}
      \end{subfigure}
      \hspace{1em}
      \begin{subfigure}{0.4\textwidth}
          \centering
          \includegraphics[width=\textwidth]{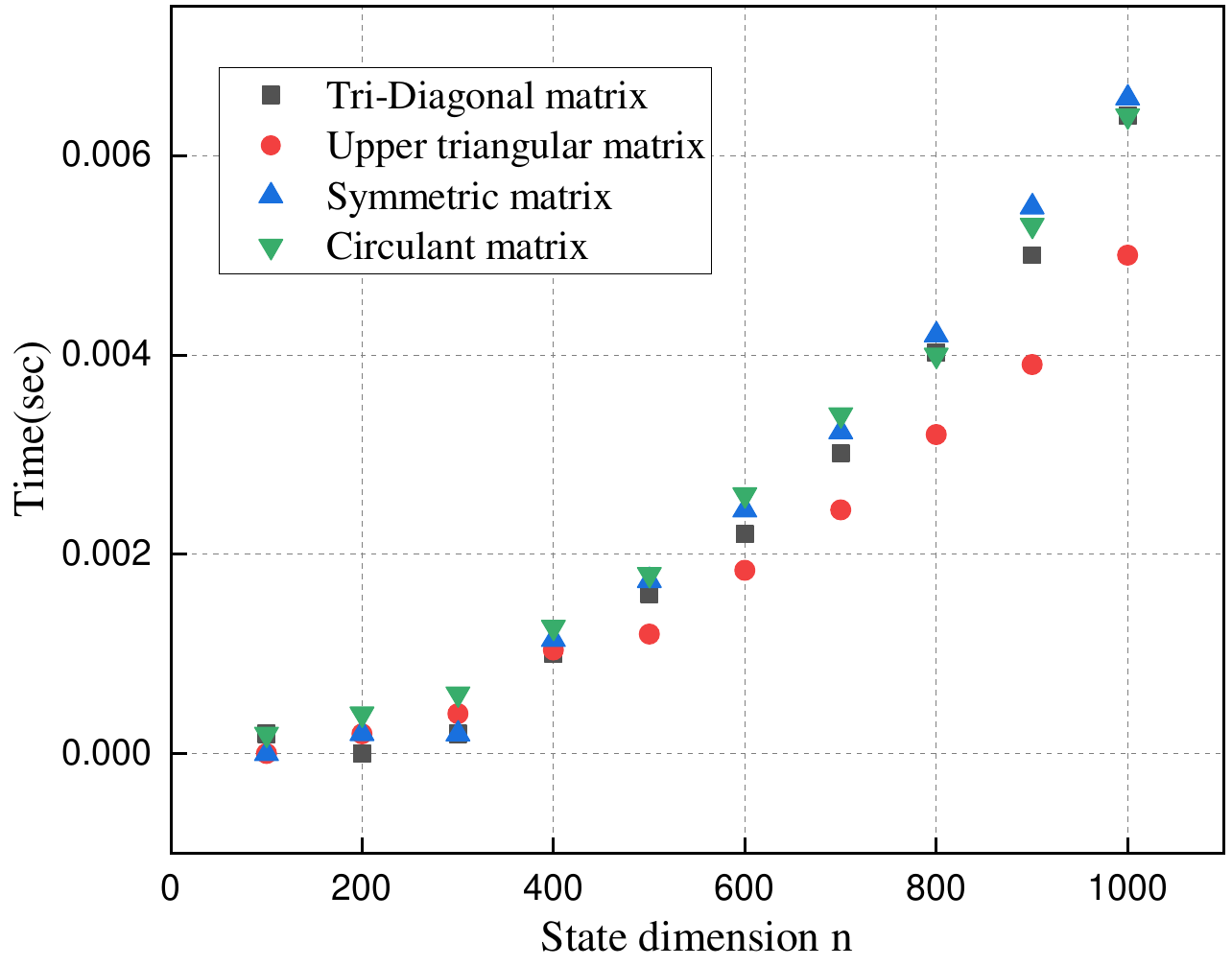}
          \caption{OPIDMD matrix at each step}
      \end{subfigure}
      \caption{Performance of the different piDMD algorithms}
      \label{fig:10}
    \end{figure}

    Thus, Figure \ref{fig:10} compares the computational time for OPIDMD and piDMD under different physical constraints across various system dimensions \(n\). Part (a) shows the computation time for piDMD across the entire dataset, using the computational method proposed in \cite{baddooPhysicsinformedDynamicMode2023}. Part (b) illustrates the computation time per run for OPIDMD under different dimensional settings \(n\).
    
    It's important to note that OPIDMD requires a certain number of iterations to converge to an optimal solution. From this perspective, if the goal is solely to determine the final step's \( A_k \), OPIDMD might not always offer a temporal advantage. 
    However, given the scenario outlined in this paper, the time required for OPIDMD to update the model in a single step demonstrates a significant performance advantage.
     This makes it exceptionally well-suited for dynamic environments where timely updates are crucial. This capability ensures that OPIDMD remains a valuable tool for modeling time-varying complex systems, particularly when real-time processing and adaptability are paramount.
    
    The OPIDMD algorithm is particularly suited for integration with modern feedback control algorithms, especially in real-time 
    operational scenarios. By continuously updating the system model and utilizing the updated linear operator for short-term predictions, OPIDMD can 
    provide accurate model support for Model Predictive Control (MPC), thereby enhancing control precision and efficiency in real-time. This feature 
    suggests that OPIDMD has promising application prospects in dynamic and complex real-time control applications.




\end{appendices}


\bibliography{sn_bibliography}

\end{document}